\documentclass{article} 

\usepackage{iclr2027_conference,times}
\iclrfinalcopy 

\usepackage{booktabs}
\usepackage{graphicx}

\usepackage{latexsym}

\usepackage{url}

\usepackage[T1]{fontenc}

\usepackage[utf8]{inputenc}

\usepackage{inconsolata}
\usepackage{microtype}
\usepackage{graphicx}
\usepackage{subcaption}
\usepackage{booktabs} 

\newcommand{\param}[1]{\texttt{\{#1\}}}

\usepackage{amsmath}
\usepackage{wrapfig}

\usepackage{amssymb}
\usepackage{mathtools}
\usepackage{amsthm}

\usepackage{tcolorbox}
\usepackage{listings}
\theoremstyle{plain}

\theoremstyle{definition}

\theoremstyle{remark}

\usepackage[textsize=tiny]{todonotes}
\newcommand{\domainA}{\texttt{2D Motion}}
\newcommand{\domainB}{\texttt{2D Gravity}}
\newcommand{\domainC}{\texttt{Fluids}}
\newcommand{\domainD}{\texttt{Bouncing}}
\newcommand{\domainE}{\texttt{Pendulum}}
\newcommand{\domainF}{\texttt{Rolling on Slope}}
\newcommand{\domainG}{\texttt{Collision}}
\newcommand{\domainH}{\texttt{Rolling on Track}}
\newcommand{\domainI}{\texttt{Pulley System}}

\title{Chain-of-Time: In-Context Physical \\ Simulation with Image Generation Models}

\author{Yingqiao Wang \\
Department of Psychology \\
Harvard University \\
\texttt{yingqiaowang@g.harvard.edu} \\
\And
Eric Bigelow \\
Department of Psychology, Harvard University \\
Goodfire AI \\
\texttt{ebigelow@g.harvard.edu} \\
\AND
Boyi Li \\
NVIDIA Research \\
\texttt{boyil@nvidia.com} \\
\And
Tomer Ullman\thanks{Corresponding author.} \\
Department of Psychology \\
Harvard University \\
\texttt{tullman@fas.harvard.edu}
}

\begin{document}

\maketitle
\lhead{} 

\begin{abstract}
  We propose a novel cognitively-inspired prompting method to stabilize and interpret the physical simulation produced by autoregressive image generation models, called \textit{Chain-of-Time}. \textit{Chain-of-Time} involves generating a series of intermediate images during a simulation, and it is motivated by mental simulation in humans, as well as in-context reasoning in LLMs. \textit{Chain-of-Time} is used at inference time, and requires no additional fine-tuning. We apply this method to synthetic and real-world domains, including 2-D graphics simulations and natural 3-D videos. These domains cover various physical dynamics, ranging from rigid-body rolling to harmonic oscillation and pulley systems. We found that \textit{Chain-of-Time} simulation stabilizes the physics simulated by the state-of-the-art image generation models. Beyond examining performance, we also analyzed the specific states of the world simulated by an image model at each time step, which sheds light on the dynamics underlying these simulations. This analysis reveals insights that are hidden from traditional evaluations of physical reasoning, highlighting cases where image generation models can simulate physical properties such as velocity, gravity, oscillations, and pulley systems. Like humans, image generation models are able to accurately predict future world states over short time periods. By providing a mechanism analogous to working memory, Chain-of-Time enables a model to stitch together a series of short-term predictions into an accurate long-term simulation.
\end{abstract}

\section{Introduction}

Recent developments 
%
%
enable auto-regressive image generation models (IGMs) to generate increasingly complex, realistic, and coherent images \cite{chen2025gpt4o_empirical, cao2025text, liu2023discovering, lu2024handrefiner}. Despite these improvements, however, these images often have distinct flaws, and may fail to capture real-world structures that are obvious to humans. 
Understanding the inner workings of Vision-Language Models (VLMs) such as IGMs has become a major topic in contemporary AI research \citep{dang2024explainable, chang2024survey, goh2021multimodal, bhalla2024interpreting, zhang2024vision}. 
In particular, there is a pressing question of how VLMs and image generation models represent physical properties which are required to predict how world states unfold over time.
In this work, we present a zero-shot prompting method for stabilizing physical reasoning in image generation models, which is inspired by psychological theories of mental simulation in humans and allows us to analyze the step-by-step process that the models use to simulate physics over time.

\begin{figure*}[t!]
    \includegraphics[width=\linewidth]{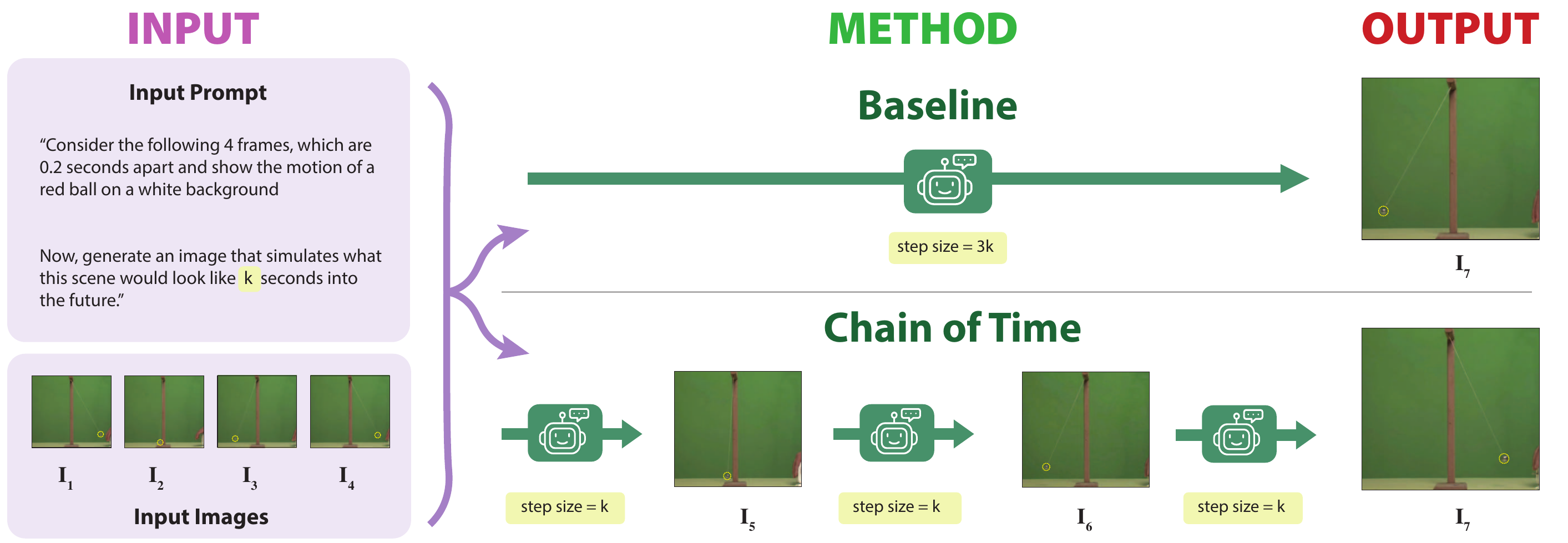}
    \caption{
    We study physical reasoning in multi-modal image generation models by providing the model with a sequence of input images of a scene at successive time steps, and having the model predict what the scene will look like some time in the future. 
    Accurately predicting future world states requires reasoning about latent physical properties such as gravity.
    We propose a novel method, Chain-of-Time (Bottom), which generates a sequence of images corresponding to a step-by-step simulation of the scene as it unfolds over time. As a baseline, Direct Prediction (Top) directly generates the final state without intermediate steps. 
    We use object segmentation algorithms to track object dynamics in both input and generated images (yellow dotted lines). 
    }
    \label{fig:intro-fig}
\end{figure*}

To our knowledge, our work is unique in studying physical reasoning abilities of autoregressive Image Generation Models (IGMs), which are similar to autoregressive LLMs, except that they can take both images and text as input and generate images or text as output. 
IGMs are rapidly becoming the gold standard for text-to-image generation, with top AI companies like OpenAI and Google replacing their flagship text-to-image models (e.g. DALL-E) with IGMs~\citep{openai_4o_addendum_2025, google_gem3_modelcard_2025}.
Given that modern IGMs build on top of pre-trained vision-language models \citep{openai_4o_image_generation_2025}, studying IGMs may also provide valuable insights into the inner workings of VLMs. 
Although our work is the first to study IGMs, prior work has evaluated the physical reasoning abilities of VLMs. Comprehensive benchmarks such as PhysBench \citep{chow2025physbench} and WM-ABench \citep{gao2025vision} test VLMs on a wide array of physical simulation capabilities. Beyond VLMs, \citet{meng2024phybench} evaluates the extent to which text-to-image models -- which generate images but do not take images as input -- can generate images matching relational and physical constraints, using a separate VLM as an evaluator.
Although these approaches provide valuable insights into how well VLMs can accomplish physical reasoning, they do not tell us \textit{how} VLMs accomplish physical simulation.
Our work aims to address this issue, providing a detailed analysis of the incremental processes underlying physical simulation.

We draw inspiration from approaches in cognitive science which empirically study analogous incremental processes when humans mentally simulate physical systems. 
%
Human mental simulation can be considered as a noisy Markov process where people simulate successive time steps in a latent world model, and there is also mounting evidence that working memory is essential for mental simulation in biological intelligence. 
We present a new method that endows IGMs with a module analogous to visual working memory, and enables them to perform physical simulation step-by-step, similar to what humans seem to do.

%
In this work, we adopt a theoretical framework of mental simulation from cognitive science to understand and stabilize physical reasoning and simulation abilities in Image Generation Models (Section 2). This framework helps us understand how IGMs reason about physical processes that unfold over time, by mapping input images to a latent state which is simulated with a Markov process to predict future time steps.
In order to both stabilize the physical reasoning abilities of IGMs, and to expose an interpretable trace of intermediate reasoning steps, we propose a novel method for in-context physical simulation, which we call Chain-of-Time (Section 3, Fig.~\ref{fig:intro-fig}).
%

\section{Mental Simulation in Humans}  \label{sec:mental-simulation}

People can reason efficiently about the physical dynamics of everyday objects, in tasks that state-of-the-art generative AI models struggle with. In this section, we will consider psychological theories of how people reason about physics in order to develop methods for evaluating and improving physical reasoning in IGMs.
For example, if you saw a pitcher full of juice begin to fall off of a table, you might quickly and intuitively predict what sequence of events would happen next. There are many competing theories that try to explain this `intuitive physics'. One current proposal is that people rely on an `internal physics engine' to carry out a mental simulation of a given scene \citep{battaglia2013simulation, ullman2017mind}. While this proposal has been criticized (see for example \citet{ludwin2021limits}), it finds support in cognitive science, computational modeling, cognitive development, and neuroscience  \citep{fischer2016functional, gerstenberg2021counterfactual, allen2021lifelong, fischer2021building, bass2021partial,balaban2025physics}. More recent work suggests it is likely that humans combine various computations to carry out physical reasoning, mental simulation being just one component \cite{hartshorne2025insights, sosa2025blending, smith2023integrating}. 


Research in cognitive science suggests that step-by-step mental simulation is an important component in human physical reasoning, and so we adopt this formalism for studying and improving upon the physical reasoning of current IGMs.
For our purposes here, we consider a basic version of the mental physics engine framework: Suppose that an agent observes an image $I$ that describes a scene at time $t$ in a pixel-based format, and wants to predict the state of the scene at a later time. A mental physics engine is a probabilistic transition function that can achieve this by composing three sub-components: a de-renderer $\phi$, a simulator $\tau$, and an optional renderer $\phi^{-1}$. The engine takes in the current image $I_t$, and de-renders it into the state of the world at that time, $X_t$. The engine then applies dynamic update rules to that state, corresponding to a transition $\tau$ that produces a distribution over future states of the world $X_{t+1}$. The engine may then render the state of the world back into a predicted image $I_{t+1}$.

A few notes on this overall formulation: 
First, while de-rendering has been studied in the context of intuitive physics in the past \citep[e.g.][]{wu2017learning,wu2017neural,smith2019modeling}, many other techniques exist for going from observations to physical states, and for our purposes here the specific technique is of less importance.
Second, while the images $I$ are pixel-based, the underlying physical state $X$ is not, and corresponds to the `game state' that describes in a lower-dimensional way the position, identity, and physical parameters of objects \citep{smith2019modeling}. 
Third, in computer graphics it is not strictly necessary to render the state of the world back into an image in order to answer various questions about the state, something that may hold for human mental physics as well \citep{balaban2025physics}.

More formally, the mental simulation formalism we consider here is:
{\small
\begin{align}
&p(X_t \ |  \ I_t) = \phi(I_t) + N(0,\sigma_{\phi}) &\textcolor{gray}{\text{De-rendering}}\nonumber \\
&p(X_{t+1} \ | \ X_t) = \tau(X_t) + N(0,\sigma_{\tau})&\textcolor{gray}{\text{Simulation}}\nonumber \\
&p(I_{t+1} \ | \  X_{t+1})  = \phi^{-1}(X_{t+1}) + N(0,\sigma_{\phi^{-1}}) \hspace{-5pt}&\textcolor{gray}{\text{Rendering}}\nonumber
\end{align}
}
The noise parameters $\sigma_{\phi}$, $\sigma_{\tau}$, and $\sigma_{\phi^{-1}}$  account for perceptual noise in the de-rendering of $I$, the cognitive complexity of mental simulation of the underlying state $X$, and imperfect imagery in the rendering of the state back to an image.

In this formulation, the state and scene at timestep $t+1$ depend only on the previous state and scene at timestep $t$. In other words, this is a linearly unfolding Markov Chain that moves from an initial observation $I_0$ step-by-step to a final state at time $T$, $X_T$, and optionally the predicted image at that time, $I_T$. While such step-by-step computations may underlie human mental simulation, it is unclear whether current IGMs tasked with predicting the future state of a scene $I$ at time $T$ go through a step-by-step simulation within their activations.
However, even if current IGMs do not do so implicitly, this framework suggests that if models can engage in short prediction steps, then with in-context reasoning we may be able to bring IGMs more in line with human-like physical simulation.

\section{Chain-of-Time Simulation}

\textit{Chain-of-Time} simulation is inspired by two bodies of prior literature: the cognitive science of mental simulation (described in Section~\ref{sec:mental-simulation}), and in-context reasoning in LLMs. 
In-context reasoning methods with LLMs coerce a model to spell out intermediate reasoning steps in its output stream before giving a final answer. This may be through prompting, as in Chain-of-Thought reasoning \citep{kojima2022large}, or through specialized training regimes \citep{guo2025deepseek, jaech2024openai}. 
These methods can significantly improve LLM performance on a variety of tasks, extending the ability of LLMs to reason over complex problems with many individual steps. Reasoning chains are also distinct from traditional LLM tasks, since intermediate steps can be directly inspected by humans to interpret what the model is doing. Although in some cases a model's intermediate reasoning tokens may not align with its final answer \citep{turpin2023language}, Chain-of-Thought reasoning has proved a valuable tool for auditing the behavior of language models. Various theories have been developed to try to explain precisely why and how these methods work or occasionally fail \citep{wang2022towards, merrill2023expressive, prystawski2023think}.

Based on the prior literature in human mental simulation and in-context reasoning, we propose a novel method for improving physical reasoning with in-context simulations. We call this method \textit{Chain-of-Time simulation} (Fig.~\ref{fig:intro-fig}).
We treat VLMs as a derenderer $\phi$ and a simulator $\tau$, and the IGMs as a renderer $\phi^{-1}$. The basic physical simulation task we consider is as follows: given a sequence of input images $I_{0:t}$ up to a given time $t$, generate a new image $\widehat{I}_{t+k}$ that accurately depicts what the scene will look like $k$ time steps into the future.

\begin{figure*}[t!]
    \includegraphics[width=\linewidth]{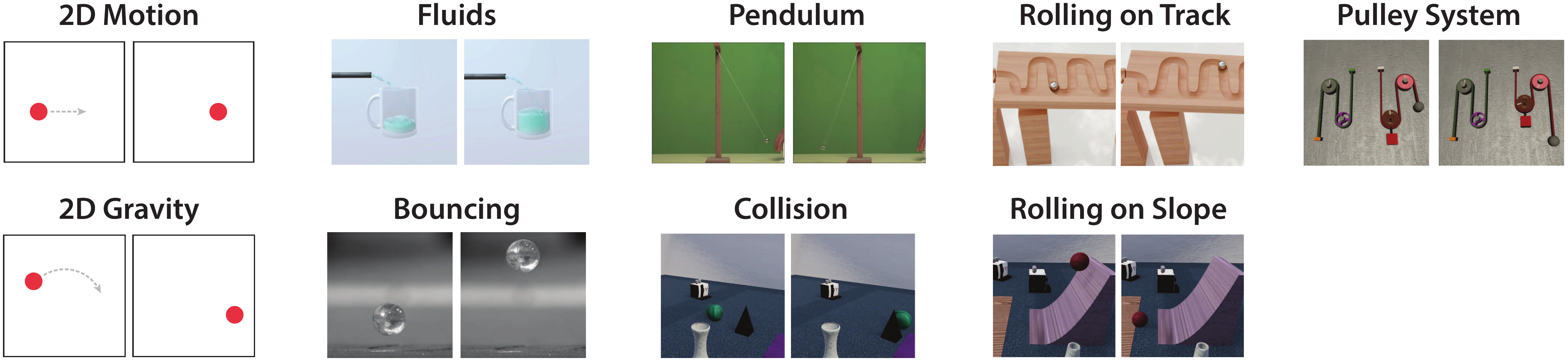}
    \caption{We use nine experimental domains to test models' ability to reason about different physical properties. Three of these domains (\domainA, \domainB, \domainC) were synthetically generated and inspired by previous work on the cognitive science of intuitive physics, four of the domains (\domainF, \domainG, \domainH, \domainI) were synthetically generated by prior works that evaluate physical reasoning in VLMs, and two of the domains (\domainD, \domainE) use real world videos sampled from various sources and demonstrate the generality of our approach.
    }
    \label{fig:intro-domains}
\end{figure*}

\begin{wrapfigure}{r}{0.5\textwidth}
    \includegraphics[width=\linewidth]{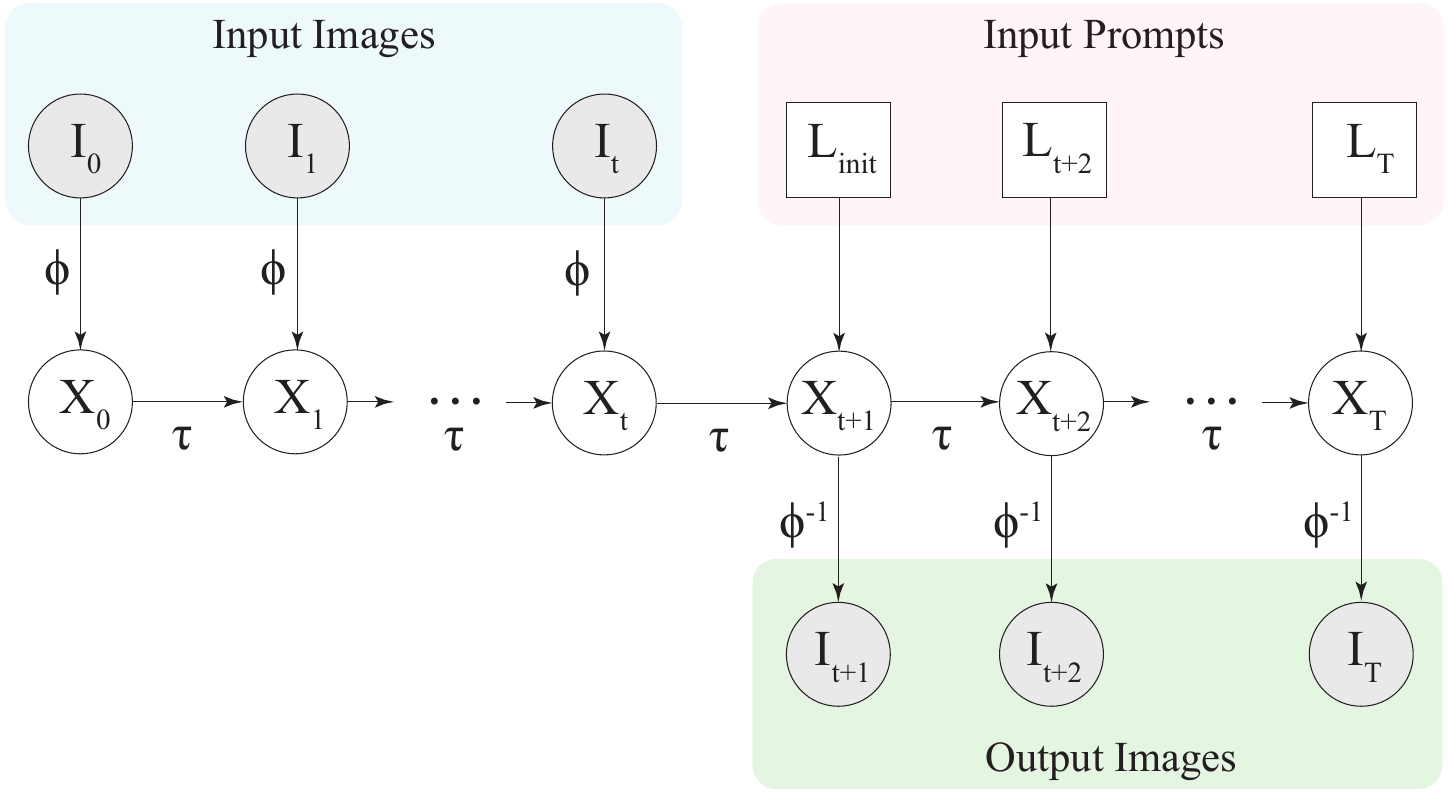}
    
    \caption{
        Chain-of-Time is a composition of three components: De-rendering $\phi$, Simulation $\tau$, and Rendering $\phi^{-1}$. De-rendering operates by converting input images $I_0 \ldots I_t$ into world states $X_t$, which represent a physical simulation over time. Chain-of-Time begins with an initial prompt $L_{\text{init}}$ and iteratively generates a sequence of in-context output images $I_{t+1} \ldots I_{T}$ with follow-up prompts $L_{t+1} \ldots L_T$
    }
    \label{fig:method-diagram}
\end{wrapfigure}

\textit{Chain-of-Time} simulation involves two prompts (provided in App.~\ref{app:prompts}): first, a Simulation Instruction prompt that, along with a sequence of input images, instructs the model to simulate an image $k$ seconds into the future. After the IGM generates a single image, we continue with our Simulation Follow-up prompt, which instructs the model to generate another image simulated an additional $k$ seconds into the future until $t+k = T$, where $T= t+t_{length}$. In our experiments, we used $T= t + 2 \ \text{secs}$ and the parameter $k$ is determined by the number of intermediate steps $s \in \{ 2  \text{ steps}, 4 \text{ steps}\}$, where we instruct the model to generate another image every $k = t_{length} / {s}$ seconds after time $t$, until it reaches time $T$.
To capture the Direct Prediction baseline, we use a method equivalent to Chain-of-Time but with a single timestep, prompting the IGM to predict $\widehat{I}_{t+k}$ given $I_{0:t}$, without intermediate prompts. In our case, to match the Chain-of-Time parameters, we consider $s= 1 \ \text{step}$ as the Direct Prediction baseline.

\paragraph{Previous Work}


Prior work has developed in-context reasoning methods for VLMs that use images to represent individual reasoning steps. Our method differs from these approaches in a few important ways. 
\citet{hu2024visual} proposed a method to solve simple reasoning problems with a VLM, such as geometry and spatial reasoning, and individual steps involve interleaved images and text outputs. Similar to our method, this approach requires no additional training. However, in contrast with \textit{Chain-of-Time}, this method is a separate tool that the VLM invokes, which uses LLM code generation along with a Python interpreter to render intermediate sketches.
\citet{xu2025visual} proposed a method for planning in which a VLM is explicitly trained with reinforcement learning to generate sequences of images to solve tasks such as maze navigation.
\textit{Chain-of-Time} simulation, on the other hand, strives to stabilize physical simulation with IGMs where ``steps'' correspond to units of time, rather than planning, where ``steps'' correspond to actions an agent takes in an environment. Our method also differs in that it can be applied to out-of-the-box IGMs with no additional training.
Further, our work presents an in-depth analysis of the sequence of images generated by \textit{Chain-of-Time}, where we consider the accuracy of simulation at each individual step. Such an analysis was not used in \citet{hu2024visual} or \citet{xu2025visual}, although \citet{xu2025visual} develops learning metrics that evaluate step-wise accuracy, and both works show examples of generated image sequences.

Our work is also similar to approaches for evaluating physical reasoning ability in video generation models \citep{bansal2024videophy, wiedemer2025video, motamed2025generative, yuan2025likephys, wei2025cophybench, le2025gravity}. The approach of \citet{zhang2025morpheus} is especially similar to our method, where they also use post-hoc image segmentation to track the state of objects in generated images, for the purpose of evaluating physical reasoning. Like prior work studying video models, we similarly evaluate pre-trained multimodal models with sequences of frames from videos, although importantly, we present a novel method which enables Image Generation Models to perform this task, which are not trained for video inputs or to generate video outputs. Additionally, Chain-of-Time uses a very small number of frames (4-5) for model inputs and outputs, compared to tens or hundreds of frames used with video models.

\section{Experiments}   \label{sec:experiment}


We hypothesize that by using Chain-of-Time, IGMs will be able to achieve better accuracy than when using direct prediction. We use the frames created by IGMs using \textit{Chain-of-Time} simulation to reveal the details about the simulation, including the initial states estimated by IGMs, physical interactions, and physical motion simulated by IGMs. To examine the overall validity and applicability of our method, we tested \textit{Chain-of-Time} on nine experimental domains: \textbf{\domainA}, \textbf{\domainB}, \textbf{\domainC}, \textbf{\domainD}, \textbf{\domainE}, \textbf{\domainF}, \textbf{\domainG}, \textbf{\domainH}, and \textbf{\domainI} (Fig.~\ref{fig:intro-domains}). We analyzed our results in two ways: first, we measured the accuracy of predicted images relative to ground truth; and second, we visualized the temporal dynamics in simulated images (such as x and y location) to assess physical motions produced by the IGM's simulations to see if it approximately follows the motion indicated by the governing physics equations, if the motion is governed by physics equations with distinctive shapes like $\sin$ function or quadratic functions.




\begin{figure*}[t!]
    \centering
    \includegraphics[width=\linewidth]{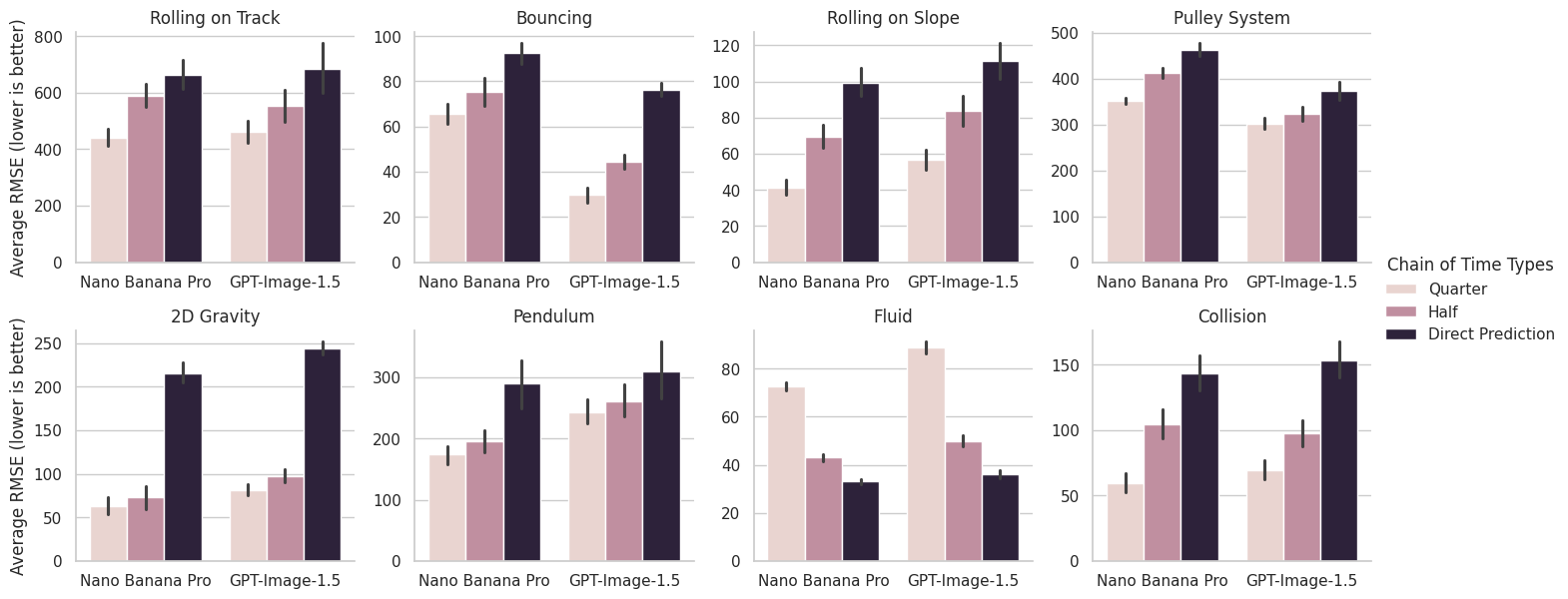}
    \caption{Prediction errors for the eight domains included in this analysis, averaged across all data for each domain. Prediction error is measured by taking the average RMSE between the ground-truth positions (location of focal object, or water level) and the final positions predicted by the IGM. Error bars are 95\% Bootstrapped CI. We generally find a monotonic relationship between \textit{Chain-of-Time} precision and performance when compared against the direct prediction baseline, indicating a more stable physical simulation under \textit{Chain-of-Time}. 
}
    \label{fig:mse}
\end{figure*}

\paragraph{Stimuli Design}
We used 9 different physical domains as our stimuli. The stimuli used in \domainA, \domainB were created in a simulation environment, and resemble stimuli in previous studies of intuitive physics \cite{smith2013sources}, \cite{bass2021partial}, \cite{gerstenberg2021counterfactual}. The stimuli used in \domainC were borrowed from \cite{wang2025resource}. In the \domainD, \domainE domains, we used real-world videos which have different physical parameters, e.g. different balls with different levels of ``bounciness''. In the \domainF, \domainG, \domainH, and \domainI, we borrowed them from physical reasoning benchmarks such as \cite{bear2021physion}, \cite{zheng2024contphy} and video websites like BiliBili and YouTube. We simulated different speeds by re-sampling input videos at different frame rates. For further details of how we designed stimuli, please refer to App.~\ref{app:stimuli}.

\paragraph{Image Generation Models}
We evaluated two state-of-the-art image generation models. The first image generation model we used is OpenAI's GPT Image 1.5 (gpt-image-1.5~\footnote{\texttt{openai.com/index/image-generation-api/}}), as of December, 2025. The second image generation model is Google DeepMind's Nano Banana Pro (gemini-3-pro-image-preview) ~\footnote{\texttt{https://ai.google.dev/gemini-api/docs\\/image-generation}}. Both models support multi-turn chat interactions and are state-of-the-art models in image generation. For derendering and simulation, the two models respectively use the VLMs of GPT 5.2 and Gemini 3 Pro.
Further, we also analyzed the newly released open-source IGM \texttt{HunyuanImage-3.0-Instruct}~\citep{cao2025hunyuanimage}. The experimental setup and results for this model are described in App.~\ref{app:open-source}.
In addition to these models, we also empirically tested other closed and open source image generation models including Dall-E 3 and Qwen Image, but we found that Dall-E 3 was unable to simulate images of future world states with any reasonable accuracy, and Qwen Image failed at multi-turn image generation due to the lack of a VLM as a simulator.
As an additional baseline, in App.~\ref{app:video-model} we also compare Chain-of-Time to physical simulation using a video generation model (\texttt{LTX-video}).

\begin{figure*}[t!]
    \centering
    \includegraphics[width=\linewidth]{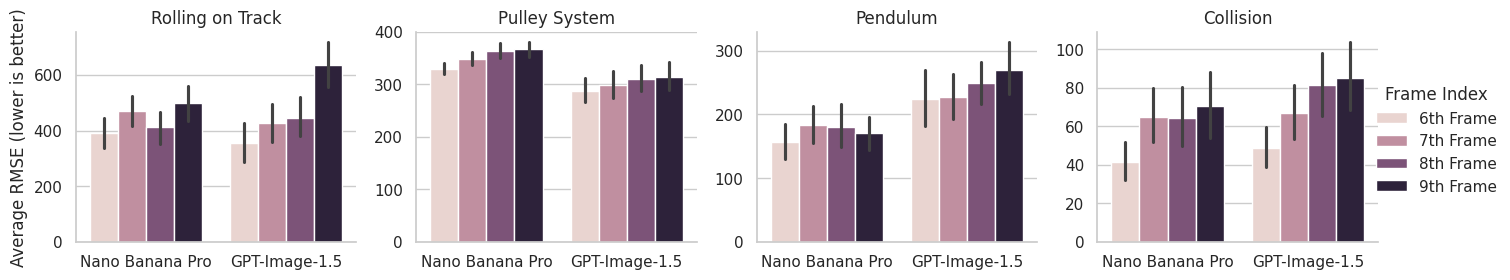}
    \caption{Prediction errors of each predicted frame for the four domains which have the most complicated object motion, averaged across all data for each domain under \textit{Chain-of-Time} Quarter. Prediction error is measured by taking the average RMSE between the ground-truth positions (location of focal object) and the positions predicted by the IGM. Error bars are 95\% Bootstrapped CI. We found that \textit{Chain-of-Time} Quarter up to 2 seconds of simulation shows small but consistent error accumulation across these domains with complex motion. 
}
    \label{fig:longhorizon}
\end{figure*}

\paragraph{Evaluating Simulation Accuracy}
Next we describe our methods for evaluating the quality of intermediate and final simulated images (also see App.~\ref{app:cv-algorithms}).
To track the specific positions of key objects, we used the Segment-Everything-Model 3 (SAM 3) ~\footnote{\texttt{https://ai.meta.com/sam3/}} to identify object positions. In each case, we specified text descriptions of the object that the SAM 3 model should segment (e.g. \textit{``tennis ball''}). For examples of successful segmentation generated by SAM 3 over the tasks we selected, see App.~\ref{sec:sam3examples}. The analysis in \domainC was conducted using a traditional CV algorithm due to consistent detection errors from SAM 3 on \domainC (further details in App.~\ref{app:fluid-cv-algo}). We used the default setting for SAM3 and all the evaluation pipeline code is available at: \url{https://anonymous.4open.science/r/Chain-of-time-8E8F/}. We conducted a evaluation based on sub-sampling to ensure the accuracy and stability of this evaluation pipeline, as detailed in App.~\ref{app:sam3_accuracy_check}.

\paragraph{Sampling Details}
At the start of each trial, the model was given 5 frames of a stimulus, showing the scene at 0, 0.5, 1.0, 1.5, and 2 seconds. Given these 5 frames, the model was asked to generate simulated frames at a time in the future, following the Initial Simulation Prompt. If \textit{Chain-of-Time} Quarter-Step or \textit{Chain-of-Time} Half-Step were used (see below), additional frames were generated following the Simulation Follow-Up Prompt. Details of the prompts we used can be found in App.~\ref{app:prompts}. In App.~\ref{app:prompts variation exp}, we also investigated the effect of varying the prompt complexity on the our method.



Chain-of-Time generates frames at different temporal resolutions, depending on a frame-rate parameter $s$. We considered two versions of \textit{Chain-of-Time} with $s$ = 2\ \text{steps} and $s$ = 4\ \text{steps}. Since the final frame was 2 seconds into the future, \textit{Chain-of-Time} with $s$ = 2\ \text{steps} generated 2 frames, corresponding to 1 and 2 seconds into the future, and\textit{Chain-of-Time} with $s$ = 4\ \text{steps} generated 4 frames, corresponding to 0.5, 1.0, 1.5, and 2.0 seconds into the future. For simplicity, we call \textit{Chain-of-Time} with $s= 4 \text{ steps}$ \textit{Chain-of-Time} Quarter, and \textit{Chain-of-Time} with $s = 2 \text{ steps}$ \textit{Chain-of-Time} Half. In addition, we had a baseline termed "Direct Prediction". In this method, we asked the model to directly generate the requested final frame, which is 2 seconds into the future. 
For all domains, we ran each stimulus 10 times (N=10) across all \textit{Chain-of-Time} simulations and Direct Prediction.

\section{Results} \label{sec:results}

\paragraph{Accuracy Analysis}


We evaluated the IGM's accuracy in predicting the ground-truth value of the target variable (object location or water height) using three methods: \textit{Chain-of-Time} Quarter, \textit{Chain-of-Time} Half, and Direct Prediction (Baseline), for both Nano Banana Pro and GPT Image 1.5. As shown in Fig.~\ref{fig:mse}, a finer \textit{Chain-of-Time} step generally improves accuracy in seven of the eight domains for both models. We excluded the 2D Motion domain from the analysis in Fig.~\ref{fig:mse} due to its simplicity, but found results similar to those in the other domains. Overall, these results suggest that \textit{Chain-of-Time} improves prediction accuracy across 2D and 3D tasks, spanning from simple to complex settings. However, performance differences between Nano Banana Pro and GPT Image 1.5 across domains indicate that gains over the Direct Prediction baseline also depend on the underlying capabilities of the base model.
In App.~\ref{app:new_baseline}, we compared Chain-of-Time to a stronger baseline which extends Direct Prediction to use self-consistency sample selection with CoT Reasoning, and Chain-of-Time is superior in accuracy to this baseline as well.
Chain-of-Time also consistently improves accuracy with the open-source IGM \texttt{HunyuanImage-3.0-Instruct} (except for the Fluids domain), and our results for this model broadly align with our results for the two closed-source IGMs (see App.~\ref{app:open-source} for further details).
Finally, we find that Chain-of-Time with these IGMs performs similarly to a state-of-the-art video generation model (App.~\ref{app:video-model}).

\paragraph{Physical Motion Analysis}


\begin{figure}
    \centering
    \includegraphics[width=.49\linewidth]{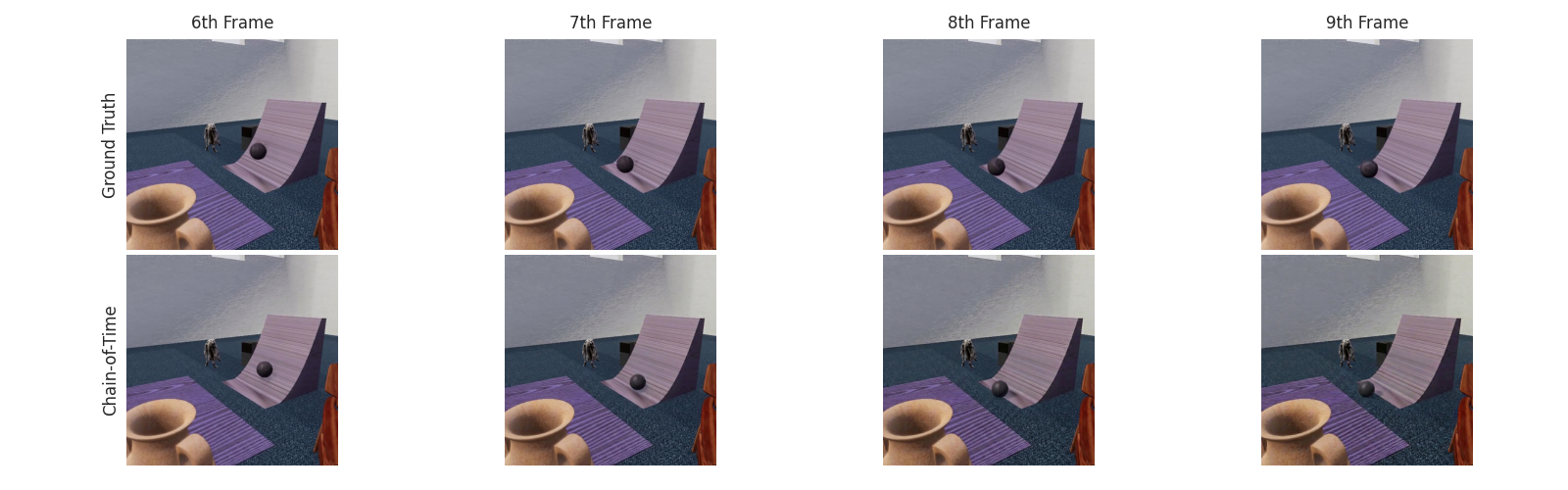}
    \includegraphics[width=.49\linewidth]{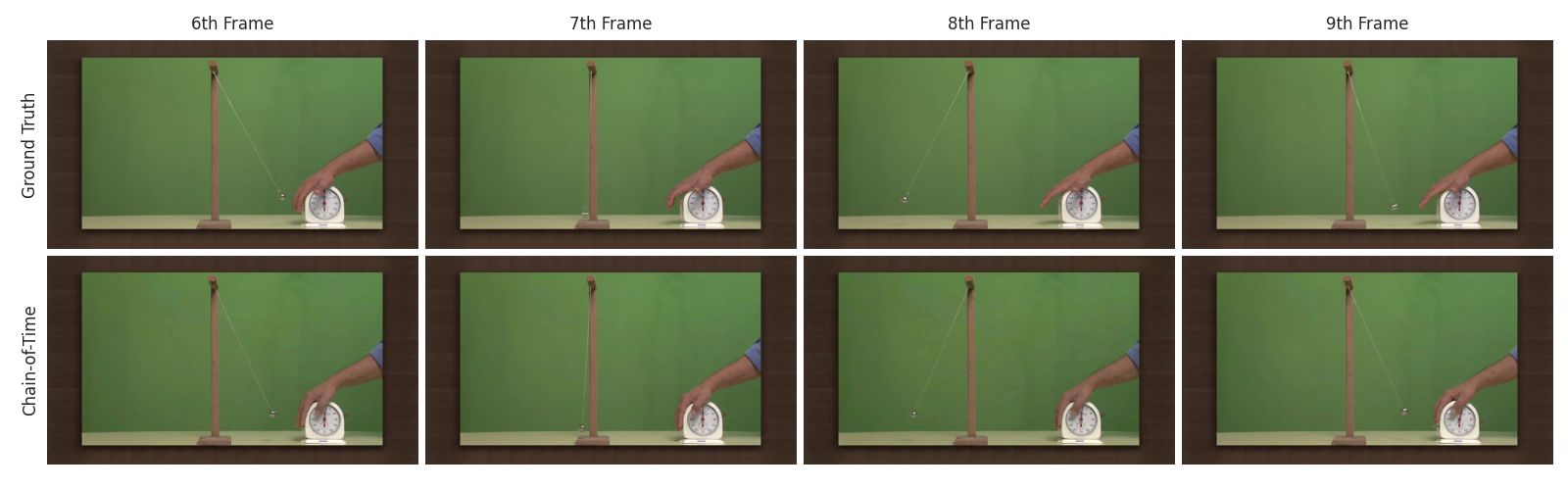}
    \
    \includegraphics[width=.45\linewidth]{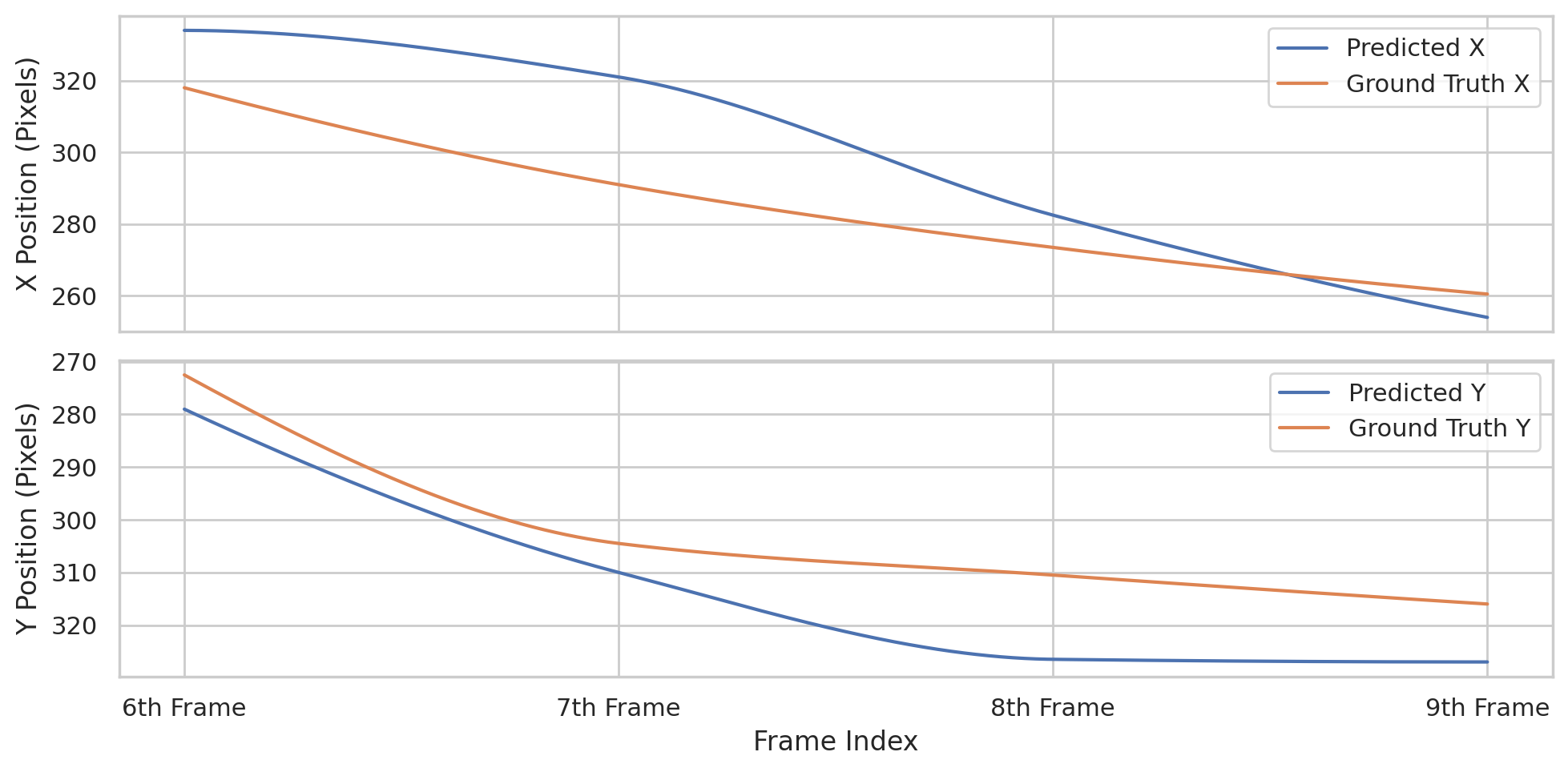}
    \hspace{20pt}
    \includegraphics[width=.45\linewidth]{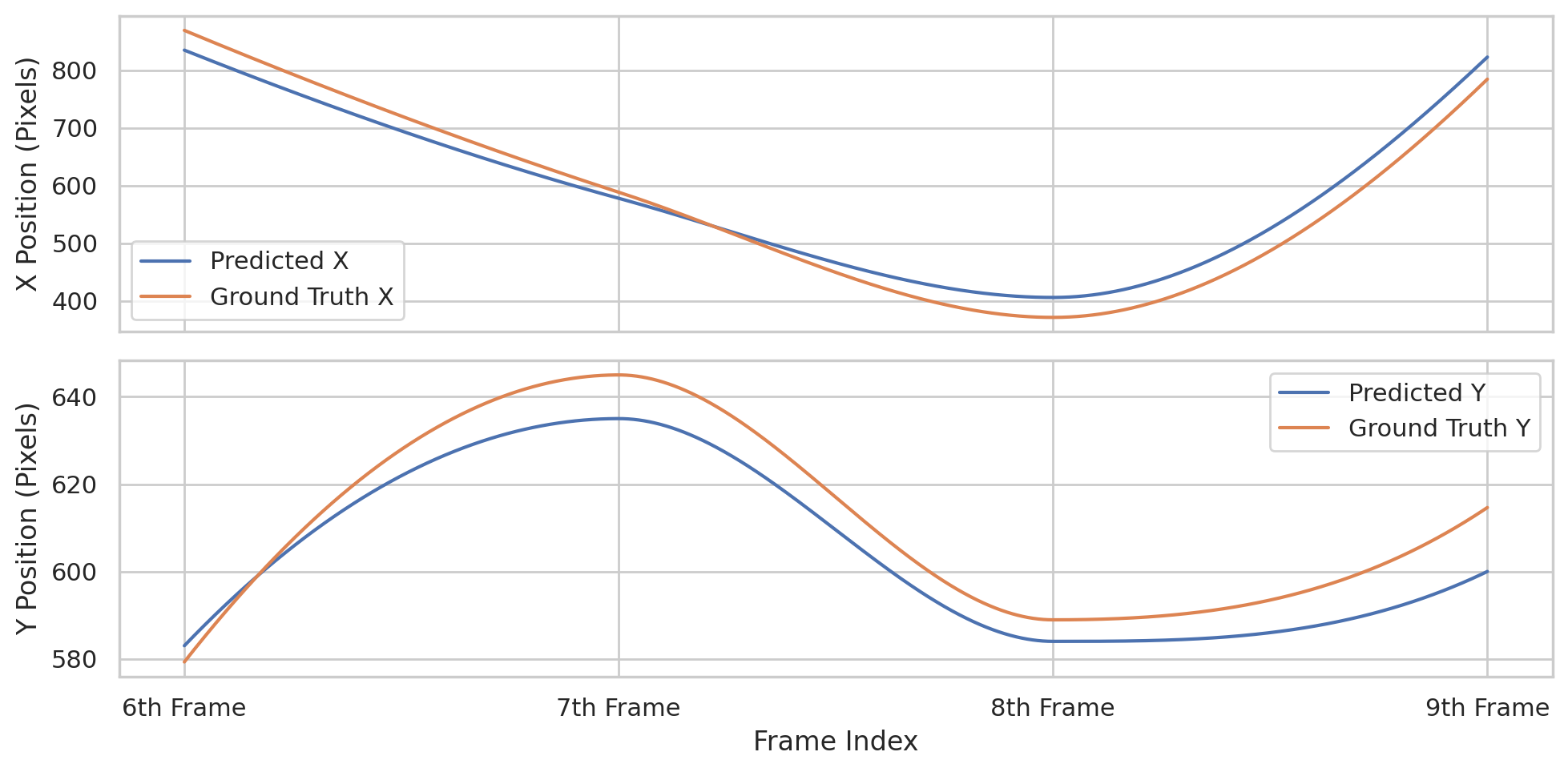}
    \caption{\textbf{Simulation analysis for \texttt{Rolling on Slope} and \texttt{Pendulum} domains} \ \  (Top) Comparison between ground truth images and images generated by Chain-of-Time (quarter) for a single trial. Frame indexing begins at 6 because the first 5 frames are provided as input to the model. 
    (Bottom Left) Extracted X and Y coordinates of the ball in each frame. The ground truth and simulated objects follow very similar trajectories, first accelerating sharply and then steadily decelerating.
    (Bottom Right) Extracted X and Y coordinates of the pendulum in each frame. We observe oscillatory motion in the pendulum for both the ground truth and simulated image sequences.
}
    \label{fig:inclined-pendulum}
\end{figure}

In our prior analysis, we found that \textit{Chain-of-Time} enables higher-fidelity physical simulation in most domains, measured by accuracy of the final generated image. This raises the question of precisely why Chain-of-Time improves the accuracy of physical simulation with IGMs. Similar to Chain-of-Thought reasoning with LLMs, Chain-of-Time simulation offers an interpretable trace of intermediate steps preceding the final output.
Therefore, we evaluate the IGM's ability to simulate physical motion by examining its intermediate simulated images, focusing on whether \textit{Chain-of-Time} stabilizes physical properties and produces object motions that are accurate and precise within the tested domains.

\begin{figure}
    \centering
    \includegraphics[width=.49\linewidth]{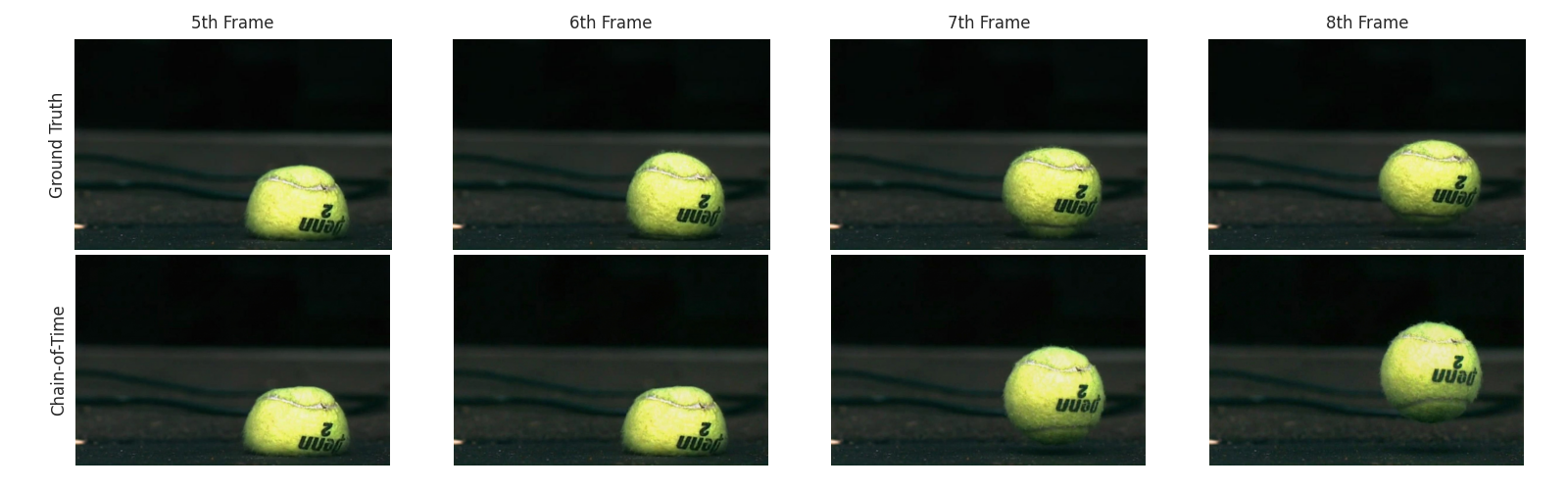}
    \includegraphics[width=.49\linewidth]{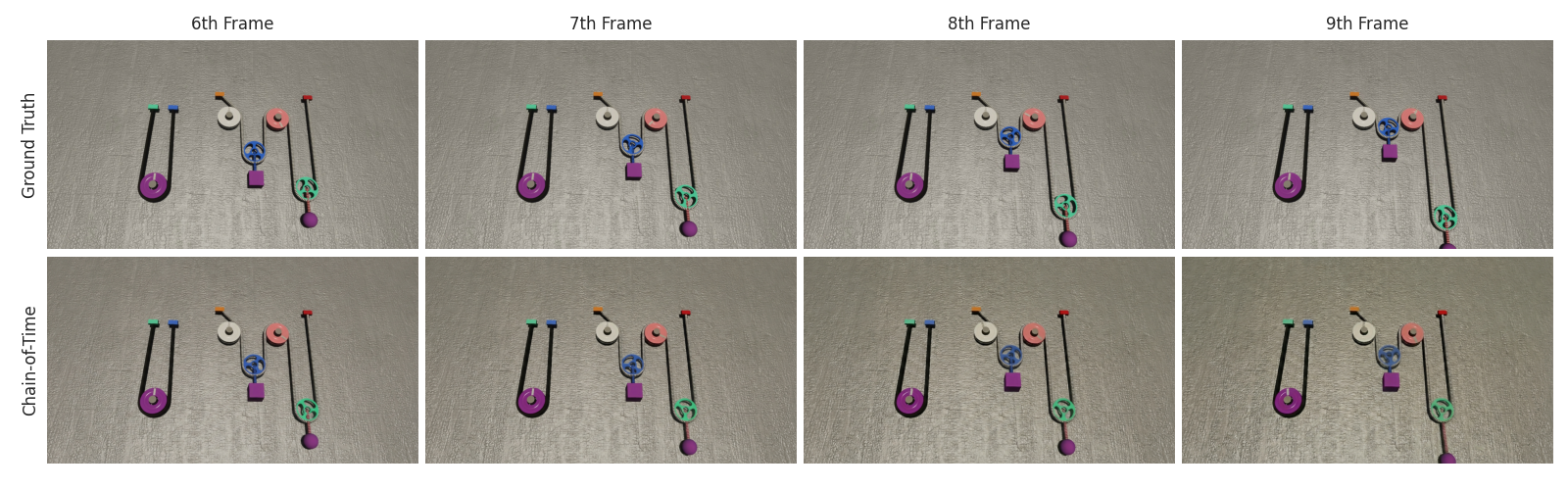}
    \\
    \includegraphics[width=.45\linewidth]{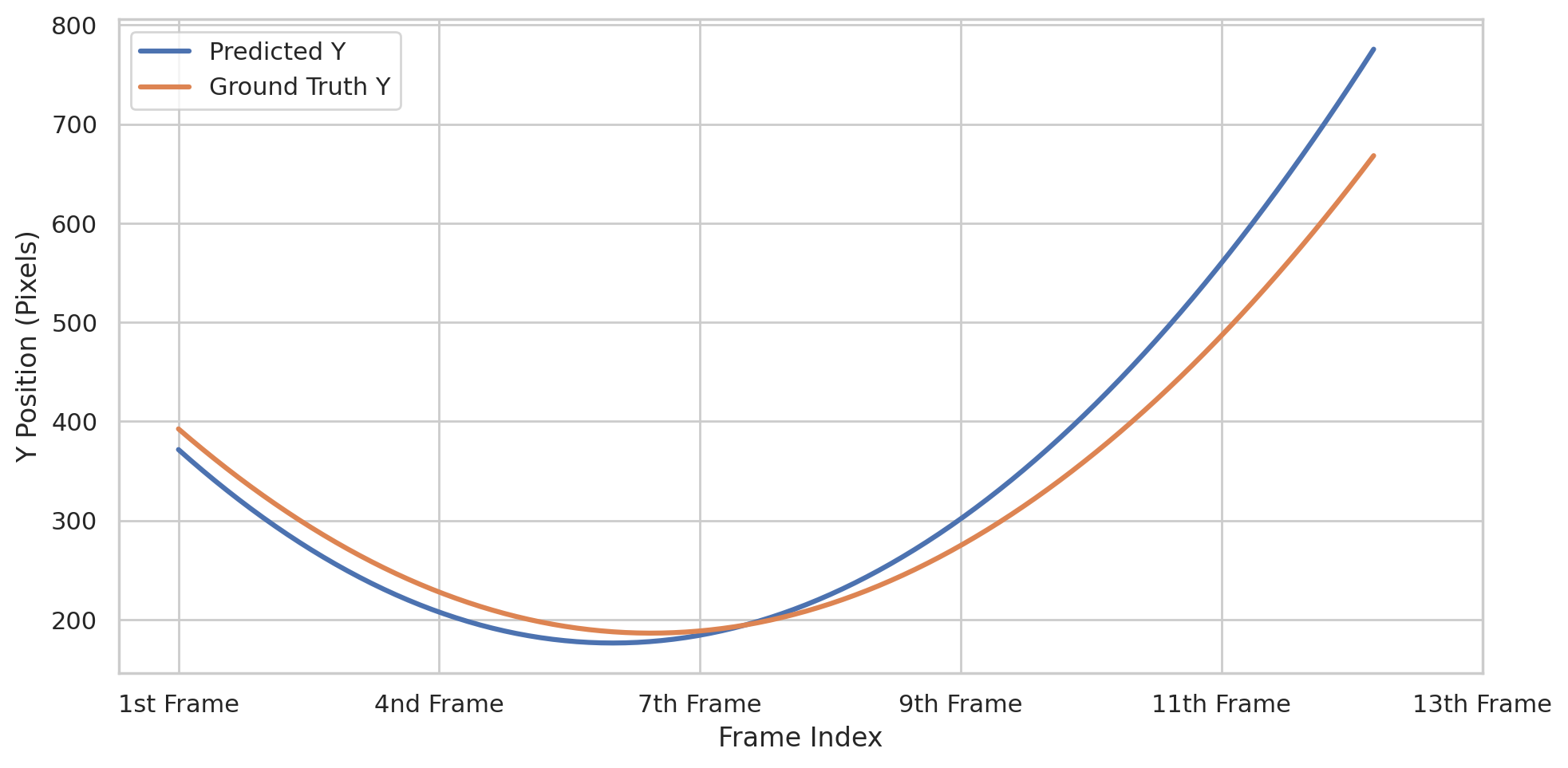}
    \hspace{20pt}
    \includegraphics[width=.45\linewidth]{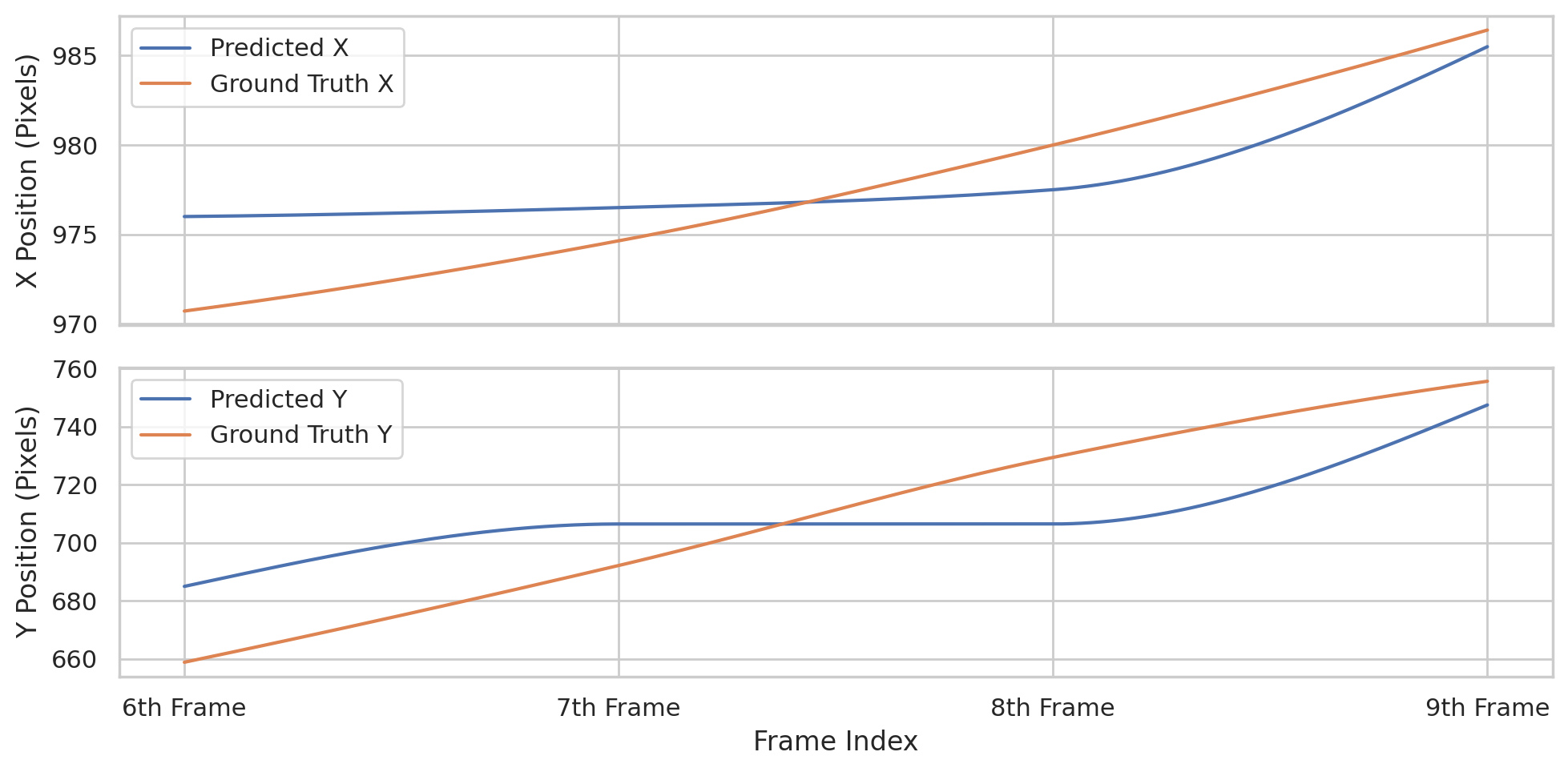}
    \caption{
    (Left) \textbf{Simulation analysis for \texttt{Bouncing} and \texttt{Pulley System} domains} \ \  (Top) Comparison between ground truth images and images generated by Chain-of-Time (quarter) for an example trial. For \texttt{Bouncing}, the images shown correspond to the ``during'' period of the collision, beginning at frame 5 of 12. (Bottom Left) Extracted Y coordinates of the tennis ball in each frame, for three independent simulation results of four frames each: one simulation before the bouncing collision, one during, and one after collision. We observe parabolic motion in the ball for both the ground truth and simulated image sequences. (Bottom Right) Extracted X and Y coordinates of the purple ball in each frame. Despite the complexity of this domain, image generation models are able to accurately simulate the ball's future motion.
}
    \label{fig:bouncing-pulley}
\end{figure}

We use \domainF, \domainD, \domainE, and \domainI as illustrative examples, since the governing equations in these domains have common functional forms with geometrically distinctive trajectories. Because \textit{Chain-of-Time} Quarter produces the greatest number of intermediate frames, we analyze the generated sequences from \textit{Chain-of-Time} Quarter in these domains. We present one representative trial per domain here. Additional examples for these and other domains are provided in App.~\ref{app:additional-analysis}.



As shown in Fig.~\ref{fig:inclined-pendulum}, the IGM-generated sequence (second row) closely matches the ground-truth motion in the reference video (first row). Because the ramp is curved, the ball's speed changes nonlinearly as it rolls downward, producing nonlinear trajectories in both the $x$ and $y$ coordinates, as illustrated in the lower panel of Fig.~\ref{fig:inclined-pendulum}.
In the \domainD domain, Fig.~\ref{fig:bouncing-pulley} shows that the IGM-generated motion likewise closely resembles the ground-truth sequence. We focus on the "during" interval, where the model must capture deformation during impact and the subsequent rebound, which requires modeling object deformation during impact and estimating physical parameters such as the coefficient of restitution. The IGM reproduces both the deformation dynamics and rebound behavior accurately, suggesting effective motion simulation and parameter inference from visual context.
A characteristic signature of bouncing motion is that the Y position follows a quadratic function over time. In the lower panel of Fig.~\ref{fig:bouncing-pulley}, we show that the IGM accurately recovers this functional form (upward U-shaped curve), further supporting its ability to simulate \domainD.




For the \domainE domain, Fig.~\ref{fig:inclined-pendulum} shows that the IGM generates realistic pendulum motion consistent with the physical dynamics of this domain. The coordinate traces in the lower panel of Fig.~\ref{fig:inclined-pendulum} exhibit the expected $\sin$ and $\cos$ dependence in $x$ and $y$, respectively matching the standard Cartesian parameterization of a simple pendulum.
In the \domainI domain, Fig.~\ref{fig:bouncing-pulley} indicates that the IGM captures the dynamics of a more complex pulley system involving both fixed and movable pulleys. The generated sequence closely matches the ground-truth video (first row), accurately reproducing the coupled motion of the two masses connected with two movable pulleys and through two fixed pulleys. These results demonstrate the IGM's capability to correctly understand the characteristics of fixed and movable pulleys, and in general the pulley system dynamics.

\paragraph{Do Simulations Degrade Over Time?}



With the prior analysis on accuracy and physical motion, we found that \textit{Chain-of-Time} substantially stabilizes the physical simulation and improves simulation accuracy in IGMs. Theoretically, we could run \textit{Chain-of-Time} for indefinitely long time periods; however, we also might expect that with very long time horizons small errors could eventually propagate into major discrepancies.
Similarly, previous work in cognitive science studying mental simulation in humans suggested that accuracy of simulation actually degrades over time due to uncertainty \cite{smith2013sources}. Therefore, we now aim to test whether \textit{Chain-of-Time} simulation's accuracy degrades, i.e. diverges from ground truth, the longer a simulation runs for.

We use \textit{Chain-of-Time} Quarter since it has the strongest overall performance (Fig.~\ref{fig:mse}) and provides the most intermediate frames for analysis.
We focus on four domains with complex dynamics and interactions: \domainF, \domainI, \domainE, and \domainG. As shown in Fig.~\ref{fig:longhorizon}, for both Nano Banana Pro and GPT-Image-1.5, the average RMSE of the 6th–9th generated frames (corresponding to approximately 0.5, 1.0, 1.5, and 2.0 seconds into the future) steadily increases with longer simulation, but at a slow rate. This demonstrates that \textit{Chain-of-Time} Quarter can maintain stable and robust simulations, although their accuracy may decrease with much longer time horizons.


In App.~\ref{app:add-experiments} we describe a number of additional experiments. First, in App.~\ref{app:add-num-input-frames} we vary the number of input frames and find diminishing returns where Chain-of-Time accuracy does not increase significantly with more than 5 input images. In App.~\ref{app:add-num-output-frames} we vary the number of intermediate simulation frames and similarly find diminishing effects with more frames, despite increasing costs. In App.~\ref{app:add-long-horizon} we find that IGMs become less accurate when predicting further into the future, similar to what has been found with human mental simulation, although Chain-of-Time still improves accuracy. Finally, in App.~\ref{app:add-vlm-as-critic} we show that Chain-of-Time simulations can be stabilized by using a VLM ``critic'' that verifies intermediate simulation steps and has the IGM regenerate low-accuracy images.

\section{Conclusion}

Inspired by mental simulation in humans and in-context reasoning in large language models, we presented a prompting method for step-by-step physical simulation in Image Generation Models. With this method, \textit{Chain-of-Time} simulation, a prediction horizon is divided into smaller time intervals, with mid-point frames being fed as input to the next step in an unfolding systematic process. We assessed \textit{Chain-of-Time} for differing degrees of temporal precision, across a variety of physical reasoning tasks, and using quantitative metrics of  step-wise accuracy compared to ground truth, and the qualitative metric of how well the IGM recovers physical parameters such as gravity and bounciness.

We also introduced a method for flexibly evaluating physical simulation accuracy in generated images by using a cutting-edge object segmentation algorithm (SAM 3). Evaluating physical simulation ability in IGMs presents an exciting new challenge for both AI evaluations and cognitive science: unlike for humans, where we must empirically probe mental simulations with secondary metrics like eye gaze, with IGMs we can directly analyze the images that the model generates.

Our results suggest that current Image Generation Models have some degree of physical simulation ability, but that ability may critically be limited in how far ahead into the future the IGM can simulate. This mirrors findings on the psychology of mental simulation in people, where people are much more accurate at predicting world states smaller amounts of time into the future, and mental simulation theoretically operates by chaining together sequences of short-term predictions. Analogously, \textit{Chain-of-Time} stabilizes the physical simulation in IGMs by chaining together sequences of short-term predictions. This is evidenced by \textit{Chain-of-Time} improving simulation accuracy for various domains involving physical reasoning. Another line of supporting evidence is our analysis of the intermediate images generated with \textit{Chain-of-Time}, which show that improved accuracy in the final state is preceded by more accurate and stable predictions of intermediate states. This also demonstrates a secondary advantage of \textit{Chain-of-Time}, which is that we can directly assess the simulated images produced by the IGM in order to evaluate the model's perception of physical parameters, motions, and interactions.

\newpage
\bibliography{refs}

@article{battaglia2013simulation,
  title={Simulation as an engine of physical scene understanding},
  author={Battaglia, Peter W and Hamrick, Jessica B and Tenenbaum, Joshua B},
  journal={Proceedings of the National Academy of Sciences},
  volume={110},
  number={45},
  pages={18327--18332},
  year={2013},
  publisher={National Academy of Sciences}
}

@article{ludwin2021limits,
  title={Limits on simulation approaches in intuitive physics},
  author={Ludwin-Peery, Ethan and Bramley, Neil R and Davis, Ernest and Gureckis, Todd M},
  journal={Cognitive Psychology},
  volume={127},
  pages={101396},
  year={2021},
  publisher={Elsevier}
}

@inproceedings{wu2017neural,
  title={Neural scene de-rendering},
  author={Wu, Jiajun and Tenenbaum, Joshua B and Kohli, Pushmeet},
  booktitle={Proceedings of the IEEE Conference on Computer Vision and Pattern Recognition},
  pages={699--707},
  year={2017}
}

@article{smith2019modeling,
  title={Modeling expectation violation in intuitive physics with coarse probabilistic object representations},
  author={Smith, Kevin and Mei, Lingjie and Yao, Shunyu and Wu, Jiajun and Spelke, Elizabeth and Tenenbaum, Josh and Ullman, Tomer},
  journal={Advances in neural information processing systems},
  volume={32},
  year={2019}
}

@article{wu2017learning,
  title={Learning to see physics via visual de-animation},
  author={Wu, Jiajun and Lu, Erika and Kohli, Pushmeet and Freeman, Bill and Tenenbaum, Josh},
  journal={Advances in neural information processing systems},
  volume={30},
  year={2017}
}

@article{sosa2025blending,
  title={Blending simulation and abstraction for physical reasoning},
  author={Sosa, Felix A and Gershman, Samuel J and Ullman, Tomer D},
  journal={Cognition},
  volume={254},
  pages={105995},
  year={2025},
  publisher={Elsevier}
}

@article{smith2023integrating,
  title={Integrating heuristic and simulation-based reasoning in intuitive physics},
  author={Smith, Kevin and Battaglia, Peter and Tenenbaum, Joshua},
  publisher={PsyArXiv}
}

@article{balaban2025physics,
  title={Physics versus graphics as an organizing dichotomy in cognition},
  author={Balaban, Halely and Ullman, Tomer D},
  journal={Trends in Cognitive Sciences},
  year={2025},
  publisher={Elsevier}
}

@misc{fischer2021building,
  title={The building blocks of intuitive physics in the mind and brain},
  author={Fischer, Jason},
  journal={Cognitive Neuropsychology},
  volume={38},
  number={7-8},
  pages={409--412},
  year={2021},
  publisher={Taylor \& Francis}
}

@article{gerstenberg2021counterfactual,
  title={A counterfactual simulation model of causation by omission},
  author={Gerstenberg, Tobias and Stephan, Simon},
  journal={Cognition},
  volume={216},
  pages={104842},
  year={2021},
  publisher={Elsevier}
}

@article{allen2021lifelong,
  title={Lifelong learning of cognitive strategies for physical problem-solving: the effect of embodied experience},
  author={Allen, Kelsey R and Smith, Kevin A and Bird, Laura-Ashleigh and Tenenbaum, Joshua B and Makin, Tamar R and Cowie, Dorothy},
  journal={bioRxiv},
  pages={2021--07},
  year={2021},
  publisher={Cold Spring Harbor Laboratory}
}

@article{fischer2016functional,
  title={Functional neuroanatomy of intuitive physical inference},
  author={Fischer, Jason and Mikhael, John G and Tenenbaum, Joshua B and Kanwisher, Nancy},
  journal={Proceedings of the national academy of sciences},
  volume={113},
  number={34},
  pages={E5072--E5081},
  year={2016},
  publisher={National Academy of Sciences}
}

@article{ullman2017mind,
  title={Mind games: Game engines as an architecture for intuitive physics},
  author={Ullman, Tomer D and Spelke, Elizabeth and Battaglia, Peter and Tenenbaum, Joshua B},
  journal={Trends in cognitive sciences},
  volume={21},
  number={9},
  pages={649--665},
  year={2017},
  publisher={Elsevier}
}

@article{smith2013sources,
  title={Sources of uncertainty in intuitive physics},
  author={Smith, Kevin A and Vul, Edward},
  journal={Topics in cognitive science},
  volume={5},
  number={1},
  pages={185--199},
  year={2013},
  publisher={Wiley Online Library}
}

@article{kojima2022large,
  title={Large language models are zero-shot reasoners},
  author={Kojima, Takeshi and Gu, Shixiang Shane and Reid, Machel and Matsuo, Yutaka and Iwasawa, Yusuke},
  journal={Advances in neural information processing systems},
  volume={35},
  pages={22199--22213},
  year={2022}
}

@article{chang2024survey,
  title={A survey on evaluation of large language models},
  author={Chang, Yupeng and Wang, Xu and Wang, Jindong and Wu, Yuan and Yang, Linyi and Zhu, Kaijie and Chen, Hao and Yi, Xiaoyuan and Wang, Cunxiang and Wang, Yidong and others},
  journal={ACM Transactions on Intelligent Systems and Technology},
  volume={15},
  number={3},
  pages={1--45},
  year={2024},
  publisher={ACM New York, NY}
}

@article{dang2024explainable,
  title={Explainable and interpretable multimodal large language models: A comprehensive survey},
  author={Dang, Yunkai and Huang, Kaichen and Huo, Jiahao and Yan, Yibo and Huang, Sirui and Liu, Dongrui and Gao, Mengxi and Zhang, Jie and Qian, Chen and Wang, Kun and others},
  journal={arXiv preprint arXiv:2412.02104},
  year={2024}
}

@article{xu2025visual,
  title={Visual Planning: Let's Think Only with Images},
  author={Xu, Yi and Li, Chengzu and Zhou, Han and Wan, Xingchen and Zhang, Caiqi and Korhonen, Anna and Vuli{\'c}, Ivan},
  journal={arXiv preprint arXiv:2505.11409},
  year={2025}
}

@article{chow2025physbench,
  title={PhysBench: Benchmarking and Enhancing Vision-Language Models for Physical World Understanding},
  author={Chow, Wei and Mao, Jiageng and Li, Boyi and Seita, Daniel and Guizilini, Vitor and Wang, Yue},
  journal={arXiv preprint arXiv:2501.16411},
  year={2025}
}

@inproceedings{gao2025vision,
  title={Do vision-language models have internal world models? towards an atomic evaluation},
  author={Gao, Qiyue and Pi, Xinyu and Liu, Kevin and Chen, Junrong and Yang, Ruolan and Huang, Xinqi and Fang, Xinyu and Sun, Lu and Kishore, Gautham and Ai, Bo and others},
  booktitle={ICLR 2025 Workshop on World Models: Understanding, Modelling and Scaling},
  year={2025}
}

@article{guo2025deepseek,
  title={Deepseek-r1: Incentivizing reasoning capability in llms via reinforcement learning},
  author={Guo, Daya and Yang, Dejian and Zhang, Haowei and Song, Junxiao and Zhang, Ruoyu and Xu, Runxin and Zhu, Qihao and Ma, Shirong and Wang, Peiyi and Bi, Xiao and others},
  journal={arXiv preprint arXiv:2501.12948},
  year={2025}
}

@article{jaech2024openai,
  title={Openai o1 system card},
  author={Jaech, Aaron and Kalai, Adam and Lerer, Adam and Richardson, Adam and El-Kishky, Ahmed and Low, Aiden and Helyar, Alec and Madry, Aleksander and Beutel, Alex and Carney, Alex and others},
  journal={arXiv preprint arXiv:2412.16720},
  year={2024}
}

@article{wang2022towards,
  title={Towards understanding chain-of-thought prompting: An empirical study of what matters},
  author={Wang, Boshi and Min, Sewon and Deng, Xiang and Shen, Jiaming and Wu, You and Zettlemoyer, Luke and Sun, Huan},
  journal={arXiv preprint arXiv:2212.10001},
  year={2022}
}

@article{turpin2023language,
  title={Language models don't always say what they think: Unfaithful explanations in chain-of-thought prompting},
  author={Turpin, Miles and Michael, Julian and Perez, Ethan and Bowman, Samuel},
  journal={Advances in Neural Information Processing Systems},
  volume={36},
  pages={74952--74965},
  year={2023}
}

@article{hu2024visual,
  title={Visual sketchpad: Sketching as a visual chain of thought for multimodal language models},
  author={Hu, Yushi and Shi, Weijia and Fu, Xingyu and Roth, Dan and Ostendorf, Mari and Zettlemoyer, Luke and Smith, Noah A and Krishna, Ranjay},
  journal={arXiv preprint arXiv:2406.09403},
  year={2024}
}

@article{hartshorne2025insights,
  title={Insights into cognitive mechanics from education, developmental psychology and cognitive science},
  author={Hartshorne, Joshua K and Jing, Mengguo},
  journal={Nature Reviews Psychology},
  pages={1--15},
  year={2025},
  publisher={Nature Publishing Group US New York}
}

@article{bass2021partial,
  title={Partial mental simulation explains fallacies in physical reasoning},
  author={Bass, Ilona and Smith, Kevin A and Bonawitz, Elizabeth and Ullman, Tomer D},
  journal={Cognitive Neuropsychology},
  volume={38},
  number={7-8},
  pages={413--424},
  year={2021},
  publisher={Taylor \& Francis}
}

@article{wang2025resource,
  title={Resource bounds on mental simulations: Evidence from a liquid-reasoning task.},
  author={Wang, YingQiao and Ullman, Tomer D},
  journal={Journal of Experimental Psychology: General},
  year={2025},
  publisher={American Psychological Association}
}

@article{goh2021multimodal,
  title={Multimodal neurons in artificial neural networks},
  author={Goh, Gabriel and Cammarata, Nick and Voss, Chelsea and Carter, Shan and Petrov, Michael and Schubert, Ludwig and Radford, Alec and Olah, Chris},
  journal={Distill},
  volume={6},
  number={3},
  pages={e30},
  year={2021}
}

@article{bhalla2024interpreting,
  title={Interpreting clip with sparse linear concept embeddings (splice)},
  author={Bhalla, Usha and Oesterling, Alex and Srinivas, Suraj and Calmon, Flavio and Lakkaraju, Himabindu},
  journal={Advances in Neural Information Processing Systems},
  volume={37},
  pages={84298--84328},
  year={2024}
}

@article{zhang2024vision,
  title={Vision-language models for vision tasks: A survey},
  author={Zhang, Jingyi and Huang, Jiaxing and Jin, Sheng and Lu, Shijian},
  journal={IEEE transactions on pattern analysis and machine intelligence},
  volume={46},
  number={8},
  pages={5625--5644},
  year={2024},
  publisher={IEEE}
}

@article{meng2024phybench,
  title={Phybench: A physical commonsense benchmark for evaluating text-to-image models},
  author={Meng, Fanqing and Shao, Wenqi and Luo, Lixin and Wang, Yahong and Chen, Yiran and Lu, Quanfeng and Yang, Yue and Yang, Tianshuo and Zhang, Kaipeng and Qiao, Yu and others},
  journal={arXiv preprint arXiv:2406.11802},
  year={2024}
}

@misc{openai_4o_image_generation_2025,
  author    = {OpenAI},
  title     = {Introducing 4o Image Generation},
  date      = {2025-03-25},
  url       = {https://openai.com/index/introducing-4o-image-generation/},
  urldate   = {2025-09-25},
  organization = {OpenAI},
    year = {2025}
}

@misc{openai_4o_addendum_2025,
  author    = {OpenAI},
  title     = {Addendum to GPT-4o System Card: Native image generation},
  date      = {2025-03-25},
  url       = {https://openai.com/index/gpt-4o-image-generation-system-card-addendum/},
  urldate   = {2025-09-25},
  organization = {OpenAI},
    year = {2025}
}

@misc{google_gem3_modelcard_2025,
  author    = {Google DeepMind},
  title     = {Addendum to GPT-4o System Card: Native image generation},
  date      = {2025-11-18},
  url       = {https://storage.googleapis.com/deepmind-media/Model-Cards/Gemini-3-Pro-Model-Card.pdf},
  organization = {Google DeepMind},
    year = {2025}
}

@article{cao2025hunyuanimage,
  title={Hunyuanimage 3.0 technical report},
  author={Cao, Siyu and Chen, Hangting and Chen, Peng and Cheng, Yiji and Cui, Yutao and Deng, Xinchi and Dong, Ying and Gong, Kipper and Gu, Tianpeng and Gu, Xiusen and others},
  journal={arXiv preprint arXiv:2509.23951},
  year={2025}
}

@misc{chen2025gpt4o_empirical,
  title         = {An Empirical Study of GPT-4o Image Generation Capabilities},
  author        = {Sixiang Chen and Jinbin Bai and Zhuoran Zhao and Tian Ye and Qingyu Shi and Donghao Zhou and Wenhao Chai and Xin Lin and Jianzong Wu and Chao Tang and Shilin Xu and Tao Zhang and Haobo Yuan and Yikang Zhou and Wei Chow and Linfeng Li and Xiangtai Li and Lei Zhu and Lu Qi},
  year          = {2025},
  month         = {apr},
  eprint        = {2504.05979},
  archivePrefix = {arXiv},
  primaryClass  = {cs.CV},
  doi           = {10.48550/arXiv.2504.05979},
  url           = {https://arxiv.org/abs/2504.05979},
  note          = {Version 2, revised 10 Apr 2025}
}

@article{cao2025text,
  title={Text-to-image diffusion models cannot count, and prompt refinement cannot help},
  author={Cao, Yuefan and Guo, Xuyang and Huo, Jiayan and Liang, Yingyu and Shi, Zhenmei and Song, Zhao and Zhang, Jiahao and Zhuang, Zhen},
  journal={arXiv preprint arXiv:2503.06884},
  year={2025}
}

@article{liu2023discovering,
  title={Discovering failure modes of text-guided diffusion models via adversarial search},
  author={Liu, Qihao and Kortylewski, Adam and Bai, Yutong and Bai, Song and Yuille, Alan},
  journal={arXiv preprint arXiv:2306.00974},
  year={2023}
}

@inproceedings{lu2024handrefiner,
  title={Handrefiner: Refining malformed hands in generated images by diffusion-based conditional inpainting},
  author={Lu, Wenquan and Xu, Yufei and Zhang, Jing and Wang, Chaoyue and Tao, Dacheng},
  booktitle={Proceedings of the 32nd ACM International Conference on Multimedia},
  pages={7085--7093},
  year={2024}
}

@article{merrill2023expressive,
  title={The expressive power of transformers with chain of thought},
  author={Merrill, William and Sabharwal, Ashish},
  journal={arXiv preprint arXiv:2310.07923},
  year={2023}
}

@article{prystawski2023think,
  title={Why think step by step? reasoning emerges from the locality of experience},
  author={Prystawski, Ben and Li, Michael and Goodman, Noah},
  journal={Advances in Neural Information Processing Systems},
  volume={36},
  pages={70926--70947},
  year={2023}
}

@article{bear2021physion,
  title={Physion: Evaluating physical prediction from vision in humans and machines},
  author={Bear, Daniel M and Wang, Elias and Mrowca, Damian and Binder, Felix J and Tung, Hsiao-Yu Fish and Pramod, RT and Holdaway, Cameron and Tao, Sirui and Smith, Kevin and Sun, Fan-Yun and others},
  journal={arXiv preprint arXiv:2106.08261},
  year={2021}
}

@article{zheng2024contphy,
  title={Contphy: Continuum physical concept learning and reasoning from videos},
  author={Zheng, Zhicheng and Yan, Xin and Chen, Zhenfang and Wang, Jingzhou and Lim, Qin Zhi Eddie and Tenenbaum, Joshua B and Gan, Chuang},
  journal={arXiv preprint arXiv:2402.06119},
  year={2024}
}

@article{motamed2025generative,
  title={Do generative video models understand physical principles?},
  author={Motamed, Saman and Culp, Laura and Swersky, Kevin and Jaini, Priyank and Geirhos, Robert},
  journal={arXiv preprint arXiv:2501.09038},
  year={2025}
}

@article{le2025gravity,
  title={What about gravity in video generation? Post-Training Newton's Laws with Verifiable Rewards},
  author={Le, Minh-Quan and Zhu, Yuanzhi and Kalogeiton, Vicky and Samaras, Dimitris},
  journal={arXiv preprint arXiv:2512.00425},
  year={2025}
}

@article{wei2025cophybench,
  title={CoPhyBench: Benchmarking Physical Reasoning from Conditional Video Observation},
  author={Wei, Fanyue and Xu, Kai and Zhang, Yizhuo and Sun, Pengzhan and Xiao, Junbin and Yao, Angela},
  journal={Openreview (Submitted to ICLR)},
  year={2025}
}

@article{yuan2025likephys,
  title={Likephys: Evaluating intuitive physics understanding in video diffusion models via likelihood preference},
  author={Yuan, Jianhao and Pizzati, Fabio and Pinto, Francesco and Kunze, Lars and Laptev, Ivan and Newman, Paul and Torr, Philip and De Martini, Daniele},
  journal={arXiv preprint arXiv:2510.11512},
  year={2025}
}

@article{wiedemer2025video,
  title={Video models are zero-shot learners and reasoners},
  author={Wiedemer, Thadd{\"a}us and Li, Yuxuan and Vicol, Paul and Gu, Shixiang Shane and Matarese, Nick and Swersky, Kevin and Kim, Been and Jaini, Priyank and Geirhos, Robert},
  journal={arXiv preprint arXiv:2509.20328},
  year={2025}
}

@article{zhang2025morpheus,
  title={Morpheus: Benchmarking physical reasoning of video generative models with real physical experiments},
  author={Zhang, Chenyu and Cherniavskii, Daniil and Tragoudaras, Antonios and Vozikis, Antonios and Nijdam, Thijmen and Prinzhorn, Derck WE and Bodracska, Mark and Sebe, Nicu and Zadaianchuk, Andrii and Gavves, Efstratios},
  journal={arXiv preprint arXiv:2504.02918},
  year={2025}
}

@article{bansal2024videophy,
  title={Videophy: Evaluating physical commonsense for video generation},
  author={Bansal, Hritik and Lin, Zongyu and Xie, Tianyi and Zong, Zeshun and Yarom, Michal and Bitton, Yonatan and Jiang, Chenfanfu and Sun, Yizhou and Chang, Kai-Wei and Grover, Aditya},
  journal={arXiv preprint arXiv:2406.03520},
  year={2024}
}

@misc{carion2025sam3segmentconcepts,
      title={SAM 3: Segment Anything with Concepts},
      author={Nicolas Carion and Laura Gustafson and Yuan-Ting Hu and Shoubhik Debnath and Ronghang Hu and Didac Suris and Chaitanya Ryali and Kalyan Vasudev Alwala and Haitham Khedr and Andrew Huang and Jie Lei and Tengyu Ma and Baishan Guo and Arpit Kalla and Markus Marks and Joseph Greer and Meng Wang and Peize Sun and Roman Rädle and Triantafyllos Afouras and Effrosyni Mavroudi and Katherine Xu and Tsung-Han Wu and Yu Zhou and Liliane Momeni and Rishi Hazra and Shuangrui Ding and Sagar Vaze and Francois Porcher and Feng Li and Siyuan Li and Aishwarya Kamath and Ho Kei Cheng and Piotr Dollár and Nikhila Ravi and Kate Saenko and Pengchuan Zhang and Christoph Feichtenhofer},
      year={2025},
      eprint={2511.16719},
      archivePrefix={arXiv},
      primaryClass={cs.CV},
      url={https://arxiv.org/abs/2511.16719},
}
\bibliographystyle{iclr2027_conference}

\newpage
\appendix

\section{Limitations}

The main limitation of our work is that we analyze only three Image Generation Models. This is because, as of submission time, there are only two AI model providers with fully featured autoregressive IGMs -- Google and OpenAI -- and we tested their most capable IGMs.
These companies have stated that IGMs are now their gold standard for text-to-image generation~\citep{openai_4o_addendum_2025, google_gem3_modelcard_2025}, although this technology is new and has not yet been adopted by other AI companies.
In App.~\ref{app:open-source}, we also provide analyses with a recently released open source IGM (HunyuanImage-3.0-Instruct). Since this model is currently under development, it does not yet support sequences of chat messages with images, although this is planned for the near future.

We recognize that the small number of available IGMs limits how well we can establish the generalization of our method.
However, we argue that this is similar to the state of research on LLMs when Chain-of-Thought reasoning was introduced. \citet{kojima2022large} showed Chain-of-Thought reasoning improved performance on only two LLMs. Although they tested open source LLMs, these models were significantly inferior to closed source LLMs and could not effectively perform zero-shot reasoning.
Zero-shot Chain-of-Thought reasoning has since been proven effective on a wide variety of LLMs, including much more powerful LLMs. Similarly, given our consistent success across all three IGMs, we expect that Chain-of-Time simulation will similarly improve physical simulation with more powerful IGMs in the future.

\newpage
\section{Stimuli Design and Specification of the Stimuli}  \label{app:stimuli}

\subsection{\domainA }
The stimulus shows a 2D red ball being launched at the left middle of the screen on a white background in a simulated environment. There is no gravity or friction enabled, and no visible boundaries on the white surface. The white surface is flat and featureless, allowing the red ball to travel without obstructions, as shown in Fig.~\ref{fig:intro-domains}. We varied the speed at which the ball rolled over the white surface with three different speeds: 100, 300, and 500 (in units of pixels/second). This gives us 3 stimuli in total. The stimuli and the generated image from the image generation model have the same resolution (1024*1024) natively.

\subsection{\domainB }
The stimulus shows a 2D red ball being launched at various positions (left-middle, left-bottom, right-middle, right-bottom) on a white background in a simulated environment. There is gravity enabled, and no visible boundaries on the white surface. The white surface is flat and featureless, allowing the red ball to travel without obstructions, as shown in Fig.~\ref{fig:intro-domains}. We varied the speed at which the ball rolled over the white surface with three different speeds: 230, 240, and 250 (in units of pixels/second) and with two different angles: $45^{\circ}$ and $60^{\circ}$. This gives us 24 stimuli in total. The stimuli and the generated image from the image generation model have the same resolution (1024*1024) natively.

\subsection{\domainC}
The stimulus shows a 3D glass mug being filled with water on a light sky-blue background in a simulated environment. The water is emitted from a dark-grey pipe into the glass mug, as shown in Fig.~\ref{fig:intro-domains}. There is gravity enabled, and the amount of water is limited so that the mug never overflows. We varied the flow rate: 25, 50, and 75 (in frames/second), the type of cups: small, medium, and large, and the ground truth water levels: 1/12 full, 3/12 full, 5/12 full, 7/12 full, 9/12 full. The stimuli are cropped to 1024*1024 resolution, keeping only the water pipe and the glass water mug. This will give us 45 stimuli in total. The generated images from the image generation model have the same resolution (1024*1024) natively.

\subsection{\domainD}
    The stimulus shows a real world 3D object being launched from the top and bouncing back after hitting the ground as shown in Fig.~\ref{fig:intro-domains}. There is gravity enabled, and the objects each have a specific coefficient of restitution. We varied the coefficient of restitution of the object by including various balls made with different materials, and the falling rate. The velocity we varied for all trials includes 10, 15, 20, and 50 (frames / second). We will also separate each stimulus into three partitions. The first partition is \textit{before}, which is the sequence of frames that shows the motion of the ball right before it hits the ground and starts to deform. The second partition is \textit{during}, which is the sequence of frames that shows the motion of the ball hitting the ground, deforming, and beginning to bounce back. The third partition is \textit{after}, which is the sequence of frames that shows the ball bouncing back upwards. This gives us 54 stimuli in total.

The stimulus images were resized to 1080*720, and the generated images from the image generation model are in 1536*1024, and were resized into 1080*720. 

\subsection{\domainE}
The stimulus shows a real world 3D simple pendulum oscillating between left and right as shown in Fig.~\ref{fig:intro-domains}. There is gravity enabled, and we vary the oscillation frequency of the simple pendulum in 10 and 15 (frames / second). We have 9 videos with different simple pendulums. This gives us 18 stimuli in total.

The stimulus images were resized to 1K resolution, with an aspect ratio of 16:9, 16:9, 4:5, 16:9, 2:3, 2:3, 16:9, 2:3, and 4:5.

\subsection{\domainF}
The stimulus from \cite{bear2021physion} shows a 3D object beginning to roll down from the top of a curved ramp to the bottom, as shown in Fig.~\ref{fig:intro-domains}. The scene includes synthetic environments with other objects, intended to mimic the real physical world. There is gravity enabled, and we vary the speed of the object rolling in 10 and 15 (frames / second). We have 9 videos with different 3D objects varying by color and shape. This gives us 18 stimuli in total.

The stimulus images had a resolution of $512\times512$ pixels, with an aspect ratio of 1:1 for all the videos.

\subsection{\domainG}
The stimulus from \cite{bear2021physion} shows a 3D object colliding with another object, as shown in Fig.~\ref{fig:intro-domains}. The scene includes synthetic environments with other objects, intended to mimic the real physical world. There is gravity enabled, and we vary the speed of the object rolling in 10 and 15 (frames / second). We have 9 videos with different 3D objects varying by color and shape. This gives us 18 stimuli in total.

The resolution of the stimuli were 512*512 resolution, with an aspect ratio of 1:1 for all the videos.

\subsection{\domainH}
The stimulus from YouTube shows various 3D objects rolling in complex tracks set up in simulated environments. The scene is extremely complex and realistic, with many realistic physics assets in order to closely resemble real-world scenes. We vary the speed of the object rolling in 10 and 15 (frames / second). We have 7 videos with different 3D objects varying by color and shape. This gives us 14 stimuli in total.

The stimulus images were resized to 1920*1080 resolution, with an aspect ratio of 9:16, 9:16, 9:16, 9:16, 9:16, 16:9, and 16:9.

\subsection{\domainI}
The stimulus from \cite{zheng2024contphy} shows a complex pulley system consisting of multiple pulley systems in a simulated environment. Each pulley system consists of multiple static or movable pulleys, and a complex background to mimic the common real world complex pulley system. The gravity is enabled, and we vary the speed of the object moving on the pulley in 10 and 15 (frames / second). We have 13 videos with different 3D objects varying by color and shape. This gives us 26 stimuli in total.

The stimulus images were resized to 1920*1080 resolution, with an aspect ratio of 16:9 for all videos.

\clearpage
\section{Prompts}
\label{app:prompts}

We defined two parameters for the prompts we used in all nine domains.

For parameter \textbf{number of seconds forward}, which is $k$, we chose from $k \in [0.5, 1.0, 2.0]$. 

For parameter \textbf{scene content}, we choose it based on the domain. In \domainA, we define the parameter as the following: \textit{``a ball rolling on a surface''}. In \domainB, we define the parameter as the following: \textit{``a ball being launched from a specific position and angle''}. In \domainC, we define the parameter as the following: \textit{``a glass mug being filled with water, at a constant rate''}. For \domainD, when simulating \textit{before} and \textit{during} partitions, we define the parameters as the following: \textit{``a bouncy ball falling towards the ground''}. When simulating \textit{after} partition, we define the parameters as the following: \textit{``a bouncy ball bouncing upward after hitting the ground''}. In \domainE, we define the parameter as the following: \textit{``a bob swinging back-and-forth in simple pendulum motion''}. In \domainF, we define the parameter as the following: \textit{``an object rolling down an inclined ramp''}. In \domainG, we define the parameter as the following: \textit{``a object moving towards and then colliding with another object that is lighter (in absolute mass).''}. In \domainH, we define the parameter as the following: \textit{``a ball rolling inside a track''}. In \domainI, we define the parameter as the following: \textit{``a running complex pulley system''}.

\subsection{Prompts for all domains}

In all domains, for our \textit{Chain-of-Time} simulation method, as well as our Direct Prediction baseline, models are provided the following prompt, with different methods (\textit{Chain-of-Time} Quarter, Half, and Direct Prediction) varying the \texttt{\{\{scene content\}\}} and \texttt{\{\{number of seconds forward\}\}} parameter:

\begin{tcolorbox}[colframe=teal, title=Simulation Instruction Prompt]
\begin{lstlisting}
Consider the following sequence of 5 images, which show {scene_content}. Each image frame video is precisely 0.5 seconds after the last frame.

Please generate an image that continues this sequence, simulating what the scene will look like {number of seconds forward} seconds further into the future after the last frame. 

{{image sequence}}
\end{lstlisting}

\end{tcolorbox}

For \textit{Chain-of-Time}, we use the following prompt to elicit subsequent simulation steps from the IGM:

\begin{tcolorbox}[colframe=brown, title=Simulation Follow-Up Prompt]

\begin{lstlisting}
Continue this simulation {number of seconds forward} seconds into the future after the last frame that you generated. 


\end{lstlisting}
\end{tcolorbox}

\subsection{Additional Experiment: Varying Prompts for all domains}

In addition to the prompt we provided above, we also probed the effect of prompt complexity on our method. In this part, we conducted an additional experiment with the aim of exploring the prompt sensitivity of our method. 

To investigate the method's sensitivity to the prompt, we designed another version of the prompt that provide more information than the prompt we listed in section \ref{app:prompts}. The prompts below were designed to intentionally include more details about the physical simulation task itself, more detailed simulation instructions, and more details on the generated images themselves. These prompts therefore contrast with our simple general prompt in App.\ref{app:prompts}.

For prompts, we designed three parameters for the prompts, and we used them in all nine domains, if necessary.

For parameter \textbf{number of seconds forward}, which is $k$, we chose from $k \in [0.5, 1.0, 1.5, 2.0]$.

For parameter \textbf{direction}, we choose from leftbottom, leftmiddle,
rightmiddle, and rightbottom.

For parameter \textbf{scene content}, we choose it based on the domain.

In \domainC, we define the parameter as the following:
\textit{``a glass mug being filled with water, at a constant rate''}.

For \domainD, when simulating \textit{before} and \textit{during}
partitions, we define the parameters as the following:
\textit{``a bouncy ball falling towards the ground''}.
When simulating \textit{after} partition, we define the parameters as the
following:
\textit{``a bouncy ball bouncing upward after hitting the ground''}.

In \domainE, we define the parameter as the following:
\textit{``a bob swinging back-and-forth in simple pendulum motion''}.

In \domainF, we define the parameter as the following:
\textit{``an object rolling down a inclined ramp''}.

In \domainG, we define the parameter as the following:
\textit{``a object moving towards and then colliding with another object that
is lighter (in absolute mass)''}.

In \domainH, we define the parameter as the following:
\textit{``a ball rolling inside a track''}.

In \domainI, we define the parameter as the following:
\textit{``a running complex pulley system''}.


\subsection{Prompt for \texttt{2D Motion}}

In the \texttt{2D Motion} domain, for our \textit{Chain-of-Time} simulation method,
as well as our Direct Prediction baseline, models are provided the following
prompt, with different methods (\textit{Chain-of-Time} Quarter, Half, and
Direct Prediction) varying the \texttt{number of seconds forward} parameter:

\begin{tcolorbox}[colframe=teal, title=Simulation Instruction Prompt]
\begin{lstlisting}
Consider the following 5 frames, which show the motion of a red ball on a white
background. Note that each frame is precisely .5 seconds apart.

Now, please generate an image that simulates what this scene would look like
{number of seconds forward} Seconds into the future.

Make sure that your image is 2d and consists of a single red circle on a solid
white background. Ensure that the circle is exactly the same size as the input
images. Assume that there is no friction, the ground is flat, and the ball can
pass through objects.

{{image sequence}}
\end{lstlisting}
\end{tcolorbox}

For \textit{Chain-of-Time}, we used the following prompt to elicit subsequent
simulation steps from the IGM:

\begin{tcolorbox}[colframe=brown, title=Simulation Follow-Up Prompt]
\begin{lstlisting}
Now, simulate additional {number of seconds forward} seconds into the future.
\end{lstlisting}
\end{tcolorbox}


\subsection{Prompts for \texttt{2D Gravity}}

In the \texttt{2D Gravity} domain, for our \textit{Chain-of-Time} simulation
method, as well as our Direct Prediction baseline, models are provided the
following prompt, with different methods (\textit{Chain-of-Time} Quarter,
Half, and Direct Prediction) varying the \param{direction} and
\param{number of seconds forward} parameters:

\begin{tcolorbox}[colframe=teal, title=Simulation Instruction Prompt]
\begin{lstlisting}
Consider the following 5 frames, which show the projectile motion of a red ball
being launched from the {direction}. Each frame is precisely 0.5 seconds apart.

You will generate an image that simulates the position of the red ball
{number of seconds forward} seconds forward into the future after the last
frame.

Assume that there is gravity and the red ball continues to follow the projectile
motion shown in the frames provided. The scene is viewed from the side, so
gravity pulls downward.

Make sure that the generated image is 2-D, that it contains a circle exactly the
same size and color as the circle in the frames provided, and that the
background color is white.

{{image sequence}}
\end{lstlisting}
\end{tcolorbox}

For \textit{Chain-of-Time}, we use the following prompt to elicit subsequent
simulation steps from the IGM:

\begin{tcolorbox}[colframe=brown, title=Simulation Follow-Up Prompt]
\begin{lstlisting}
Generate an additional image that simulates the position of the red ball
{number of seconds forward} seconds forward into the future after the last
frame that you generated.

Assume that there is gravity and the red ball continues to follow the projectile
motion shown in the frames provided. The scene is viewed from the side, so
gravity pulls downward.

Make sure that the generated image is 2-D, that it contains a circle exactly the
same size and color as the circle in the frames provided, and that the
background color is white.
\end{lstlisting}
\end{tcolorbox}


\subsection{Prompts for \texttt{Fluids} and \texttt{Bouncing}}

In domains \texttt{Fluids} and \texttt{Bouncing}, for our
\textit{Chain-of-Time} simulation method, as well as our Direct Prediction
baseline, models are provided the following prompt, with different methods
(\textit{Chain-of-Time} Quarter, Half, and Direct Prediction) varying the
\param{scene content} and \param{number of seconds forward} parameters:

\begin{tcolorbox}[colframe=teal, title=Simulation Instruction Prompt]
\begin{lstlisting}
Consider the following sequence of 5 images, which show {scene_content}.
Each image frame video is precisely 0.5 seconds after the last frame.

Please generate an image that continues this sequence, simulating what the scene
will look like {number of seconds forward} seconds further into the future
after the last frame. Make sure that the generated image is from exactly the
same perspective as the input images and that the background color remains the
same color.

{{image sequence}}
\end{lstlisting}
\end{tcolorbox}

For \textit{Chain-of-Time}, we use the following prompt to elicit subsequent
simulation steps from the IGM:

\begin{tcolorbox}[colframe=brown, title=Simulation Follow-Up Prompt]
\begin{lstlisting}
Continue simulating this scene {number of seconds forward} seconds into the
future after the last frame that you generated. Make sure that the generated
image is from exactly the same perspective as the input images and that the
background color remains the same color.
\end{lstlisting}
\end{tcolorbox}


\subsection{Prompts for all other domains}

In all other domains, for our \textit{Chain-of-Time} simulation method, as well
as our Direct Prediction baseline, models are provided the following prompt,
with different methods (\textit{Chain-of-Time} Quarter, Half, and Direct
Prediction) varying the \param{scene content} and
\param{number of seconds forward} parameters:

\begin{tcolorbox}[colframe=teal, title=Simulation Instruction Prompt]
\begin{lstlisting}
Consider the following sequence of 5 images, which show {scene_content}.
Each image frame video is precisely 0.5 seconds after the last frame.

Please generate an image that continues this sequence, simulating what the scene
will look like {number of seconds forward} seconds further into the future
after the last frame. Make sure that the generated image is from exactly the
same perspective as the input images and that the background color remains the
same color. In addition, make sure the object moving in the images you generated
follows the same motion shown in the images provided.

{{image sequence}}
\end{lstlisting}
\end{tcolorbox}

For \textit{Chain-of-Time}, we use the following prompt to elicit subsequent
simulation steps from the IGM:

\begin{tcolorbox}[colframe=brown, title=Simulation Follow-Up Prompt]
\begin{lstlisting}
Continue this simulation {number of seconds forward} seconds into the future
after the last frame that you generated. Make sure that the generated image is
from exactly the same perspective as the input images, the background color
remains the same color, and the object moving in the image you generated follows
the same motion shown in the images provided.
\end{lstlisting}
\end{tcolorbox}

Here are the experimental results obtained using the prompts listed in App.\ref{app:prompts variation exp}:

\begin{figure*}[t!]
    \centering
    \includegraphics[width=\linewidth]{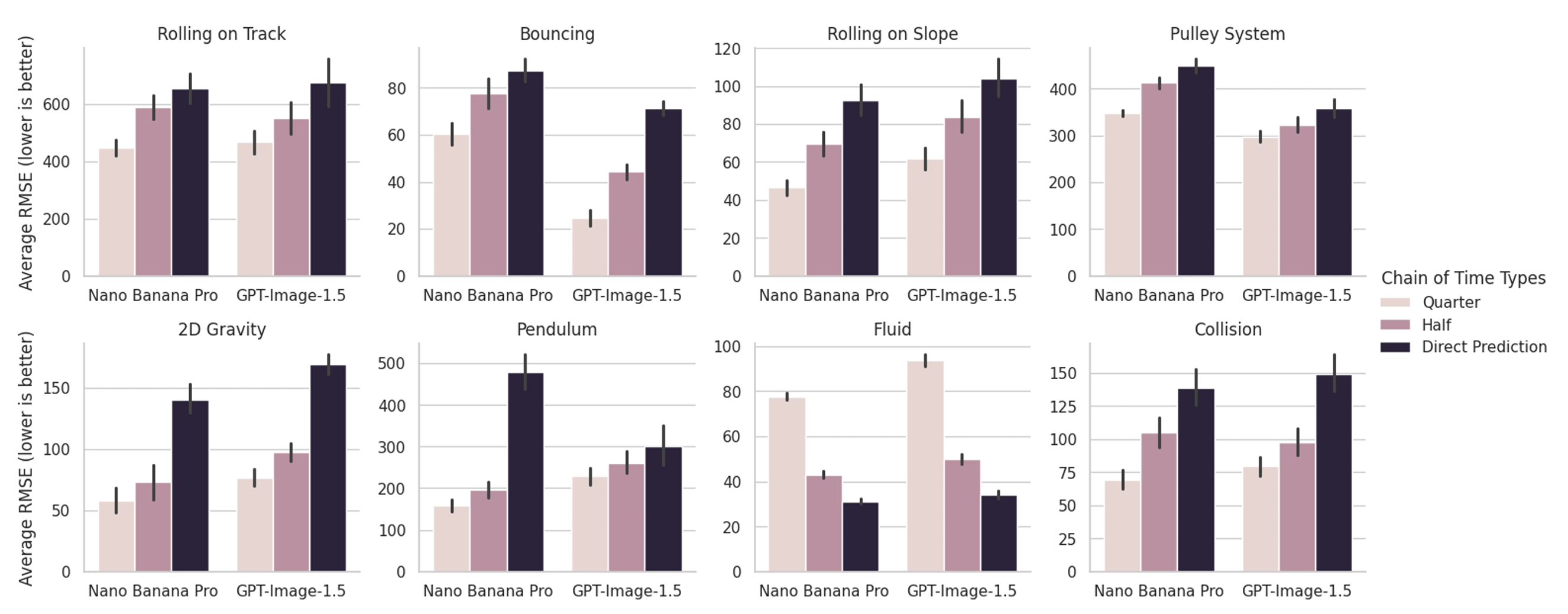}
    \caption{Prediction errors for the eight domains included in this analysis, averaged across all data for each domain using prompts listed in App.~\ref{app:prompts variation exp}. Prediction error is measured by taking the average RMSE between the ground-truth positions (location of focal object, or water level) and the final positions predicted by the IGM. Error bars are 95\% Bootstrapped CI. We observe similar prediction between the results generated by the prompts in App.~\ref{app:prompts variation exp} and the results generated by the prompts in App.~\ref{app:prompts}, which is shown in Fig.~\ref{fig:mse}. 
}
\end{figure*}

\begin{figure*}[t!]
    \centering
    \includegraphics[width=\linewidth]{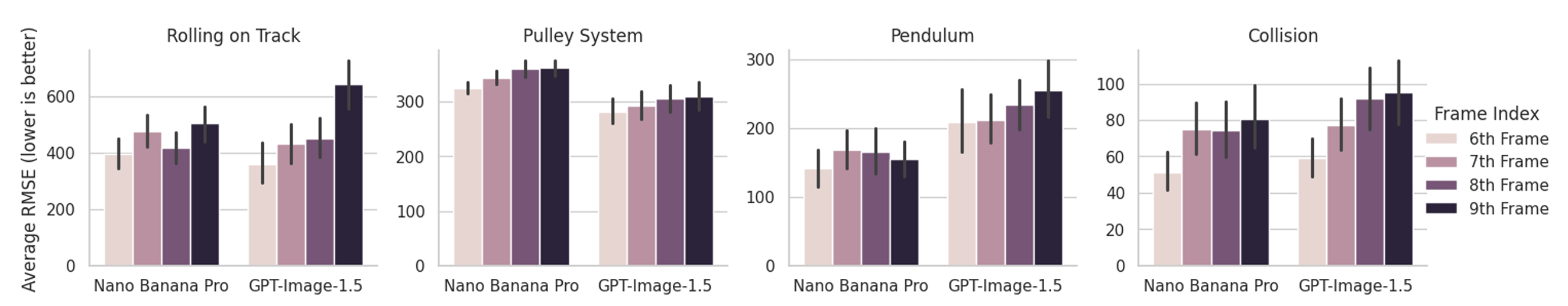}
    \caption{Prediction errors of each predicted frame for the four domains which have the most complicated object motion, averaged across all data for each domain under \textit{Chain-of-Time} Quarter, using the data generated by App.~\ref{app:prompts variation exp}. Prediction error is measured by taking the average RMSE between the ground-truth positions (location of focal object) and the positions predicted by the IGM. Error bars are 95\% Bootstrapped CI. We found that the result generated by using prompt listed in App.~\ref{app:prompts variation exp} shows similar quantitative values and patterns when compared with the result generated by using prompt listed in App.~\ref{app:prompts}, which is shown in Fig.~\ref{fig:longhorizon}.
}
\end{figure*}

\label{app:prompts variation exp}

\clearpage

\section{Details of Object Detection Algorithms}
\label{app:cv-algorithms}

\subsection{SAM3 Details}

In our analysis, we utilized Segment Everything Model 3\cite{carion2025sam3segmentconcepts} to track the objects of interest in all domains except \domainC. The prompts that we used to extract the objects of interest are listed below by domain.

\subsubsection{\domainA,\domainB}

\begin{tcolorbox}[colframe=blue, title=Prompt for Object Detection in \domainA and \domainB]

A Red Ball

\end{tcolorbox}

\subsubsection{\domainD}

\begin{tcolorbox}[colframe=blue, title=Prompt for Object Detection in \domainD]

\begin{enumerate}
    \item A transparent ball
    \item A ball with smiley face
    \item A ball with Thomas the Tank Engine
    \item A black ball
    \item A tennis ball
    \item A soccer ball
    \item A tennis ball
    \item A purple ball
    \item A tennis ball
\end{enumerate}
\end{tcolorbox}

\subsubsection{\domainE}

\begin{tcolorbox}[colframe=blue, title=Prompt for Object Detection in \domainE]

\begin{enumerate}
    \item A Silver Ball
    \item A Silver Ball
    \item A Silver Object
    \item A Silver Ball
    \item A Blue Ball
    \item A Brass Ball
    \item A Silver Ball
    \item A Brass Ball
    \item A Silver Object
\end{enumerate}
\end{tcolorbox}

\subsubsection{\domainF}

\begin{tcolorbox}[colframe=blue, title=Prompt for Object Detection in \domainF]

\begin{enumerate}
    \item A Brown Ball
    \item A Brown Object
    \item A Brown Ball
    \item A Dark Purple Ball
    \item An Orange Object
    \item A Dark Purple Ball
    \item A Green Cube
    \item A Dark Green Cube
    \item A Dark Green Object
\end{enumerate}
\end{tcolorbox}

\subsubsection{\domainG}

\begin{tcolorbox}[colframe=blue, title=Prompt for Object Detection in \domainG]

\begin{enumerate}
    \item A Black Object
    \item A Green Object
    \item A Green Ball
    \item A Green Ball 
    \item A Green Ball
    \item A Green Cube
    \item A Green Ball
    \item A Green Cylinder 
    \item A Green Object
    \item A Green Ball
\end{enumerate}
\end{tcolorbox}

\subsubsection{\domainH}

\begin{tcolorbox}[colframe=blue, title=Prompt for Object Detection in \domainH]

\begin{enumerate}
    \item An Orange Ball
    \item A Silver Ball
    \item A Silver Ball
    \item A Red Ball
    \item A Silver Ball 
    \item A Silver Ball
    \item A Silver Ball
\end{enumerate}
\end{tcolorbox}

\subsubsection{\domainI}

\begin{tcolorbox}[colframe=blue, title=Prompt for Object Detection in \domainI]

\begin{enumerate}
    \item A White Cube
    \item A Yellow Cube 
    \item A Yellow Ball 
    \item An Orange Cube 
    \item A Purple Ball 
    \item A Blue Ball 
    \item A Purple Ball 
    \item A Cyan Ball 
    \item A White Cube 
    \item A White Ball 
    \item A Grey Ball
    \item A Flesh-Color Ball 
    \item A Red Ball
\end{enumerate}
\end{tcolorbox}

\subsection{SAM3 Accuracy Check}
\label{app:sam3_accuracy_check}

To ensure the stability and the accuracy of our evaluation pipeline, we investigated the robustness of the segmentation masks generated by SAM3 by conducting a mask evaluation based on subsampling. We sub-sampled 3 generated frames for each stimuli in each subdomain, and had a human evaluator to evaluate the correctness of SAM3 segmentation masks. We found that SAM3 received the maximum possible score from the human evaluator in each domain, except for fluids; for the fluids domain, our human evaluator assigned the maximum score to the masks generated by our hand-engineered algorithm. To demonstrate segmentation mask validity for the reader, we included a sample of these segmentation masks generated by SAM3 and fluid detector in App.~\ref{sec:sam3examples}. While we believe that a more ideal validation would compare SAM3 masks to ground truth masks generated by humans, this process would be very resource intensive and beyond the scope of our present work.

\subsection{Examples of SAM3 Segmentation}
\label{sec:sam3examples}

\begin{figure}[b]
    \centering
    \fbox{\includegraphics[width=0.5\linewidth]{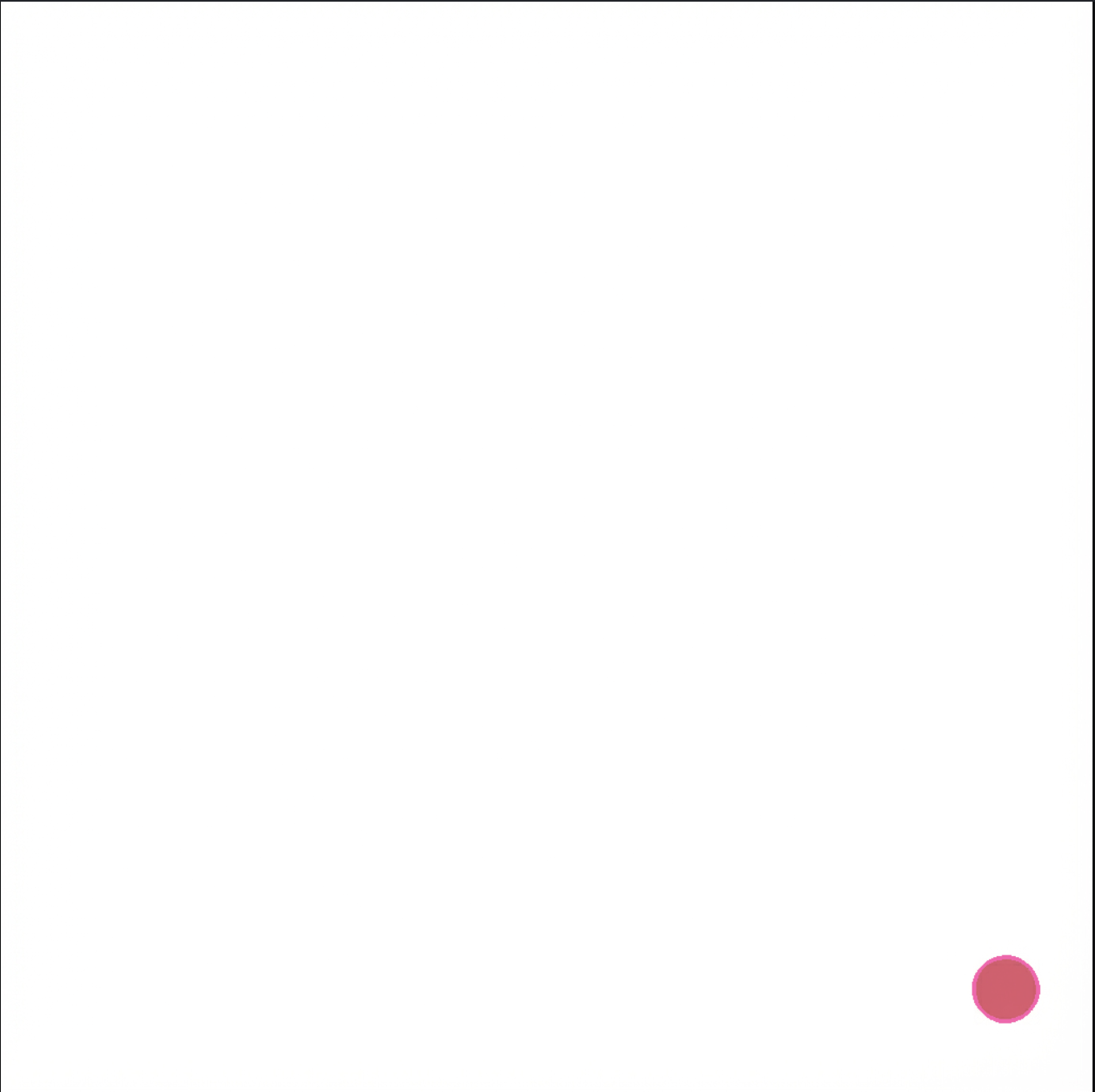}}
    
    \caption{Examples of the SAM3 detecting the red ball for stimuli used in \domainB.
    \label{sam3_example1}
}
\end{figure}

\begin{figure}[b!]
    \centering
    \fbox{\includegraphics[width=0.5\linewidth]{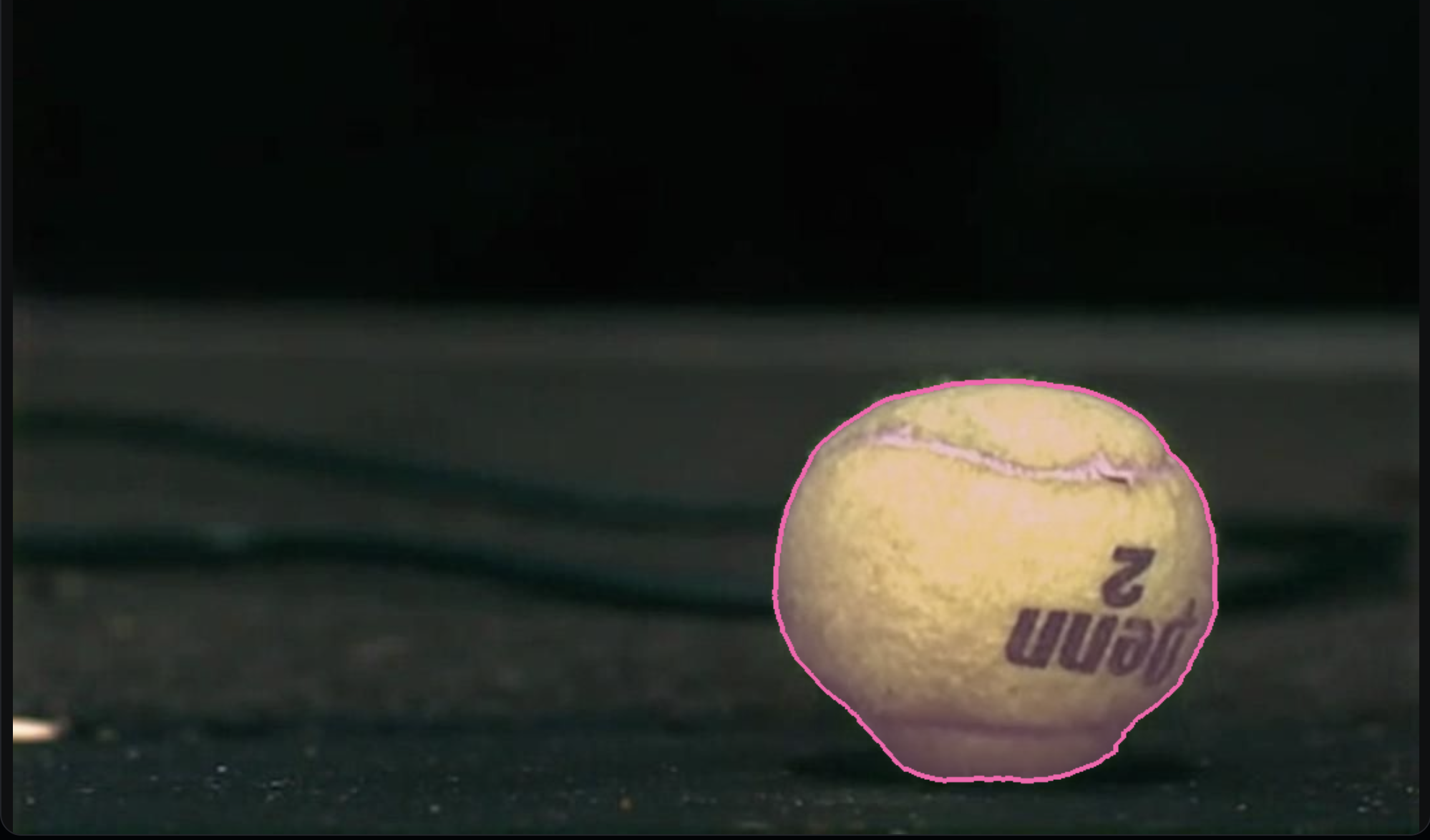}}
    
    \caption{Examples of the SAM3 detecting the tennis ball for stimuli used in \domainD.
}
\label{sam3_example2}
\end{figure}

\begin{figure}[b!]
    \centering
    \fbox{\includegraphics[width=0.5\linewidth]{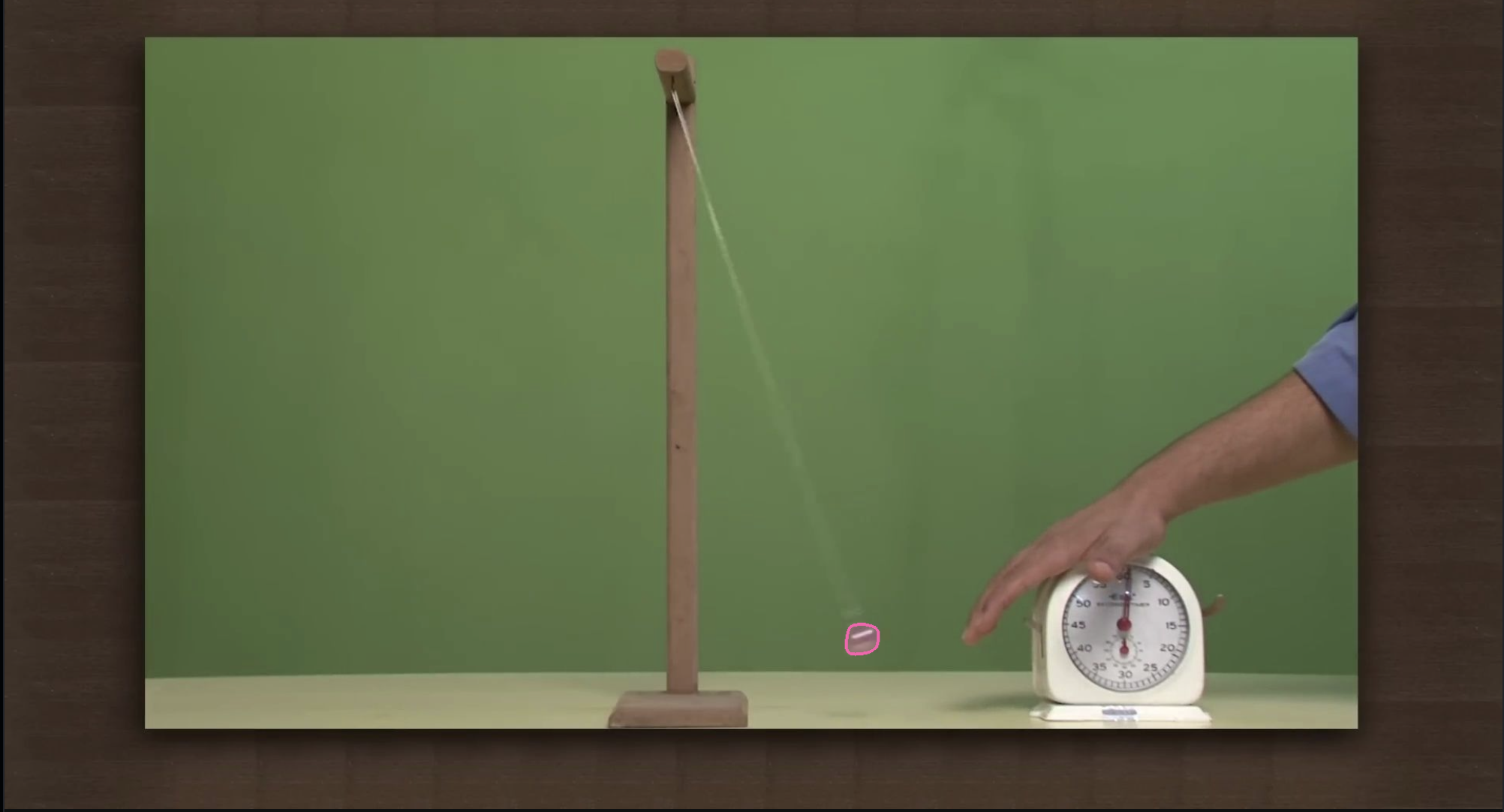}}
    
    \caption{Examples of the SAM3 detecting the silver ball for stimuli used in \domainE.
}
\label{sam3_example3}
\end{figure}

\begin{figure}[b!]
    \centering
    \fbox{\includegraphics[width=0.5\linewidth]{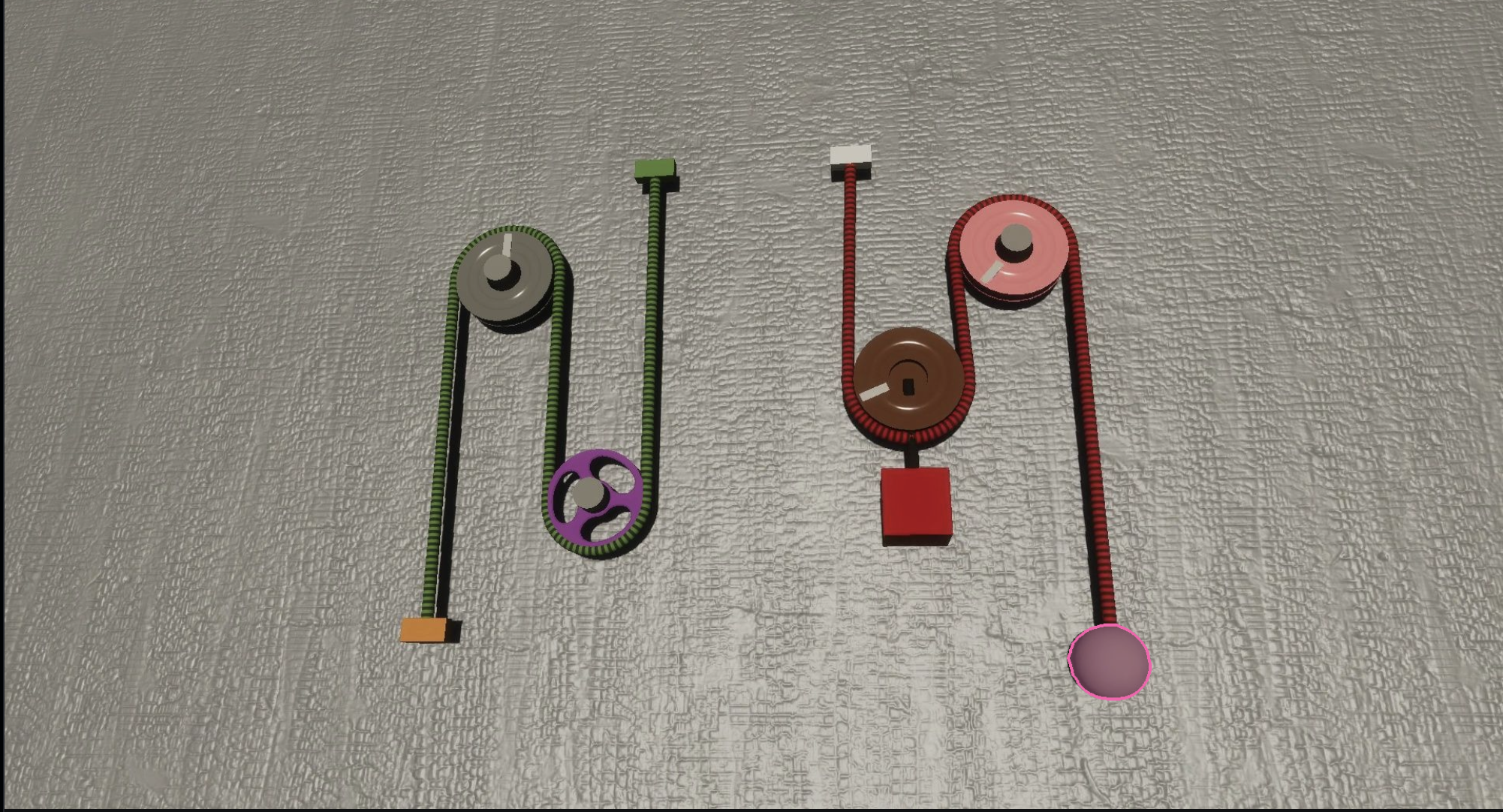}}
    
    \caption{Examples of the SAM3 detecting the grey ball for stimuli used in \domainI.
}
\label{sam3_example4}
\end{figure}

Here we present the example segmentation masks \ref{sam3_example1}, \ref{sam3_example2}, \ref{sam3_example3}, \ref{sam3_example4} generated by SAM3 for the domains we used in the experiment.

\newpage

\subsection{Computer Vision Algorithm used in The \domainC Domain}  \label{app:fluid-cv-algo}

We found SAM 3 to be unreliable in appropriately tracking the water level in the \texttt{Fluids} domain. When using SAM, it consistently identified the stream of water pouring in addition to the water in the mug. This segmentation mask presented challenges for our metric for this domain, which is the top edge of the water level. In order to better identify the water level of the mug, we instead appealed to traditional hand-engineered (so to speak) computer vision algorithms.

The computer vision algorithms we used to detect the water blob and water level in the images generated by the image generation models were coded by ChatGPT-5. The code description provided by the model is as follows:

\texttt{Detect the water ``blob'' inside a glass mug and report its top/bottom y.}

\textbf{Inputs}

\begin{itemize}
  \item \texttt{IGM\_path}: path to RGB/RGBA image.
  \item \texttt{y\_cut}: only analyze rows $y \ge y\_cut$ (ignore the upper part of the image).
  \item \texttt{alpha\_thresh}: pixels with alpha $\le$ this are treated as transparent (ignored).
  \item \texttt{hsv\_lower} / \texttt{hsv\_upper}: HSV thresholds for sky-blue liquid (OpenCV $H\in[0,179]$).
  \item \texttt{erode\_iters}, \texttt{close\_kernel}: morphology parameters to suppress thin rims/stream and fill gaps.
  \item \texttt{coverage\_main} / \texttt{coverage\_fallback}: minimum horizontal coverage (fraction of bbox width) required for a row to be considered liquid; a fallback is used if the main threshold yields none.
  \item \texttt{ignore\_top\_rows}: discard the first $N$ rows inside the ROI to avoid picking the crop boundary.
\end{itemize}

\textbf{Steps}
\begin{enumerate}
  \item Read the image with \texttt{cv2.IMREAD\_UNCHANGED} so the alpha channel (if present) is preserved.
  \item Build an alpha mask: \(\text{alpha\_mask} = (\alpha > \text{alpha\_thresh})\). If no alpha channel exists, use an all-ones mask (all pixels opaque).
  \item Crop to the region of interest (ROI): rows \(y \ge y\_cut\). Crop the alpha mask the same way.
  \item Convert the ROI to HSV and threshold with \([\texttt{hsv\_lower}, \texttt{hsv\_upper}]\) to obtain a binary color mask for the sky-blue liquid.
  \item AND the color mask with the alpha mask to suppress fully/mostly transparent background.
  \item Morphological cleanup:
  \begin{itemize}
    \item Erode (\texttt{erode\_iters} times) to remove thin bright rims and the pouring stream.
    \item Close (kernel size \(=\) \texttt{close\_kernel}) to reconnect the eroded blob and fill small holes.
  \end{itemize}
  \item Find connected components (external contours) and select the largest-area component as the liquid blob.
  \item Compute the blob's bounding box \((x, y, w, h)\) in ROI coordinates. Inside this box:
  \begin{itemize}
    \item For each row, count foreground pixels (\(\text{row\_counts}\)).
    \item Mark rows as liquid if \(\text{row\_counts} \ge \text{coverage\_main} \times w\).
    \item If none qualify, relax to \(\text{coverage\_fallback} \times w\).
    \item Discard candidate rows whose index \(<\) \texttt{ignore\_top\_rows}.
  \end{itemize}
  \item Determine water-level bounds:
  \begin{itemize}
    \item Top \(y\) (global) \(= y\_cut + y + \min(\text{valid\_row\_indices})\) if valid rows exist; otherwise \(y\_cut + \min(\text{contour\_y})\).
    \item Bottom \(y\) (global) \(= y\_cut + y + \max(\text{valid\_row\_indices})\) if valid rows exist; otherwise \(y\_cut + \max(\text{contour\_y})\).
  \end{itemize}
  \item Optional visualization:
  \begin{itemize}
    \item Red rectangle: blob bounding box.
    \item Green line: top \(y\).
    \item Blue line: bottom \(y\).
  \end{itemize}
\end{enumerate}

\textbf{Notes}
\begin{itemize}
  \item HSV bounds are intentionally broad; tune to lighting and hue variations.
  \item Erosion suppresses 1–2\,px rims and stream artifacts that otherwise bias the ``top'' level.
  \item The row-coverage rule prefers rows where a substantial horizontal span is filled, making the estimate robust to foam/splash highlights.
  \item Complexity is approximately \(O(HW)\) per frame (thresholding + morphology + a single bbox scan).
\end{itemize}

\begin{figure}[b!]
    \centering
    \fbox{\includegraphics[width=0.5\linewidth]{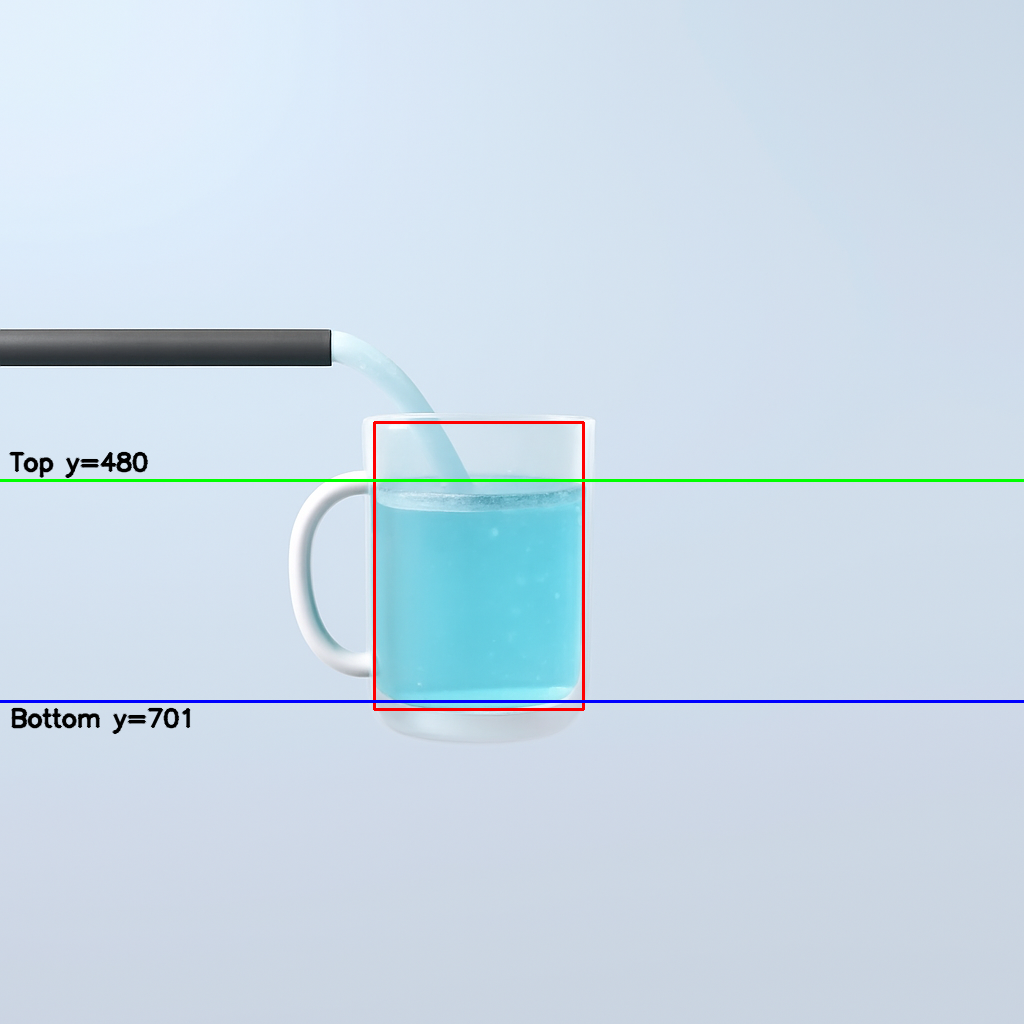}}
    
    \caption{Examples of the algorithm detecting the water level for stimuli used in \domainC.
}
    \label{fig:example2}
\end{figure}

And Fig.~\ref{fig:example2} is an example of the algorithm detecting the water level in images generated in \domainC.

\clearpage

\section{Additional Analysis}
\label{app:additional-analysis}

\subsection{The \domainC Domain}

In Section \ref{sec:results}, we observed that \textit{Chain-of-Time} Quarter and Half exhibited higher average error than Direct Prediction in the fluid-dynamics domain. This motivated us to examine how the model perceives the initial state sequence $X_{0:t}$, which in the case of \domainC includes physical parameters such as flow rate and initial water level.

\begin{figure}[h!]
    \centering
    \includegraphics[width=.9\linewidth]{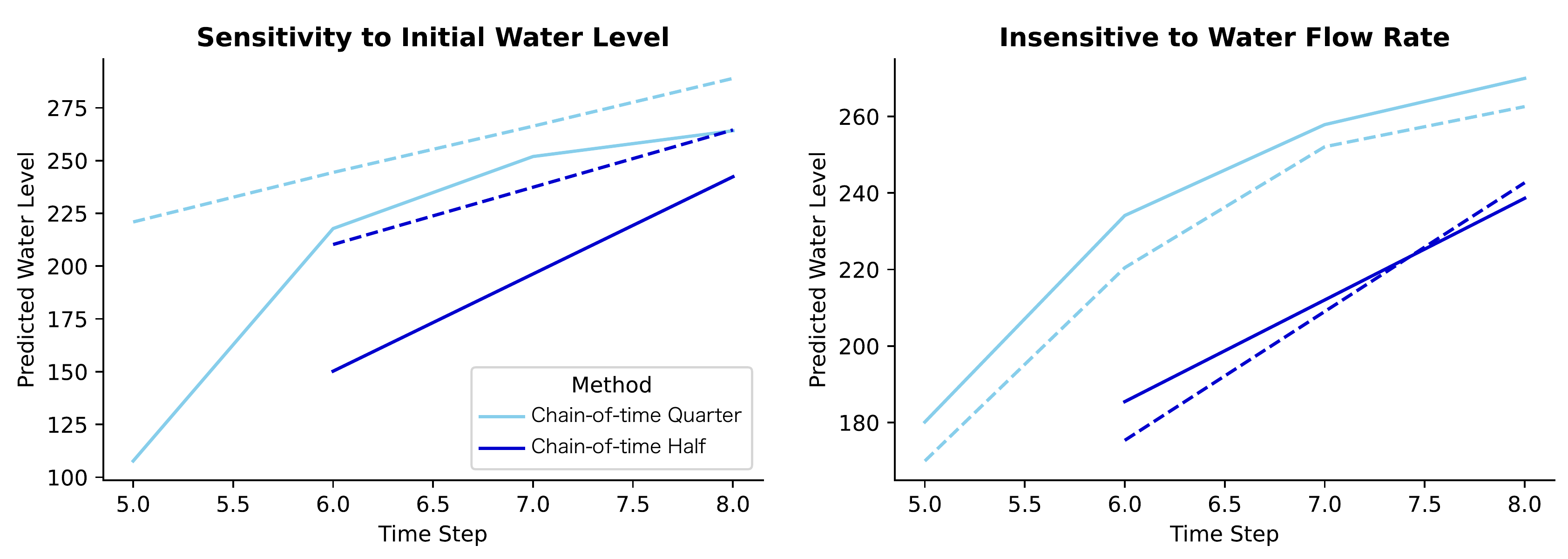}
    \caption{In the \domainC \ domain, we find that the IGMs are able to simulate water levels increasing over time. Here we show water levels in generated images steadily increasing as a function of time. (Left) The model is sensitive to initial water levels, with solid lines representing a low water level at the initial time of simulation $t$ and dotted lines representing a high initial water level. (Right) However, the model is insensitive to the flow rate of water, with water level consistently increasing at the same rate during simulations. Solid lines represent a slow flow rate and dotted lines represent a 3x faster flow rate.}
    \label{fig:fluids_ts}
\end{figure}

We analyzed the simulated image sequences $\widehat{I}_{t:T}$ generated by the IGM in \domainC. Specifically, we examined the predicted water-level time series for stimuli with slow ($25$ frames/s) and fast ($75$ frames/s) flow rates, and with low ($1/12$ and $3/12$ full) versus high ($7/12$ and $9/12$ full) initial water levels. As shown in Fig.~\ref{fig:fluids_ts}, the IGM is sensitive to the initial water level—lowering the initial level shifts the intercept (dotted to solid line; left panel)—but largely insensitive to flow rate—the slope remains similar for $25$ and $75$ frames/s (right panel).

The sensitivity to the initial water level indicates that the IGM leverages salient visual cues in the inputs, whereas its insensitivity to flow rate suggests substantial error in estimating that parameter. Flow rate is a critical physical variable in this task, ensuring that the glass mugs are filled with the correct amount of water over time. These findings imply that errors in estimating the initial state $X_{0:t}$ can occur, especially for parameters that are not directly observable. Such estimation error accumulates over the rollout: as the \textit{Chain-of-Time} step size $k$ decreases (i.e., more, finer steps), the accumulated error grows. This helps explain why \textit{Chain-of-Time} underperforms Direct Prediction in the fluid-dynamics domain, and why the Quarter configuration performs worse than Half and Direct Prediction.

\clearpage

\subsection{The \domainD Domain}

Here are the additional trials in \domainD:

\begin{figure}[h]
    \centering
    \includegraphics[width= \linewidth]{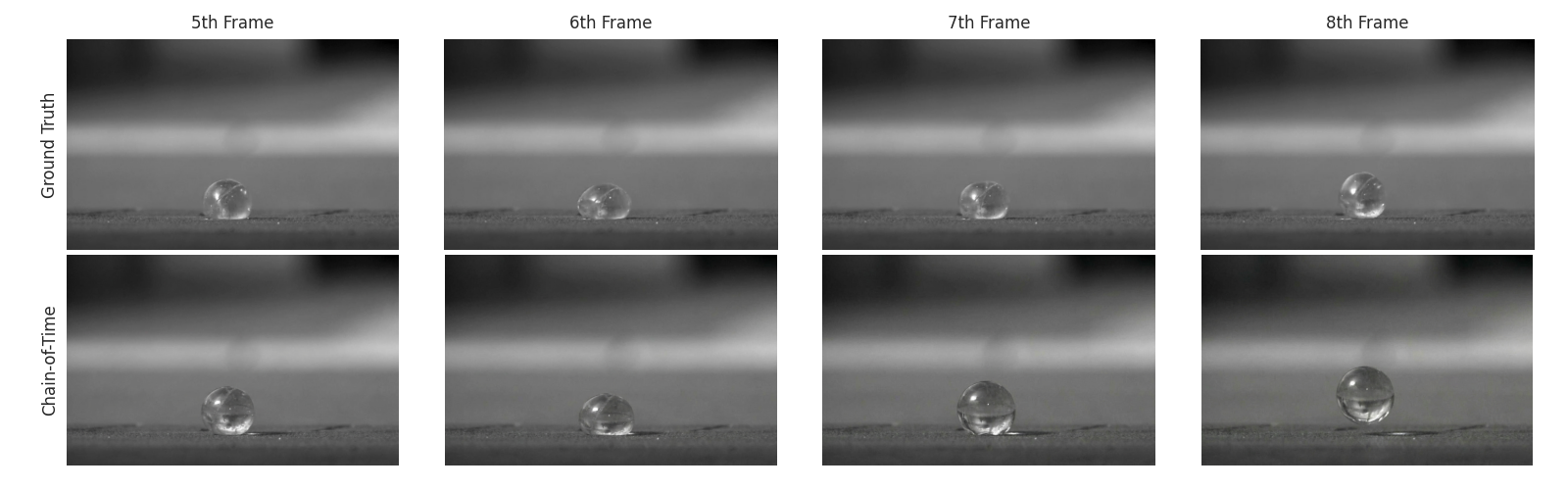}
    \includegraphics[width= \linewidth]{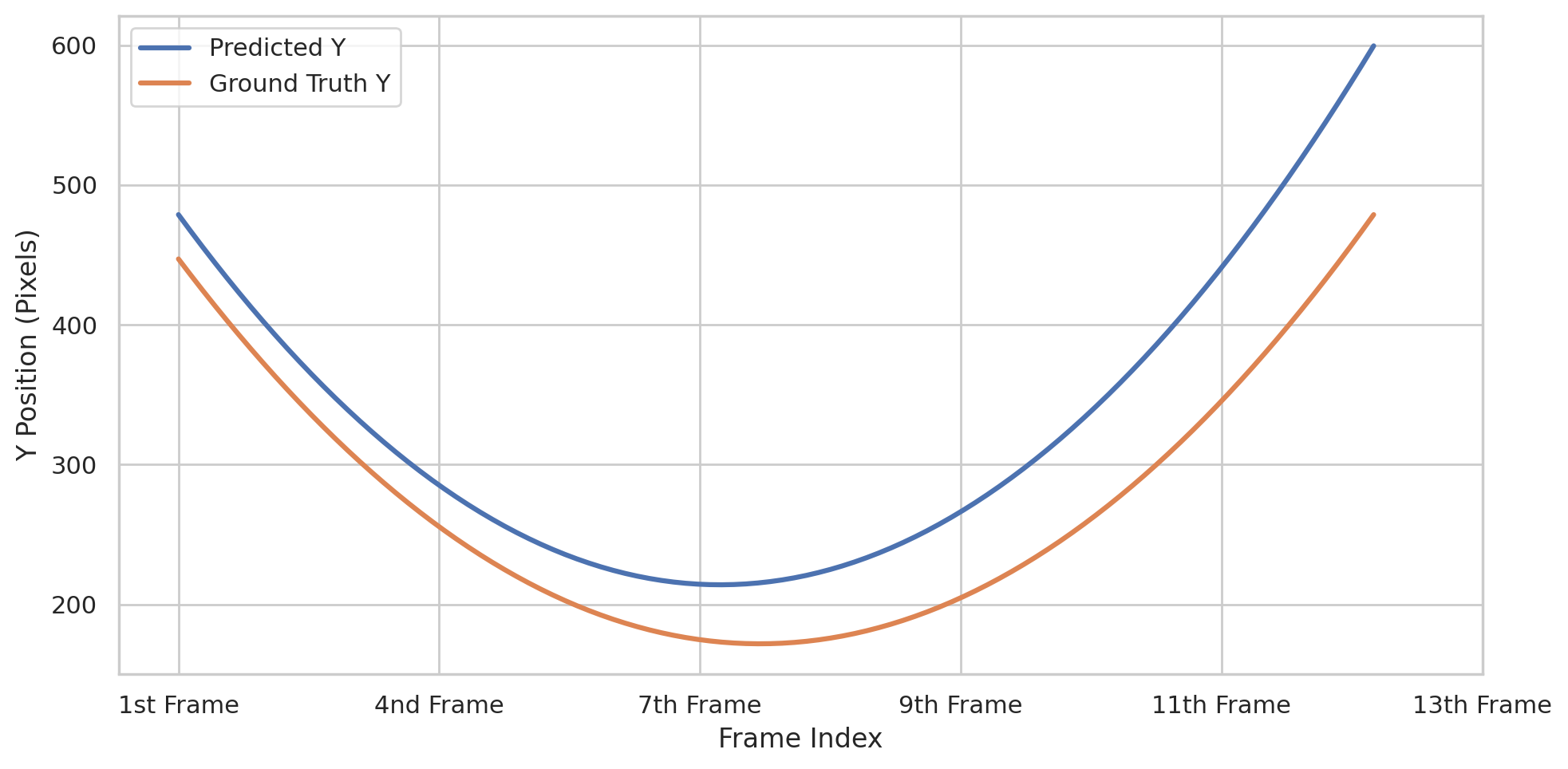}
    \caption{
    \textbf{Simulation analysis for \texttt{Bouncing} domain} \ \  (Top) Comparison between ground truth images and images generated by Chain-of-Time (quarter) for an example trial. The images shown correspond to the ``during'' period of the collision, beginning at frame 5 of 12. (Bottom) Extracted Y coordinates of the transparent ball in each frame, for three independent simulation results of four frames each: one simulation before the bouncing collision, one during, and one after collision. We observe parabolic motion in the ball for both the ground truth and simulated image sequences.
}
\end{figure}

\begin{figure}[h]
    \centering
    \includegraphics[width= \linewidth]{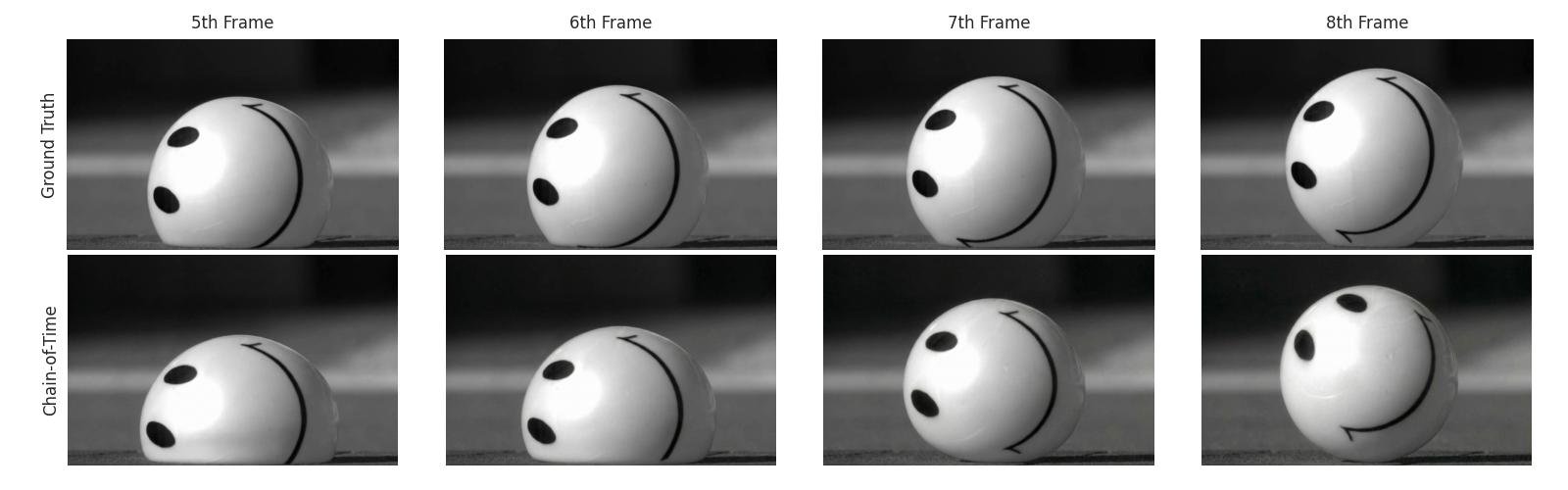}
    \includegraphics[width= \linewidth]{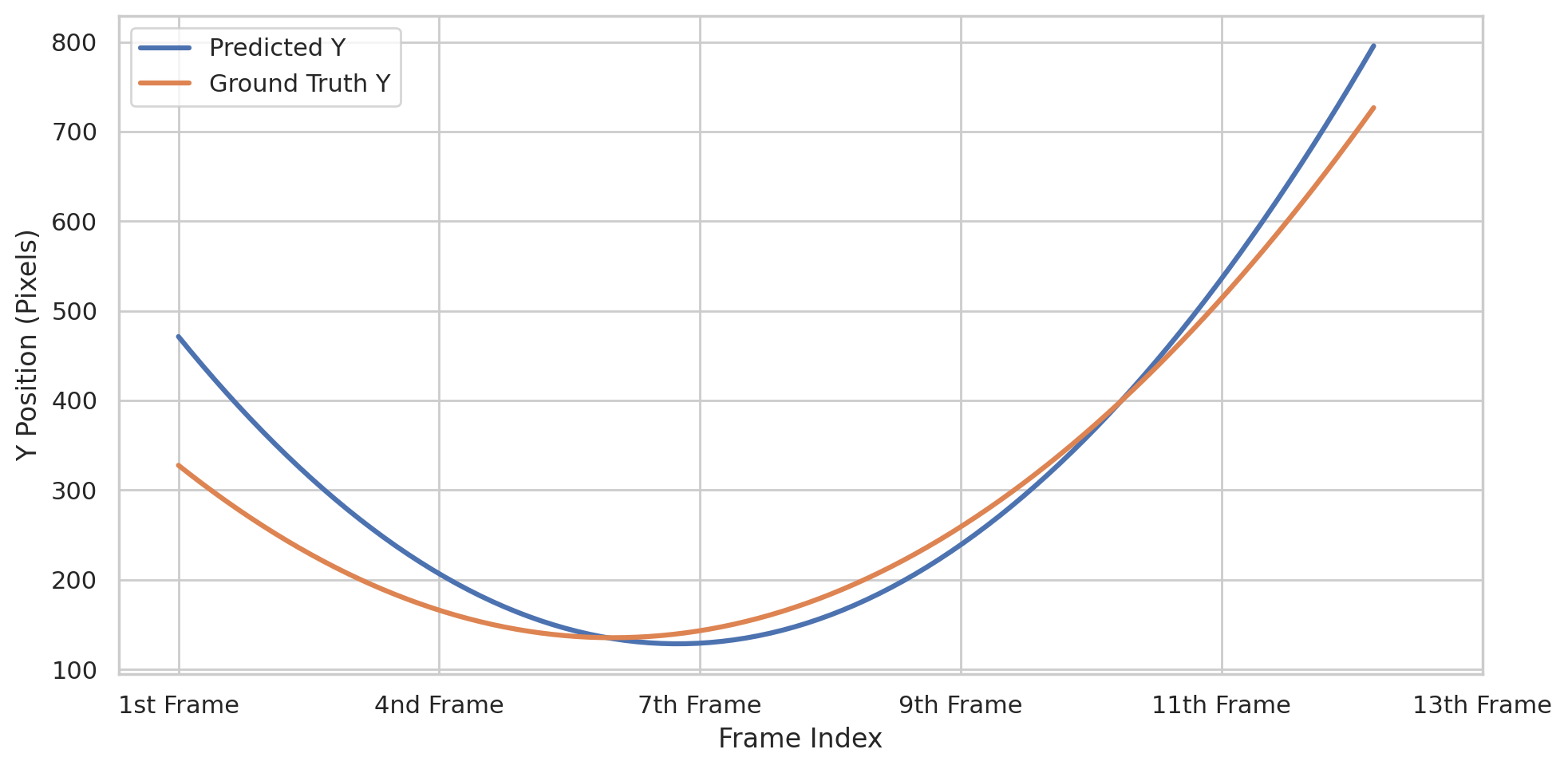}
    \caption{\textbf{Simulation analysis for \texttt{Bouncing} domain} \ \  (Top) Comparison between ground truth images and images generated by Chain-of-Time (quarter) for an example trial. The images shown correspond to the ``during'' period of the collision, beginning at frame 5 of 12. (Bottom) Extracted Y coordinates of the smiley face ball in each frame, for three independent simulation results of four frames each: one simulation before the bouncing collision, one during, and one after collision. We observe parabolic motion in the ball for both the ground truth and simulated image sequences.
}
\end{figure}

\clearpage

\subsection{The \domainE Domain}

Here are the additional trials in \domainE:

\begin{figure}[h]
    \centering
    \includegraphics[width= \linewidth]{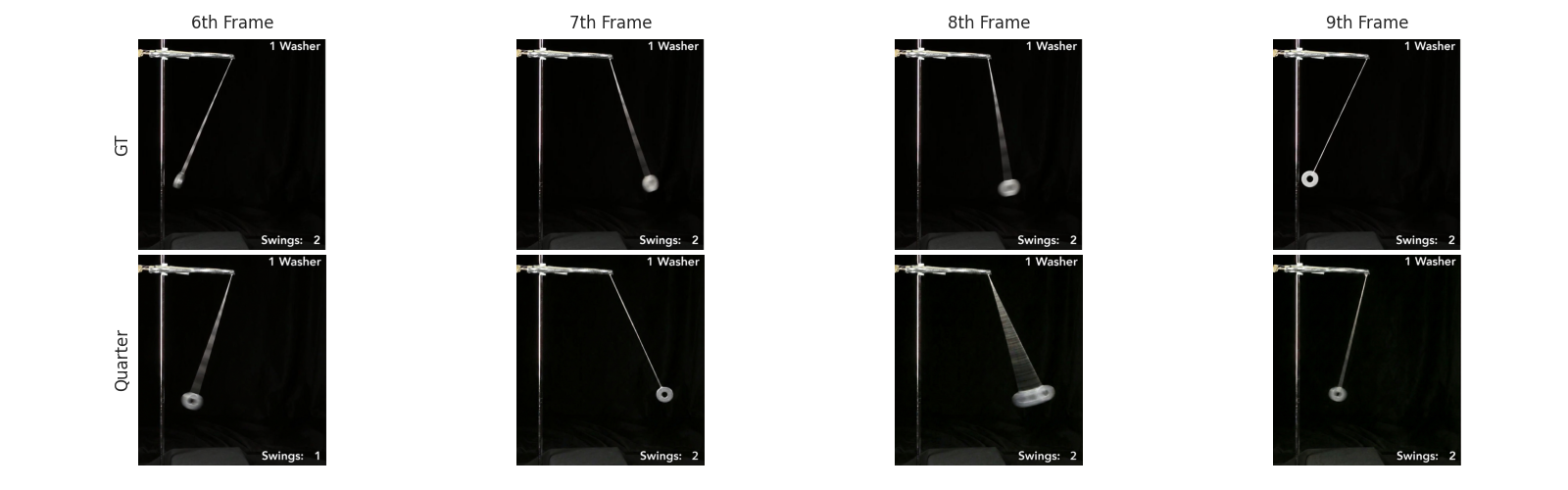}
    \includegraphics[width= \linewidth]{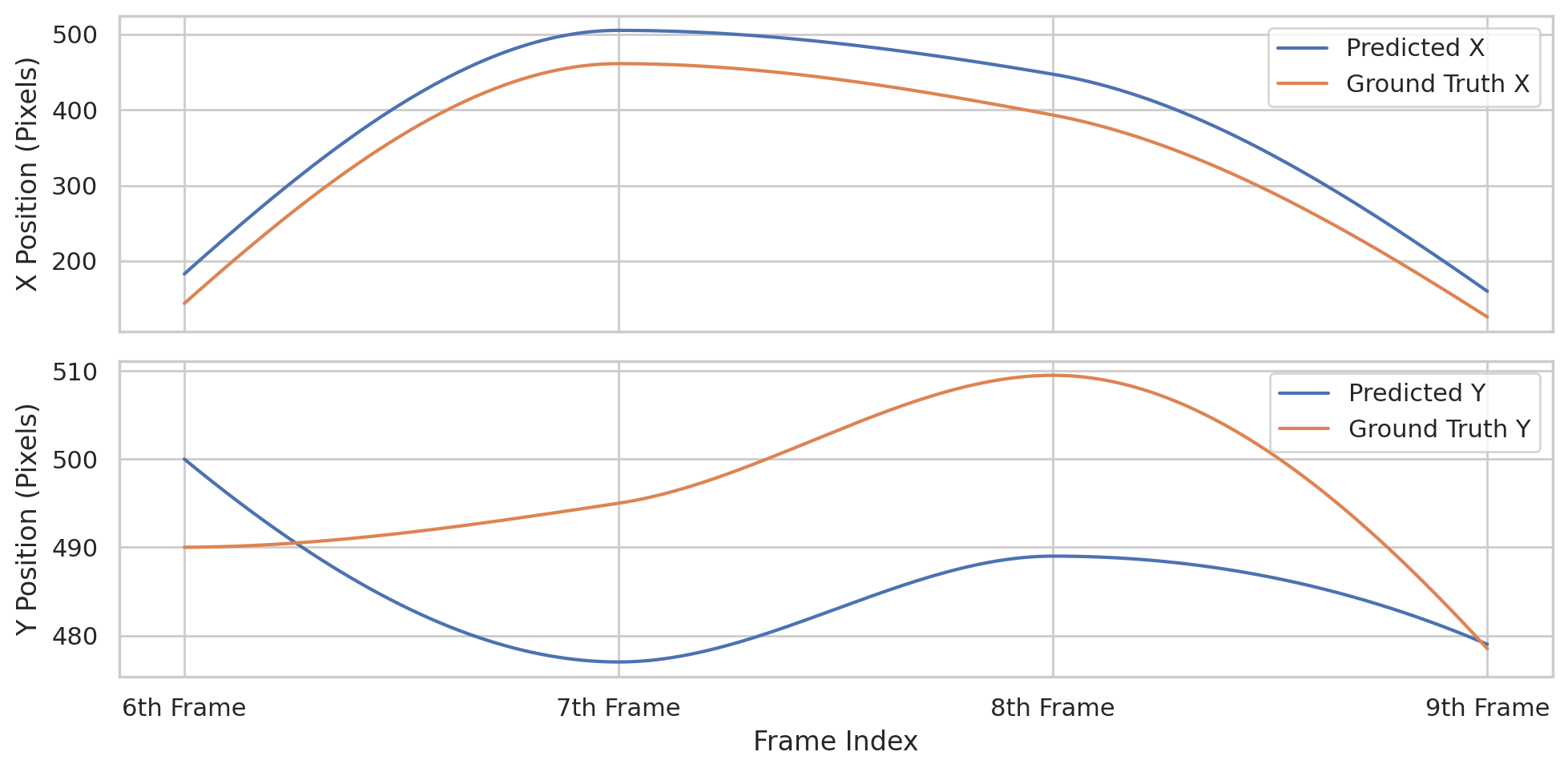}
    \caption{\textbf{Simulation analysis for \texttt{Pendulum} domain} \ \  (Top) Comparison between ground truth images and images generated by Chain-of-Time (quarter) for an example trial.  (Bottom) Extracted X and Y coordinates of the pendulum in each frame. We observe oscillatory motion in the pendulum for both the ground truth and simulated image sequences.
}
\end{figure}

\begin{figure}[h]
    \centering
    \includegraphics[width= \linewidth]{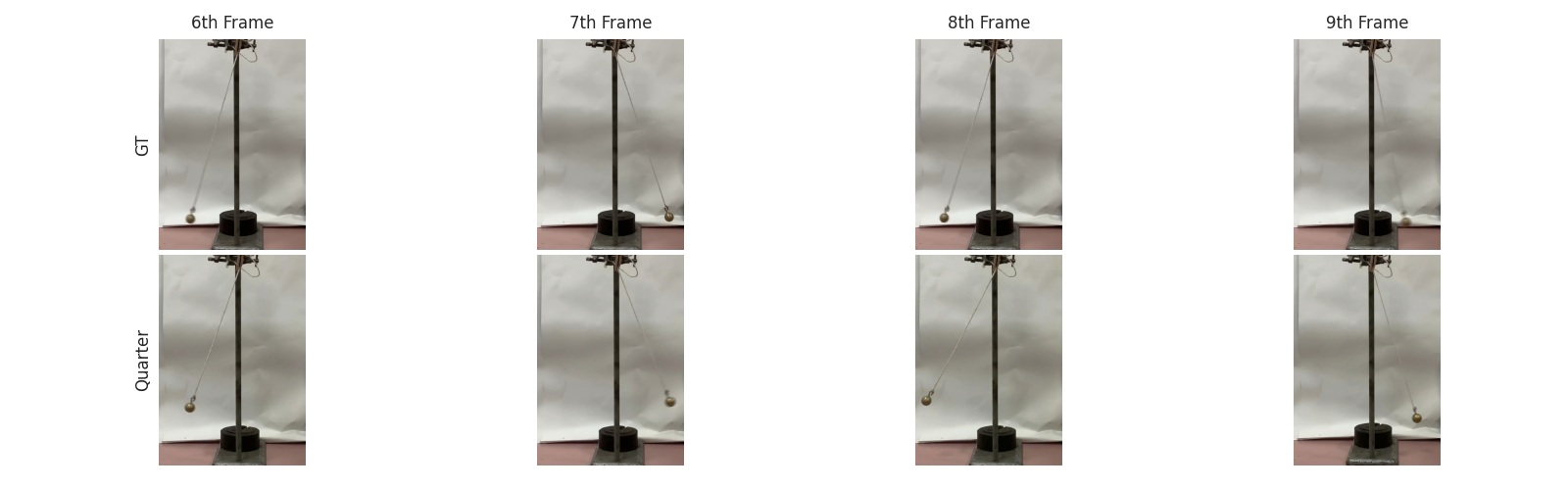}
    \includegraphics[width= \linewidth]{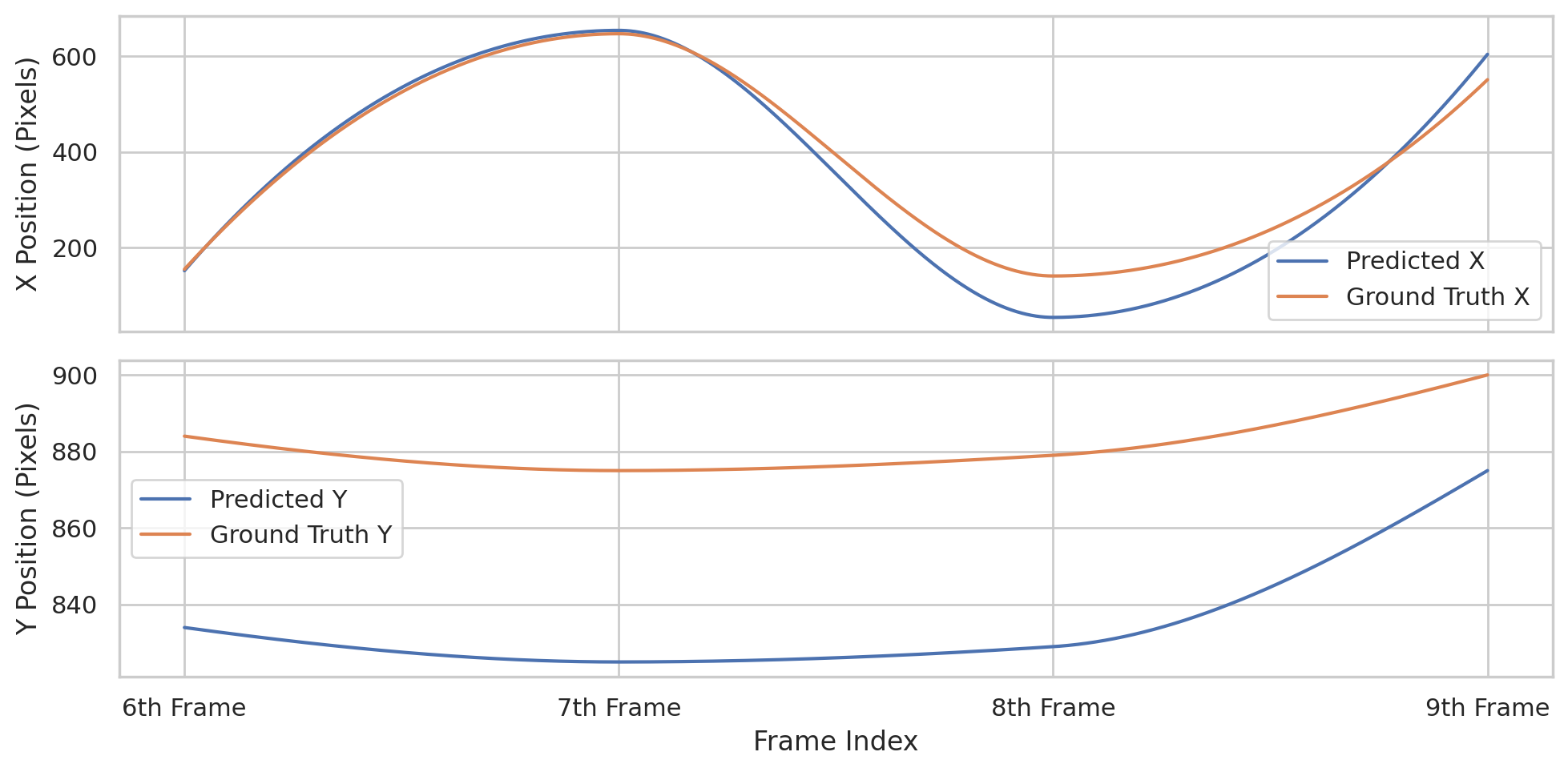}
    \caption{\textbf{Simulation analysis for \texttt{Pendulum} domain} \ \  (Top) Comparison between ground truth images and images generated by Chain-of-Time (quarter) for an example trial.  (Bottom) Extracted X and Y coordinates of the pendulum in each frame. We observe oscillatory motion in the pendulum for both the ground truth and simulated image sequences.
}
\end{figure}

\clearpage

\subsection{The \domainF Domain}

Here are the additional trials in \domainF:

\begin{figure}[h]
    \centering
    \includegraphics[width=\linewidth]{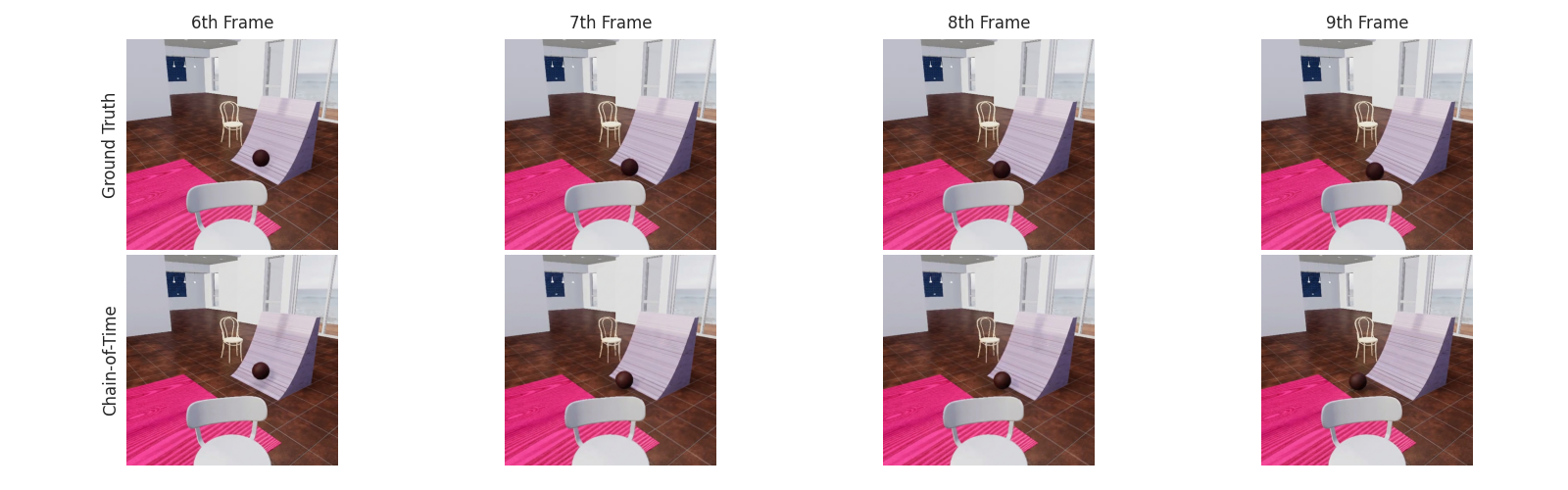}
    \includegraphics[width=\linewidth]{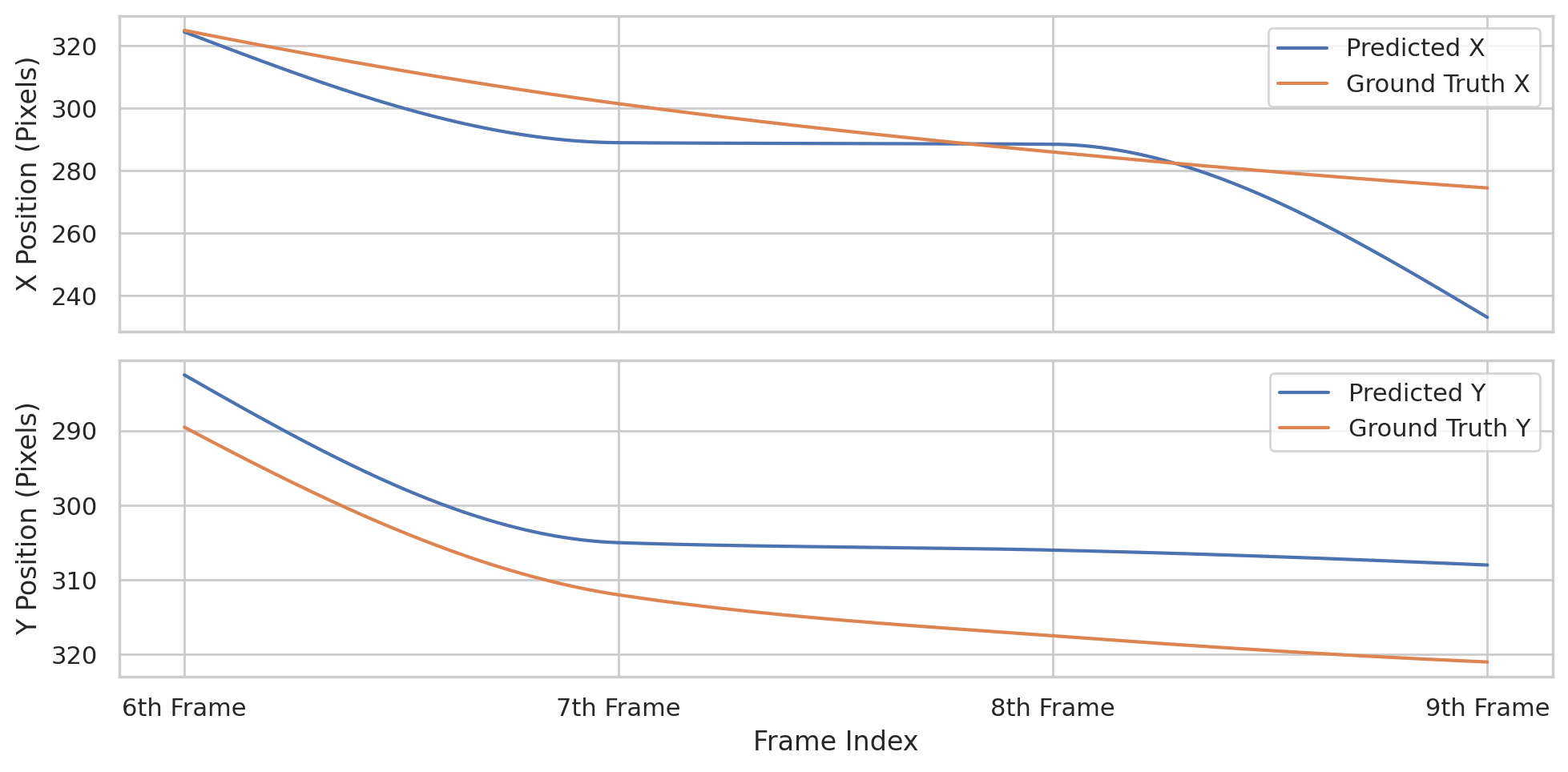}
    \caption{\textbf{Simulation analysis for \texttt{Rolling on Slope} domain} \ \  (Top) Comparison between ground truth images and images generated by Chain-of-Time (quarter) for a single trial. Frame indexing begins at 6 because the first 5 frames are provided as input to the model. (Bottom) Extracted X and Y coordinates of the ball in each frame. The ground truth and simulated objects follow very similar trajectories, first accelerating sharply and then steadily decelerating.
}
\end{figure}

\begin{figure}[h]
    \centering
    \includegraphics[width=\linewidth]{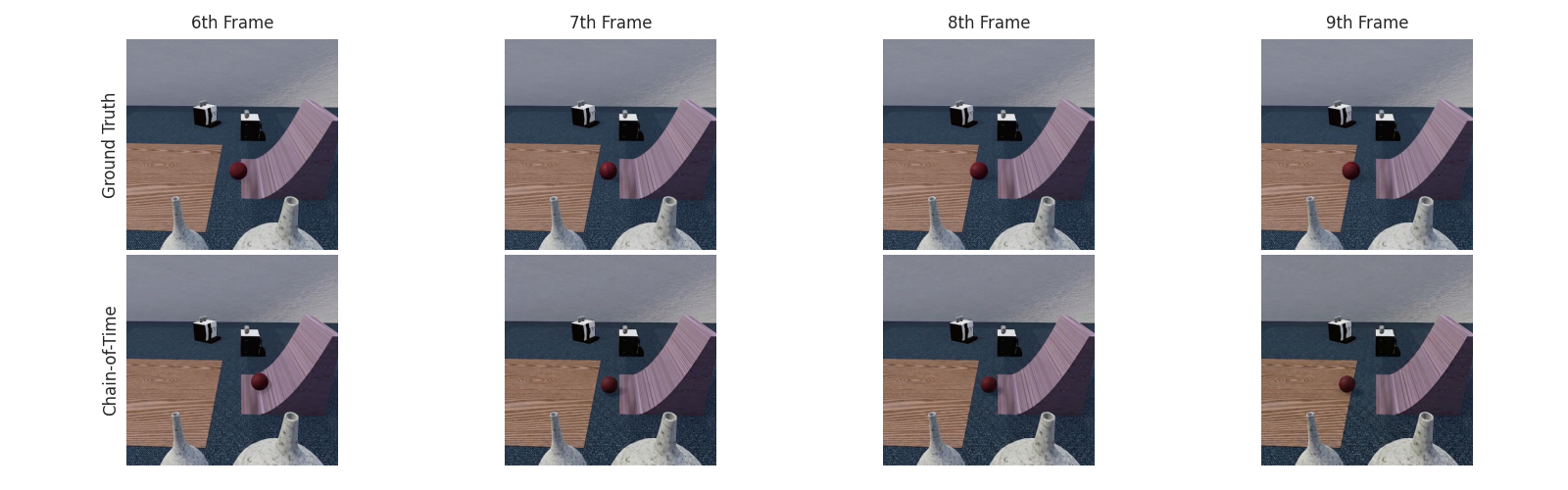}
    \includegraphics[width=\linewidth]{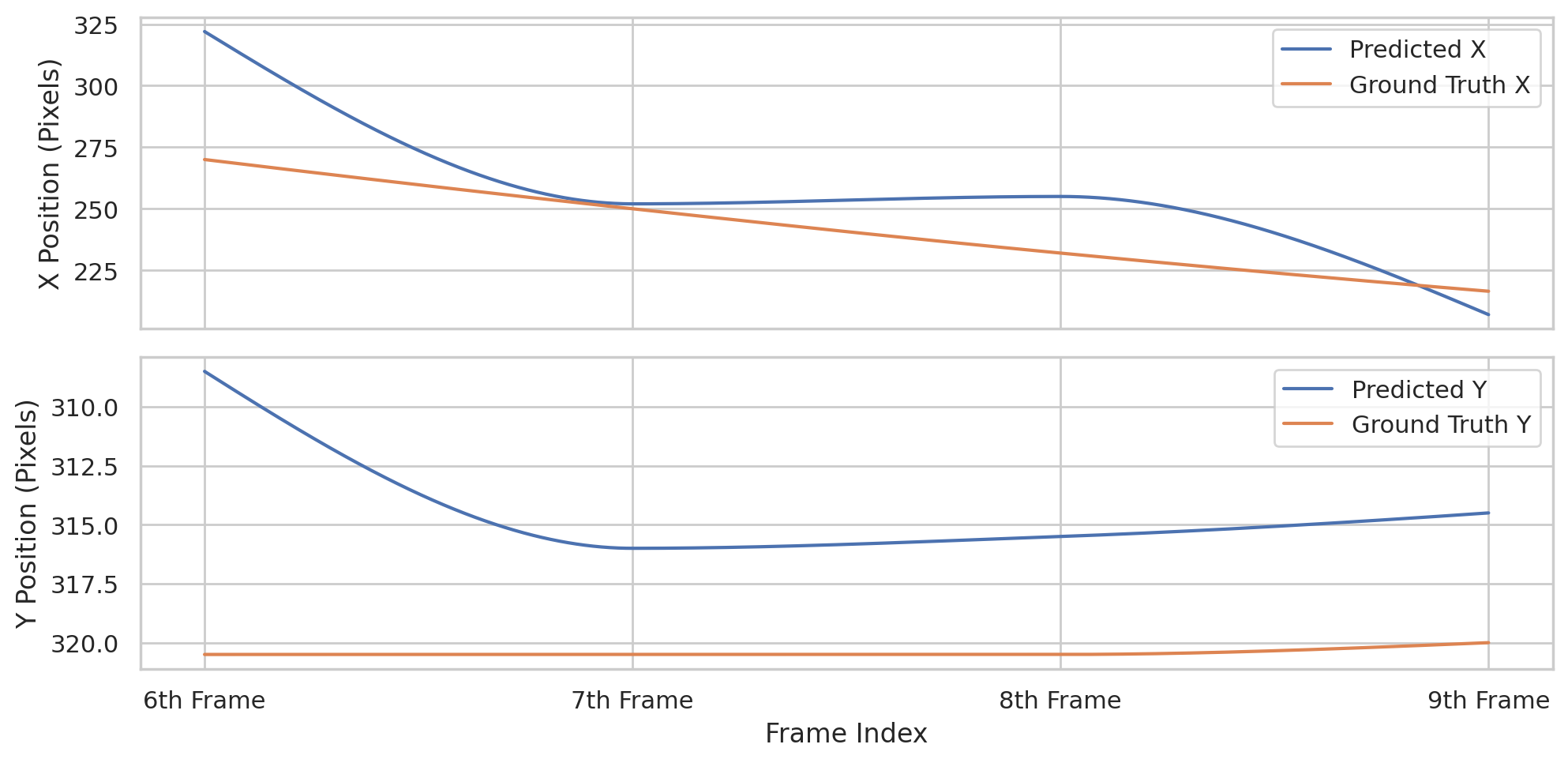}
    \caption{\textbf{Simulation analysis for \texttt{Rolling on Slope} domain} \ \  (Top) Comparison between ground truth images and images generated by Chain-of-Time (quarter) for a single trial. Frame indexing begins at 6 because the first 5 frames are provided as input to the model. (Bottom) Extracted X and Y coordinates of the ball in each frame. The ground truth and simulated objects follow very similar trajectories, first accelerating sharply and then steadily decelerating.
}
\end{figure}

\clearpage

\subsection{The \domainG Domain}

Here are the additional trials in \domainG:

\begin{figure}[h]
    \centering
    \includegraphics[width=\linewidth]{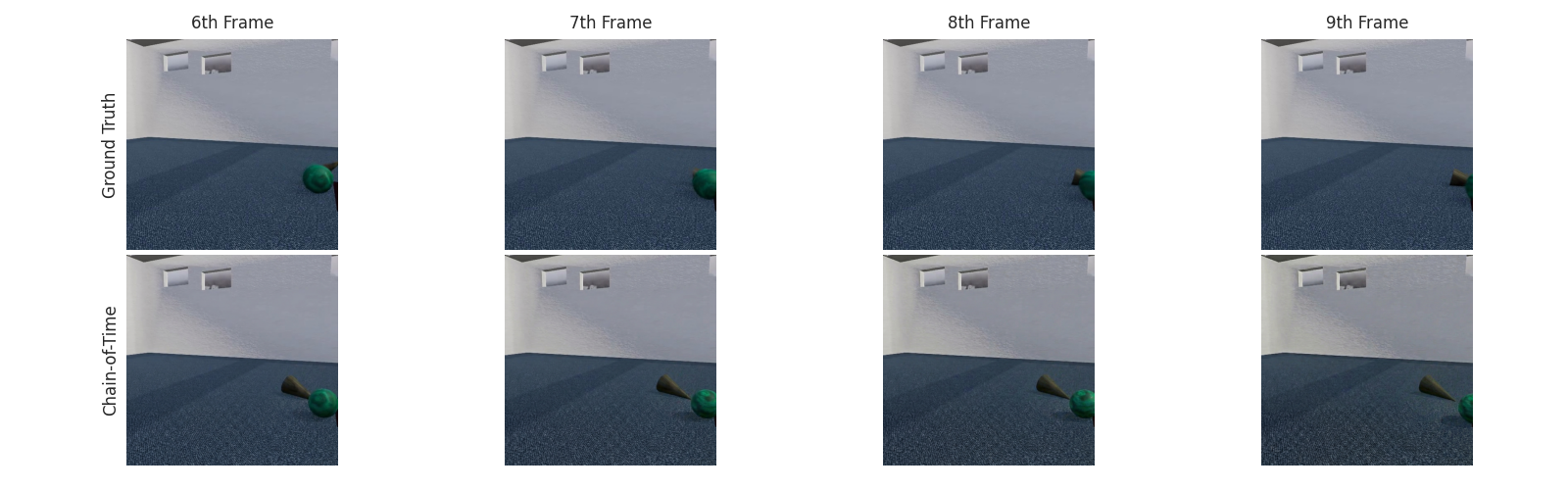}
    \includegraphics[width=\linewidth]{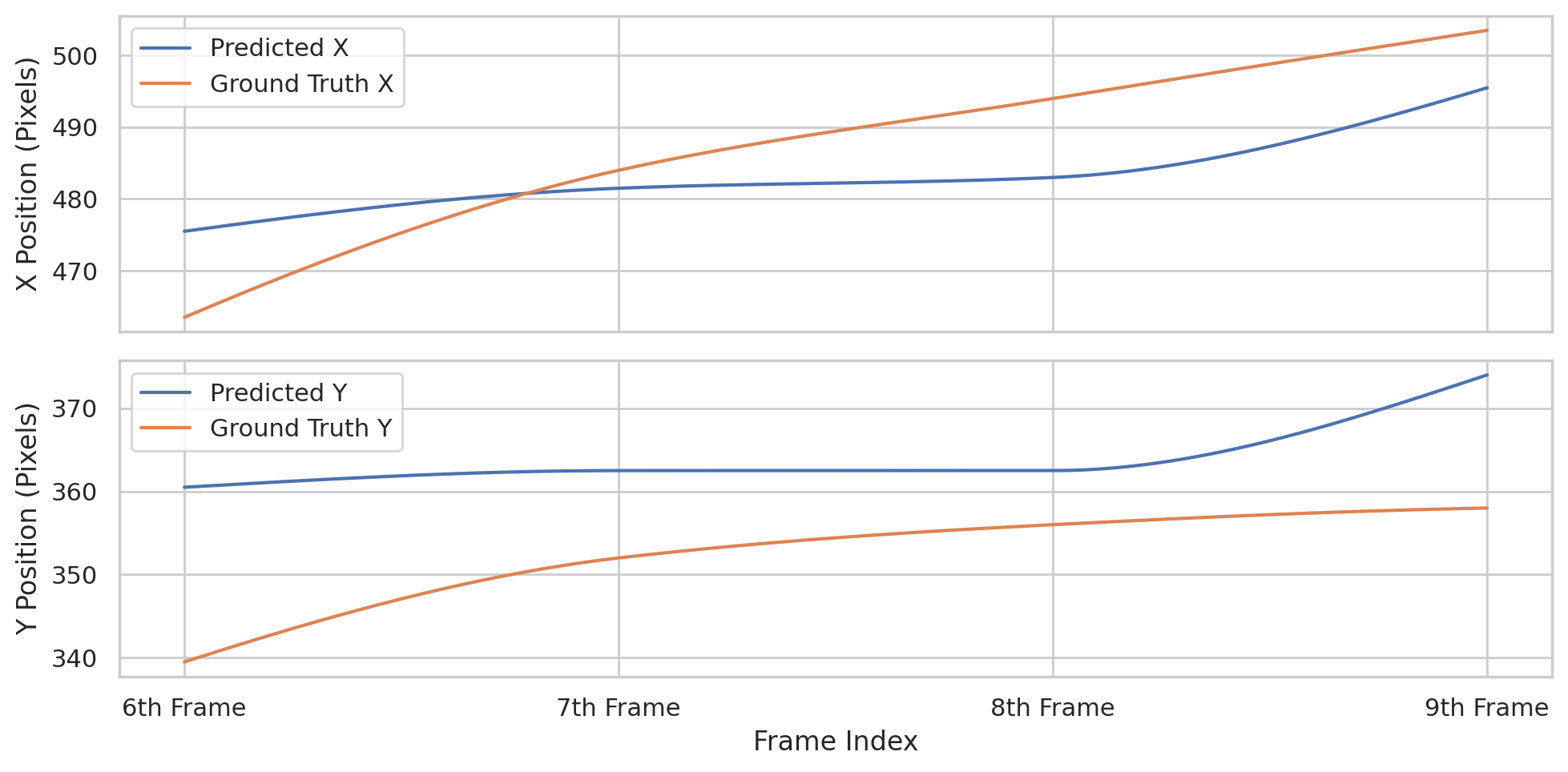}
    \caption{\textbf{Simulation analysis for \texttt{Collision} domain} \ \  (Top) Comparison between ground truth images and images generated by Chain-of-Time (quarter) for a single trial. Frame indexing begins at 6 because the first 5 frames are provided as input to the model. (Bottom) Extracted X and Y coordinates of the ball in each frame. The ground truth and simulated objects follow very similar trajectories.
}
\end{figure}

\begin{figure}[h]
    \centering
    \includegraphics[width=\linewidth]{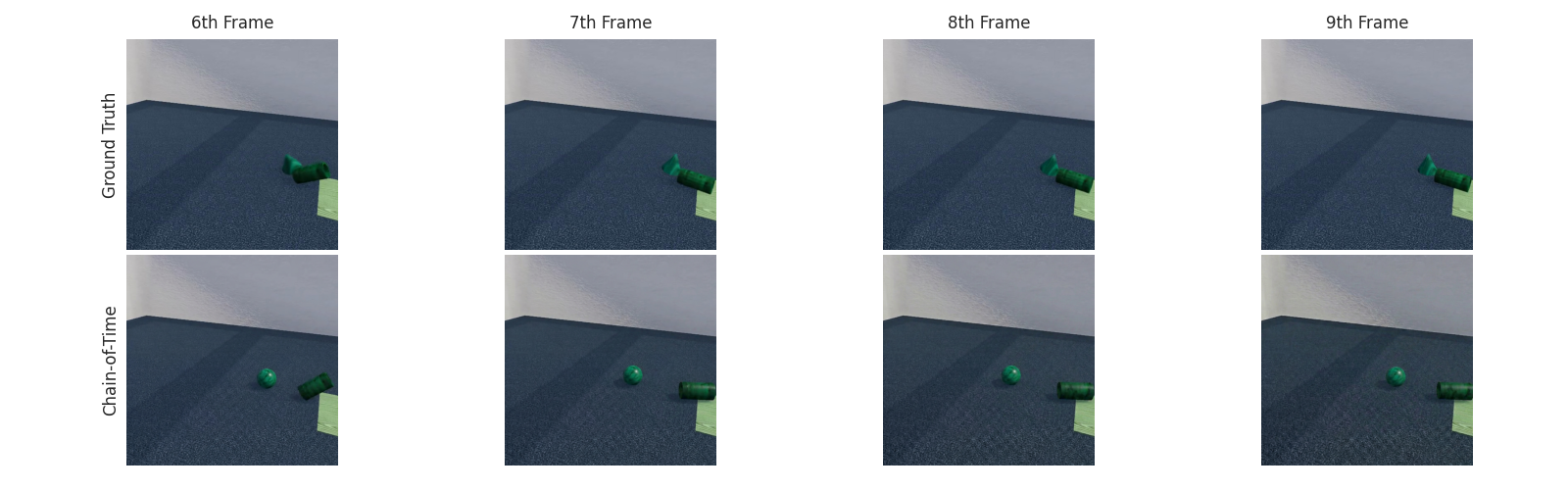}
    \includegraphics[width=\linewidth]{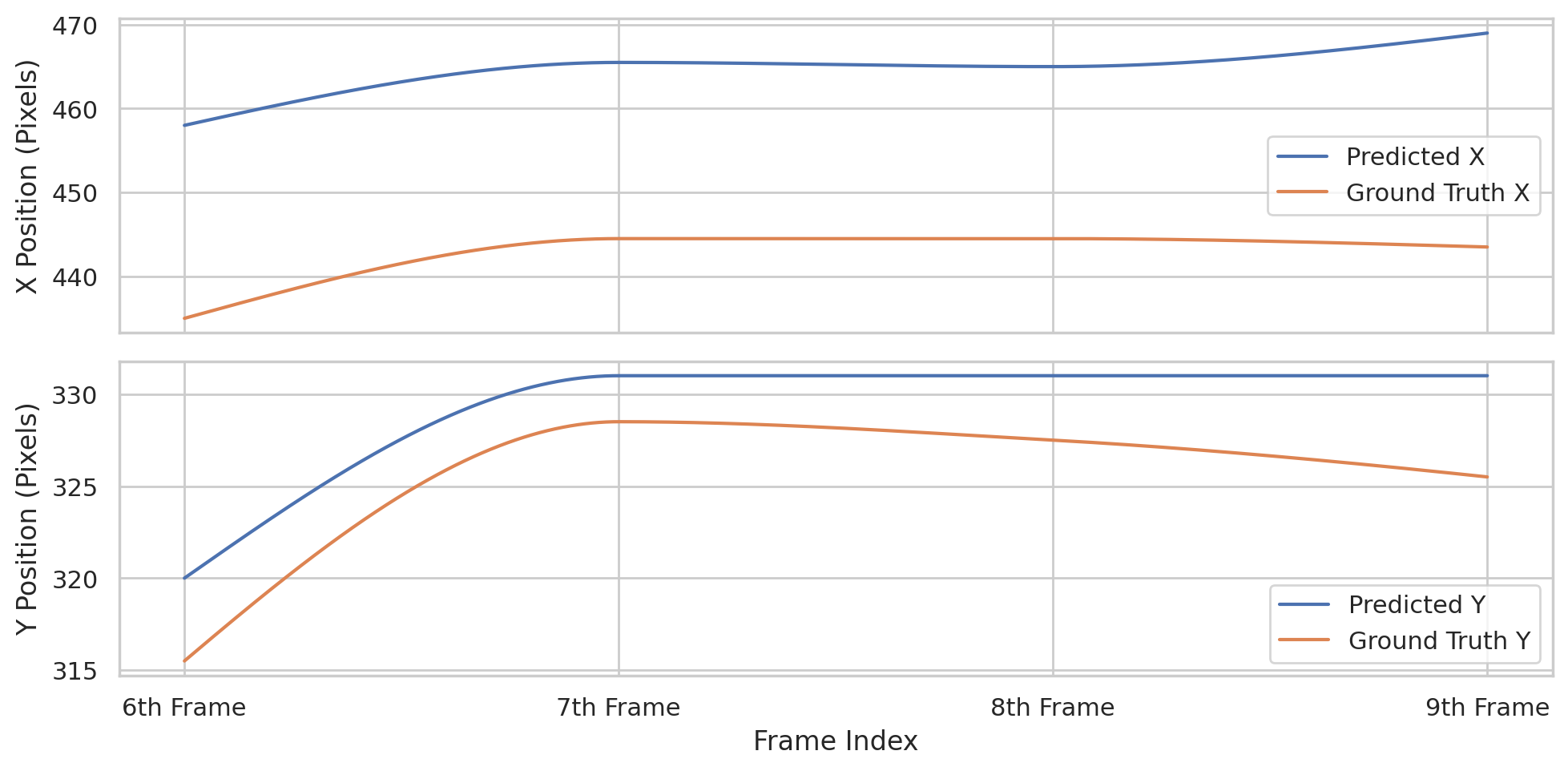}
    \caption{\textbf{Simulation analysis for \texttt{Collision} domain} \ \  (Top) Comparison between ground truth images and images generated by Chain-of-Time (quarter) for a single trial. Frame indexing begins at 6 because the first 5 frames are provided as input to the model. (Bottom) Extracted X and Y coordinates of the ball in each frame. The ground truth and simulated objects follow very similar trajectories.
}
\end{figure}

\clearpage

\subsection{The \domainH Domain}

Here are the additional trials in \domainH:

\begin{figure}[h]
    \centering
    \includegraphics[width=\linewidth]{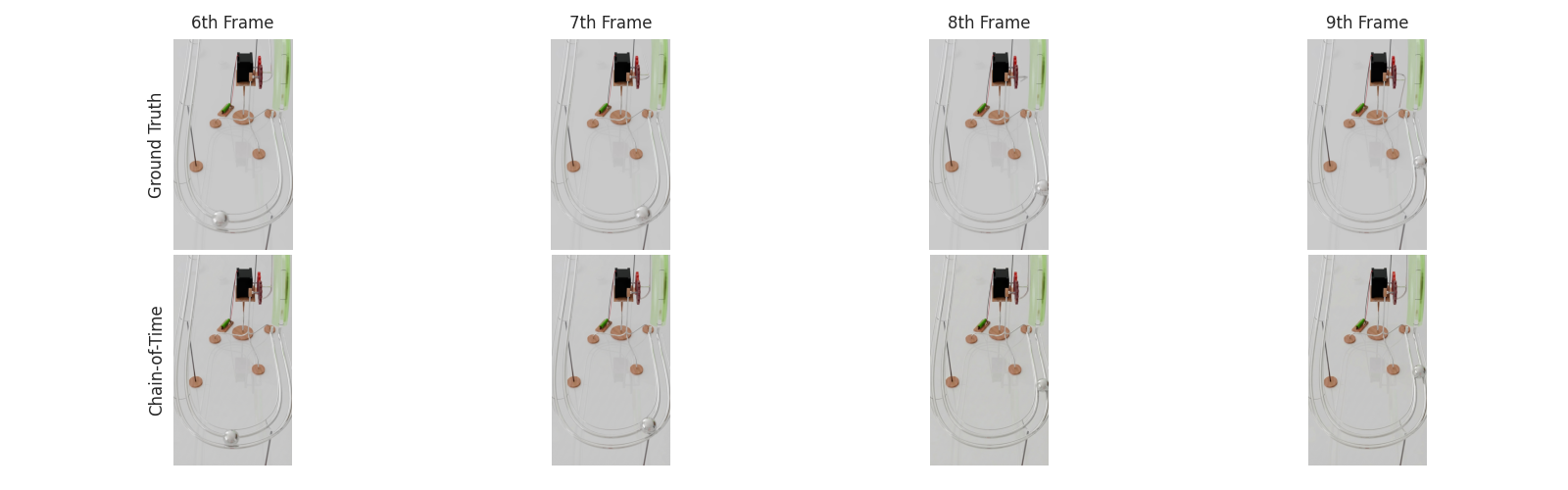}
    \includegraphics[width=\linewidth]{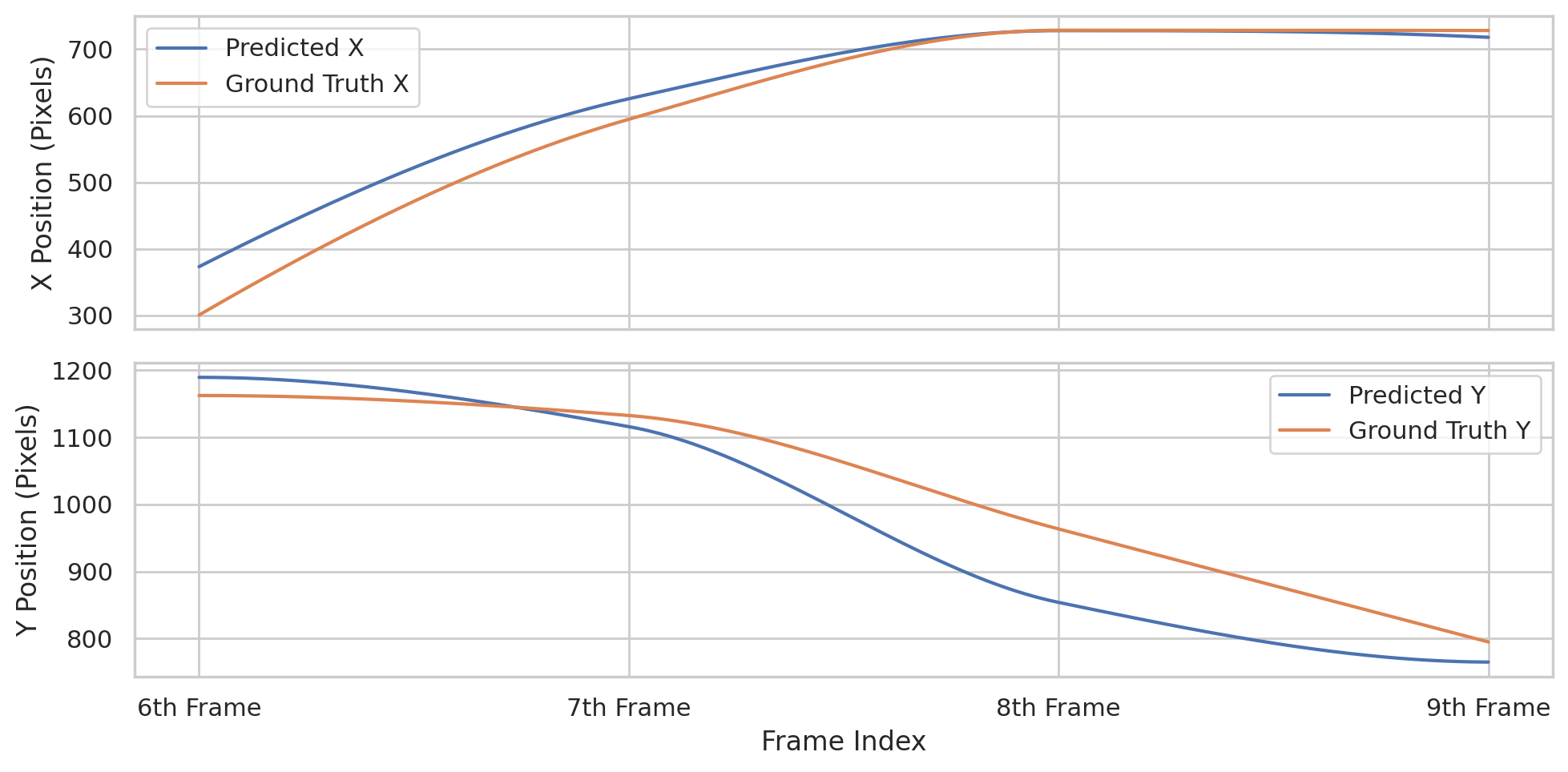}
    \caption{\textbf{Simulation analysis for \texttt{Rolling on Track} domain} \ \  (Top) Comparison between ground truth images and images generated by Chain-of-Time (quarter) for a single trial. Frame indexing begins at 6 because the first 5 frames are provided as input to the model. (Bottom) Extracted X and Y coordinates of the ball in each frame. The ground truth and simulated objects follow very similar trajectories.
}
\end{figure}

\begin{figure}[h]
    \centering
    \includegraphics[width=\linewidth]{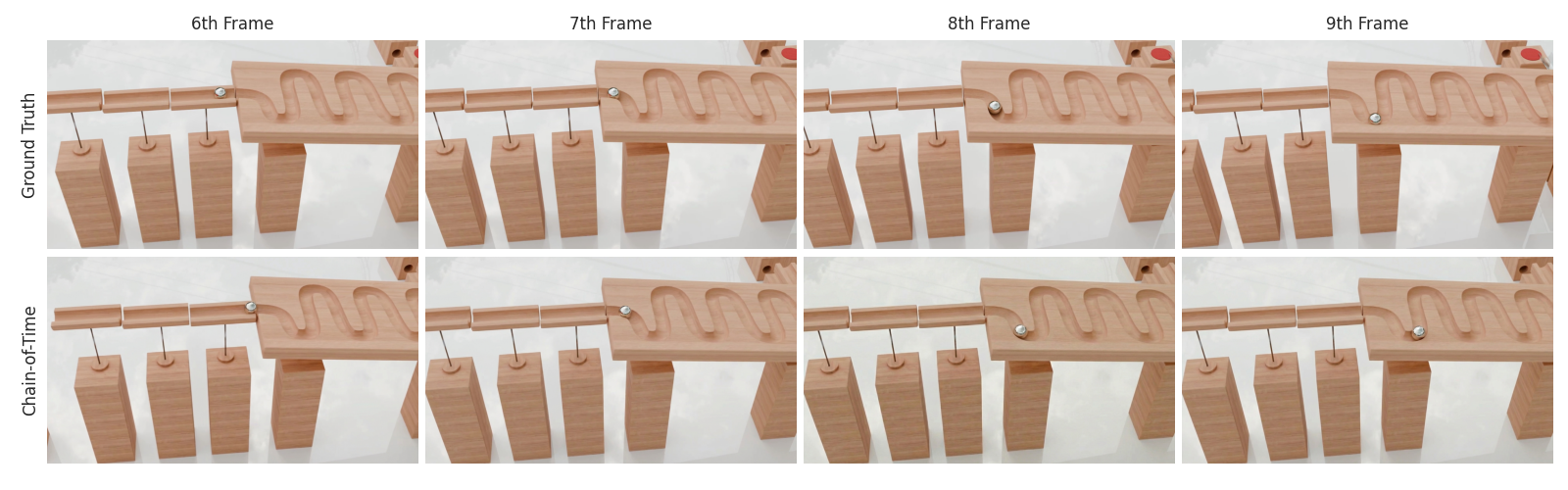}
    \includegraphics[width=\linewidth]{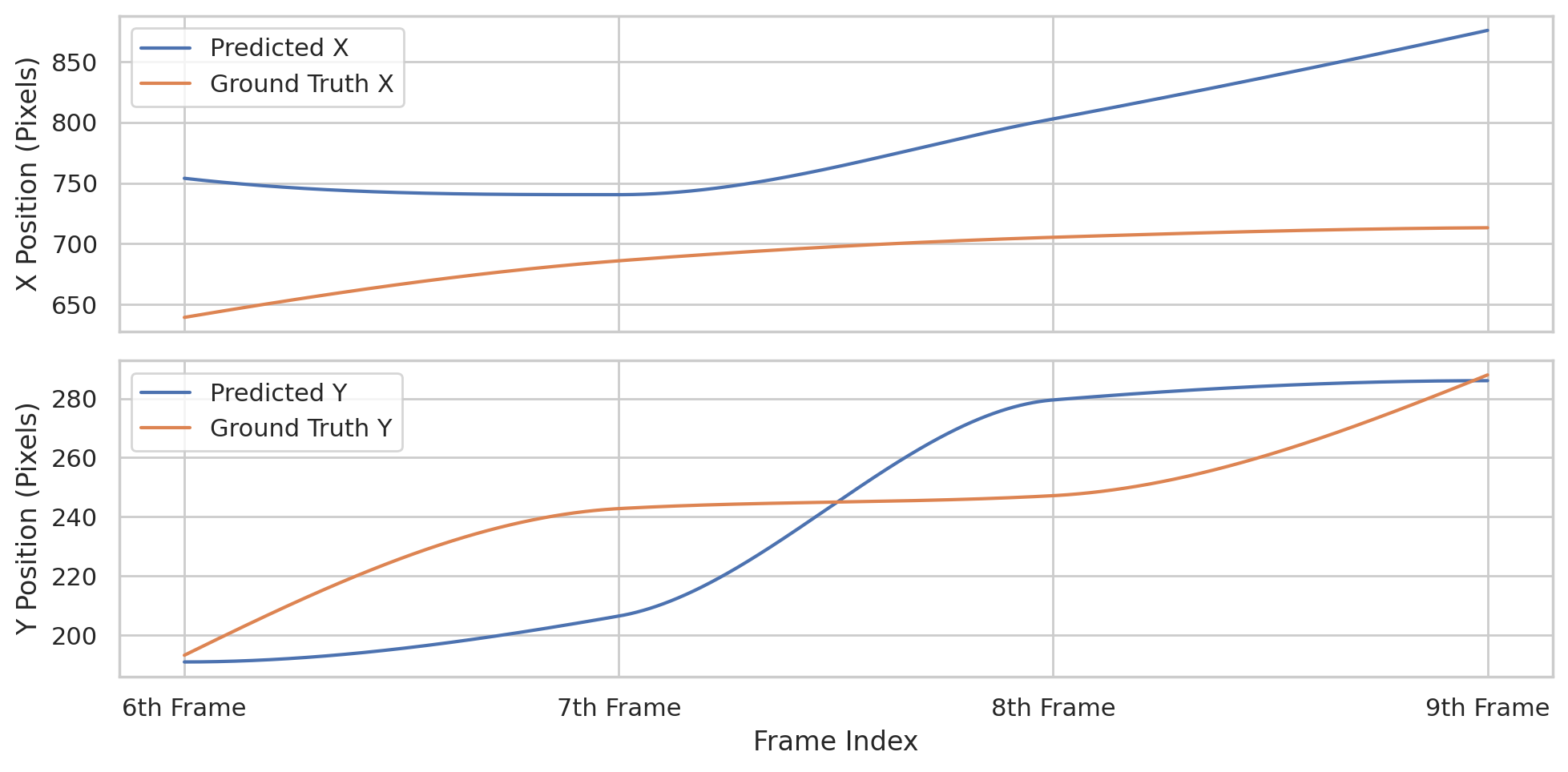}
    \caption{\textbf{Simulation analysis for \texttt{Rolling on Track} domain} \ \  (Top) Comparison between ground truth images and images generated by Chain-of-Time (quarter) for a single trial. Frame indexing begins at 6 because the first 5 frames are provided as input to the model. (Bottom) Extracted X and Y coordinates of the ball in each frame. The ground truth and simulated objects follow very similar trajectories.
}
\end{figure}

\clearpage

\subsection{The \domainI Domain}

Here are the additional trials in \domainI:

\begin{figure}[h]
    \centering
    \includegraphics[width= \linewidth]{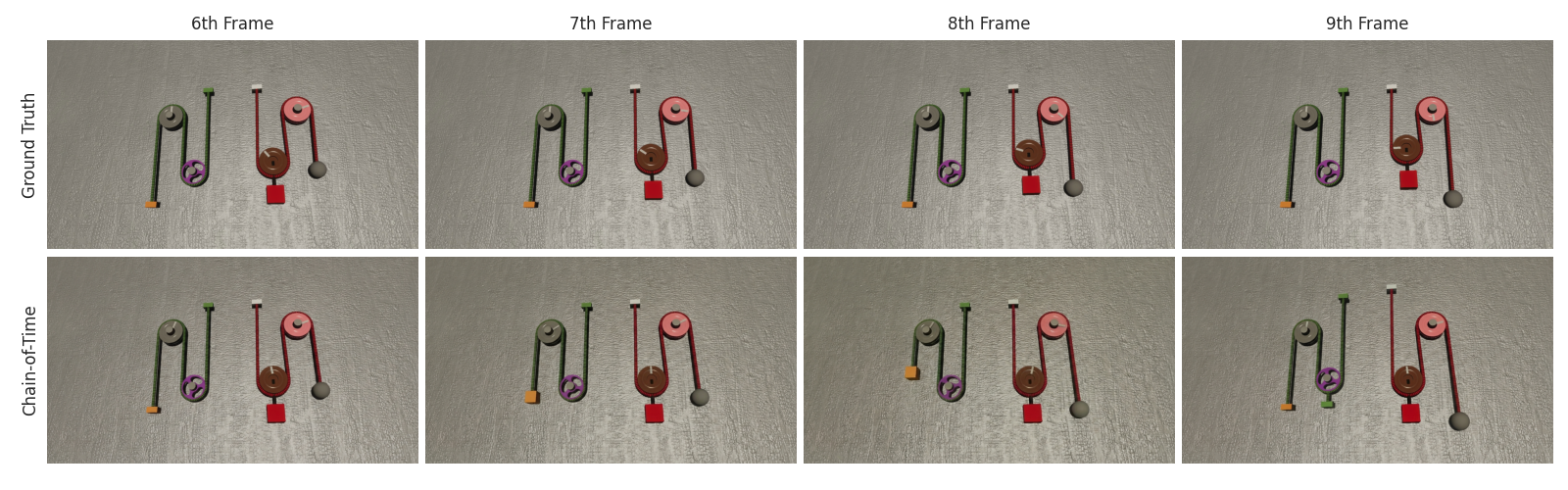}
    \includegraphics[width= \linewidth]{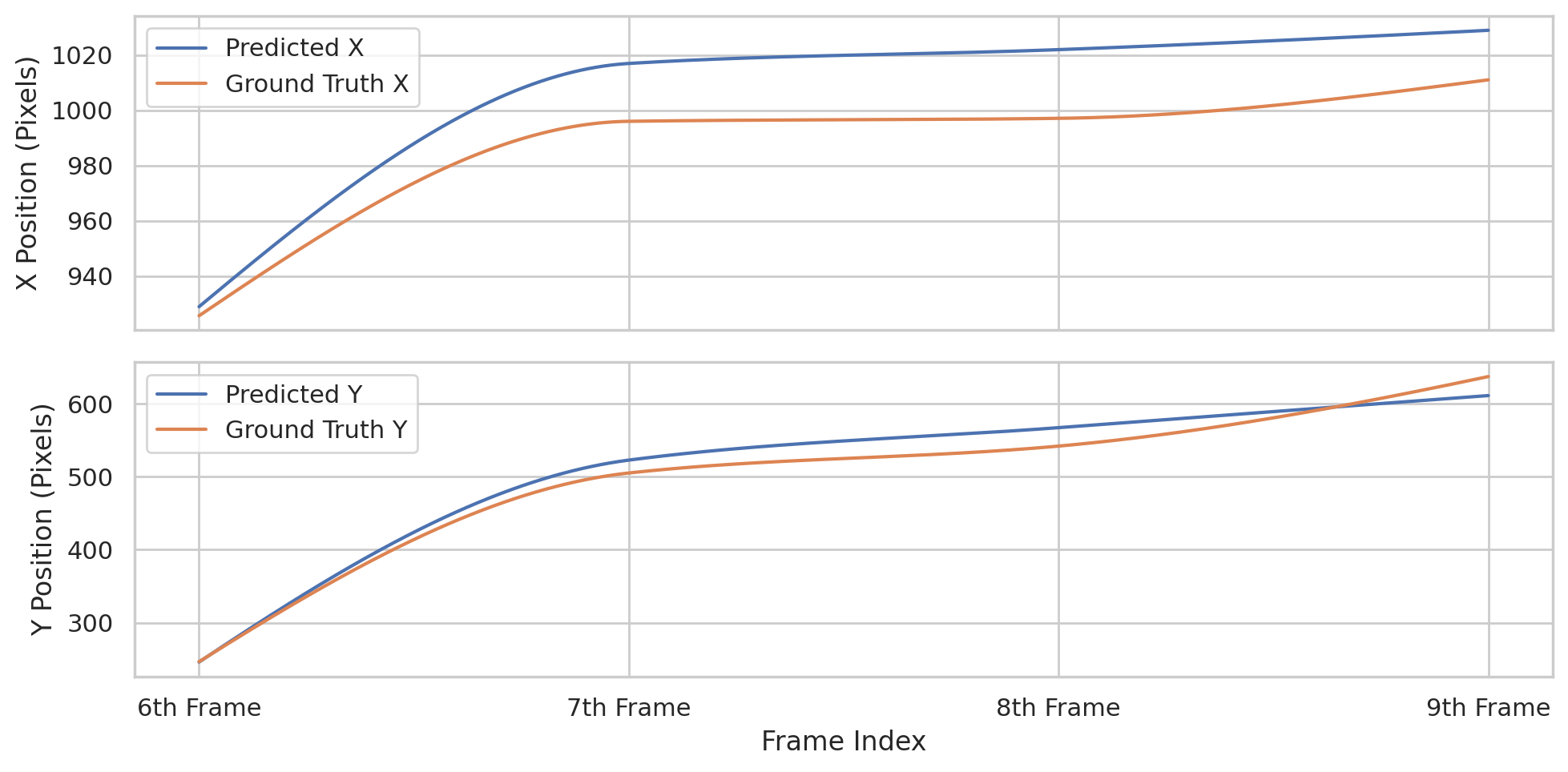}
    \caption{\textbf{Simulation analysis for \texttt{Pulley System} domain} \ \  (Top) Comparison between ground truth images and images generated by Chain-of-Time (quarter) for an example trial.  (Bottom) Extracted X and Y coordinates of the grey ball in each frame. Despite the complexity of this domain, image generation models are able to accurately simulate the ball's future motion.
}
\end{figure}

\begin{figure}[h]
    \centering
    \includegraphics[width= \linewidth]{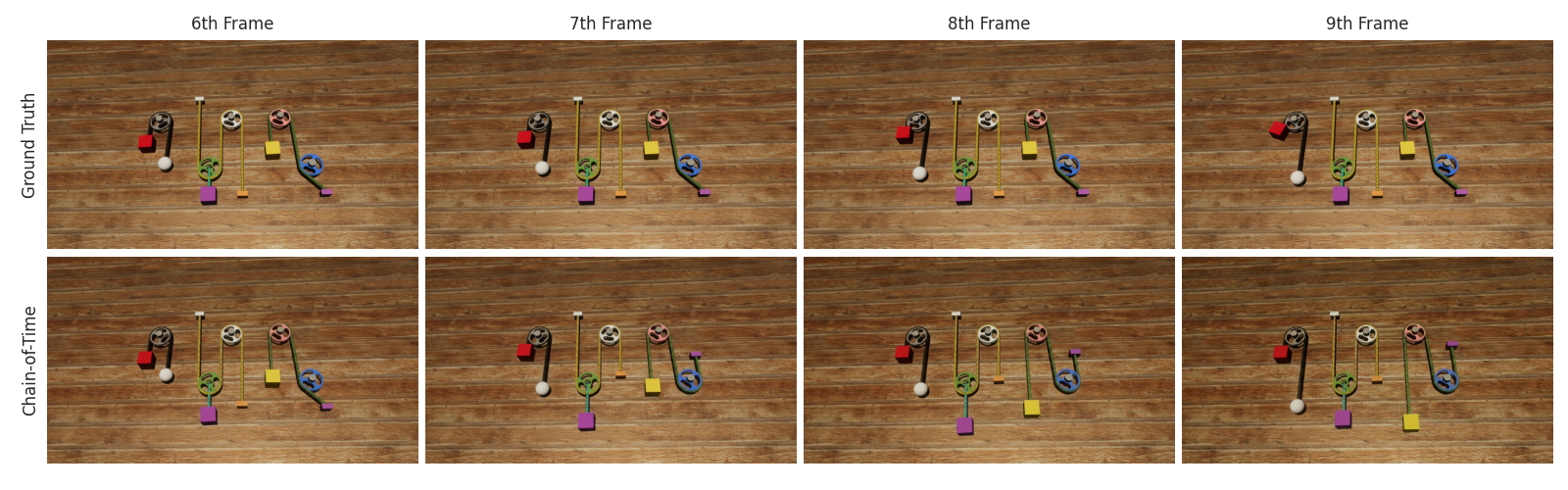}
    \includegraphics[width= \linewidth]{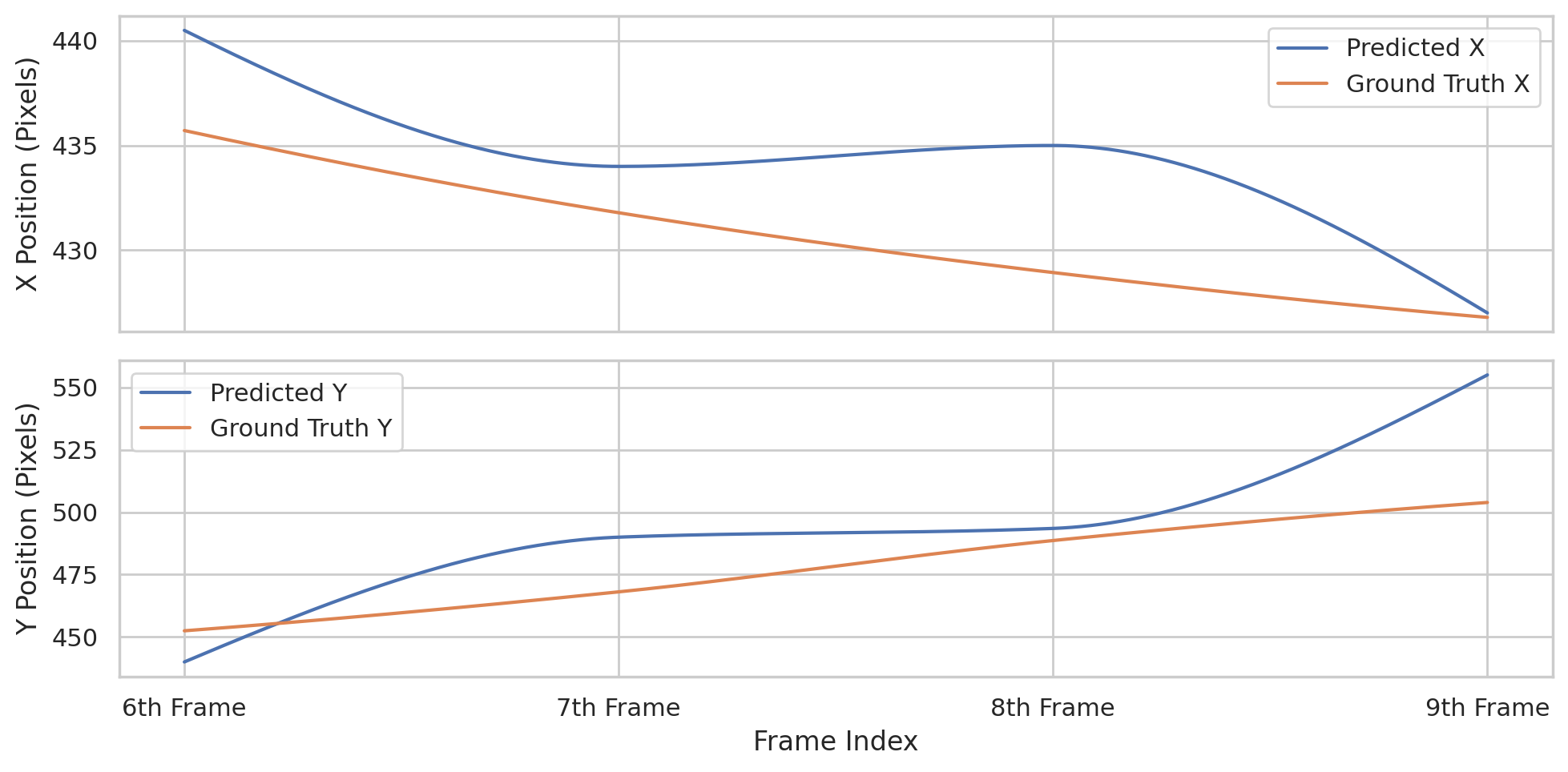}
    \caption{\textbf{Simulation analysis for \texttt{Pulley System} domain} \ \  (Top) Comparison between ground truth images and images generated by Chain-of-Time (quarter) for an example trial.  (Bottom) Extracted X and Y coordinates of the white ball in each frame. Despite the complexity of this domain, image generation models are able to accurately simulate the ball's future motion.
}
\end{figure}

\clearpage

\section{Open Source Model Results}   ~\label{app:open-source}

In addition to the closed-source models we evaluated in our main paper, we also evaluated an open source autoregressive IGM that was recently released, \texttt{HunyuanImage-3.0-Instruct}~\citep{cao2025hunyuanimage}. However, since this model is currently under development, a necessary feature for Chain-of-Time simulation -- multi-turn chat -- is not yet available as it is for closed-source models (Nano Banana Pro and GPT-Image-Gen 1.5), although it is on the roadmap in the future.

Since this is the only commercial-grade open-source autoregressive image generation model available in the market right now, we adapt our Chain-of-Time method to operate without multi-turn chat. Instead of having the IGM generate each image iteratively as a new chat message (Fig.~\ref{fig:intro-fig}), we instead append simulated frames to the list of input images, and constrain this list to a limit of 3 input images, which for now is the maximum number of input images for this model.
When multi-turn chat is added as a feature and the limit of 3 input images is expanded for this model in the near future, we will include results for this IGM along with our other models in our manuscript.

Here are the results from HunyuanImage-3.0-Instruct when we evaluated the domains with this model:

\begin{table}[ht]
\centering
\begin{tabular}{lccccc}
\hline
\textbf{Method} & \textbf{2D Gravity} & \textbf{Bouncing} & \textbf{Rolling on Slope} & \textbf{Fluid} & \textbf{Pendulum} \\
\hline
Chain-of-time Quarter & $112 \pm 5.2$ & $72 \pm 3.1$ & $103 \pm 3.2$ & $92 \pm 6.7$ & $274 \pm 4.8$ \\
Chain-of-time Half & $133 \pm 4.1$ & $87 \pm 5.9$ & $117 \pm 2.8$ & $78 \pm 4.2$ & $312 \pm 3.3$ \\
Direct Prediction & $152 \pm 6.7$ & $103 \pm 4.2$ & $128 \pm 3.6$ & $64 \pm 5.4$ & $357 \pm 3.6$ \\
\hline
\end{tabular}
\caption{RMSE results across different methods and domains.}
\label{tab:rmse_methods_domains}
\end{table}

Our results with HunyuanImage-3.0-instruct are broadly similar to our findings with Nano Banana Pro and GPT-Image-Gen. We find that Chain-of-time consistently improves the performance of HunyuanImage-3.0-instruct over the Direct Prediction baseline, with the best performance at quarter-time. As with the other IGMs, this is true for all domains except fluids.

\section{Comparison with Video Model}  \label{app:video-model}

Although our work studies IGMs, Chain-of-Time simulation is similar in some ways to video generation models, which also can be used to predict future frames given a sequence of input images. It is worth noting, though, that video models predict future frames in a dramatically different manner from Chain-of-Time, since these models typically rely on hundreds of input frames while Chain-of-Time only uses a few sparse input frames, and since video models are explicitly trained on sequential image prediction whereas IGMs are not.

We ran and analyzed a state-of-the-art open source video model (\texttt{LTX-video}) on two of our domains: the Pendulum and Rolling on Slope domains. 
Before describing our results, we would like to reiterate an important difference between our method - Chain-of-Time - and video generation models, which is that our method requires far less data as input, only a few image frames. 
When we tested video generation models with the same sparse input data that we use for Chain-of-Time, their output was incoherent, for example showing no motion at all. Therefore, in the following analysis we provided as input to LTX-video all frames of the first 2 seconds of the video and prompted the model to generate 2 seconds into the future at 30 frames per second using a prompt similar to what we have in App.~\ref{app:prompts}. We found that the simulations generated by the video model are less accurate (higher RMSE) or comparably accurate when compared to quarter-time Chain-of-Time, despite the former having 8 times more input frames and 15 times more output frames:

\begin{table}[ht]
\centering
\begin{tabular}{lcc}
\hline
\textbf{Model Type} & \textbf{RMSE in Pendulum Domain} & \textbf{RMSE in Rolling on Slope Domain} \\
\hline
LTX-Video & $143 \pm 19.1$ & $89.8 \pm 11.2$ \\
Nano Banana Pro & $169 \pm 14.6$ & $46 \pm 5.1$ \\
GPT-Image-Gen 1.5 & $243 \pm 20.2$ & $52 \pm 6.6$ \\
\hline
\end{tabular}
\caption{RMSE results across pendulum and rolling on slope domains. Three models were tested here: LTX-Video, Nano Banana Pro, and GPT-Image-Gen 1.5.}
\label{tab:rmse_results}
\end{table}

This indicates that video generation models may not possess any fundamental advantage in terms of simulation accuracy under a fixed future horizon when compared with an IGM paired with Chain-of-Time simulation, despite using substantially more computational resources.

\section{Additional Experiments}   \label{app:add-experiments}

\subsection{Varying The Number of Input Frames}   \label{app:add-num-input-frames}

We set the number of input frames to 5, but our method could theoretically work with different numbers of input frames, for example, 2,3,4,6. Therefore, we performed an additional analysis of the impact of input frames in selected domains, aiming to investigate and address the impact of the number of input frames on the final accuracy between Chain-of-time Quarter and Direct Prediction. Varying the number of input frames also changes the temporal gaps in the input. We varied the number of input frames to [2,3,4,5,6], equally spaced out across a 2-second interval, so that e.g. with 2 input frames, the frames are 2 seconds apart and thus in-distribution for our Direct Prediction baseline. Vice versa, if input frames are not equally spaced out across a 2-second interval, e.g. 3,4,5,6 input frames, such interval is out-of-distribution for our Direct Prediction baseline. Our results for the Rolling on Slope domain are shown below:

\begin{table}[ht]
\centering
\begin{tabular}{lccccc}
\hline
\textbf{Domain} & \textbf{2 Frames} & \textbf{3 Frames} & \textbf{4 Frames} & \textbf{5 Frames} & \textbf{6 Frames} \\
\hline
Rolling on Track & 12.3\% & 21.7\% & 27.3\% & 31.8\% & 33.2\% \\
Bouncing & 18.7\% & 32.7\% & 42.0\% & 49.2\% & 53.5\% \\
Rolling on Slope & 14.3\% & 35.9\% & 42.0\% & 50.5\% & 56.1\% \\
\hline
\end{tabular}
\caption{Accuracy improvements and gains over DP with Chain-of-Time across different input frame counts in three different domains.}
\label{tab:accuracy_improvements_frames}
\end{table}

We see the relative improvement in accuracy of Chain-of-Time (quarter time) compared to Direct Prediction. And according to the result above, accuracy gain diminishes as the number of in-context frames increases, and 3-5 in-context frames are sufficient to generate simulations with reasonable accuracy.

\subsection{Varying The Number of Output Frames}    \label{app:add-num-output-frames}

In our main experiment, we set the number of output frames in [4,2,1], corresponding to Chain-of-time Quarter, Chain-of-time Half, and Direct Prediction. But theoretically, the number of output frames could vary, meaning that it could be more than 4, or it could be 3. As you increase the number of output frames generated, you are increasing your computational cost too. Therefore, the relationship between the number of output frames and accuracy is important to understand.

To investigate this relationship, we performed a simple trade-off analysis between accuracy gains and token consumption. Since in our setup, the token consumption is varied only when an additional intermediary image is generated, we consider token consumption in terms of how many additional intermediate images are generated, with which we will analyze the accuracy gain (reduction in RMSE) vs one additional intermediary image generated, considering all subdomains. Here is the result:

\begin{table}[ht]
\centering
\begin{tabular}{lccc}
\hline
\textbf{Additional Image Generation} & \textbf{1} & \textbf{2} & \textbf{3} \\
\hline
Accuracy Gain & 11.07\% & 13.1\% & 14.47\% \\
\hline
\end{tabular}
\caption{Accuracy gain across additional image generations.}
\label{tab:additional_image_generation_accuracy_gain}
\end{table}

In addition, we investigated the relationship between the tokens needed for 1\% accuracy gain when additional images are generated. Here is the result:

\begin{table}[ht]
\centering
\begin{tabular}{lccc}
\hline
\textbf{Additional Image Generation} & \textbf{1} & \textbf{2} & \textbf{3} \\
\hline
Tokens needed for 1\% Accuracy Gain & 309.03 & 540.98 & 1043.86 \\
\hline
\end{tabular}
\caption{Tokens needed for each 1\% accuracy gain across additional image generations.}
\label{tab:additional_image_generation_tokens}
\end{table}

According to this analysis, we could see that the relationship between accuracy gain and token consumption is not linear, indicating that generating more images yields diminishing returns in terms of accuracy gain, and it costs more tokens for 1\% accuracy gain as accuracy gain increases.

\subsection{Analysis of Longer Time Horizons}      \label{app:add-long-horizon}

We recognize that the longer time horizon is an important topic to explore in our method. To address this point, we conducted an additional experiment which tests our method's ability when simulating longer horizons of up to 10 seconds on Pendulum and Rolling on Track domains. The results of these analyses are shown below:

\begin{table}[ht]
\centering
\begin{tabular}{lccc}
\hline
\textbf{Model Type} & \textbf{CoT Quarter RMSE} & \textbf{CoT Half RMSE} & \textbf{Direct Prediction RMSE} \\
\hline
Nano Banana Pro & $232 \pm 25.3$ & $337 \pm 24.8$ & $420 \pm 30.1$ \\
GPT-Image-Gen 1.5 & $293 \pm 31.2$ & $373 \pm 29.2$ & $436 \pm 35.9$ \\
\hline
\end{tabular}
\caption{RMSE results in the Pendulum domain.}
\label{tab:rmse_pendulum_domain}
\end{table}

\begin{table}[ht]
\centering
\begin{tabular}{lccc}
\hline
\textbf{Model Type} & \textbf{CoT Quarter RMSE} & \textbf{CoT Half RMSE} & \textbf{Direct Prediction RMSE} \\
\hline
Nano Banana Pro & $580 \pm 37.1$ & $646 \pm 25.3$ & $730 \pm 30.8$ \\
GPT-Image-Gen 1.5 & $562 \pm 34.2$ & $651 \pm 28.6$ & $710 \pm 32.7$ \\
\hline
\end{tabular}
\caption{RMSE results in the Rolling on Track domain.}
\label{tab:rmse_complex_system_domain}
\end{table}

Here, we find that Chain-of-Time still vastly outperforms Direct Prediction, although the errors are higher than we find with the 2-second data we tested in our main experiments. This suggests that our method will generalize to videos with much longer time horizons than 2 seconds.

\subsection{VLM-as-Critic For Error Correction}     \label{app:add-vlm-as-critic}

In Section~\ref{sec:results}, we found that error accumulates as the time horizon increases. Here, we investigate one method that could potentially serve as the method to reduce error accumulation.

We implemented a VLM-as-critic strategy~\footnote{We thank an anonymous reviewer for this suggestion.}, aiming to correct incremental errors during the Chain-of-Time. A VLM is implemented as a \textit{``critic''} to verify if an intermediary image is physically plausible, considering the previous images (input images + previously generated intermediary images). Running a preliminary experiment on the Pendulum domain, we found that on average the accuracy (RMSE) improved by 18\%, and the VLM reduces the accumulation of visual artifacts and physical inaccuracies, thus stabilizing the simulation process. In future work building on Chain-of-time, we plan to explore the potential of a VLM as a source of reward signals for both inference-time simulation and model fine-tuning.

\subsection{New Baseline with Self-Consistency Selection with CoT Reasoning over Multiple Sampled Images}

In the original Direction Prediction baseline, we averaged across all samples in each domain and calculated the average RMSE as an interval. But here, we aim to create an different baseline using self-consistent selection with CoT reasoning. For each stimuli within each domain, direct prediction will generate N=10 FINAL FRAMES. We use these N=10 generated FINAL FRAMES from Direct Prediction as the input along with the original 5 input frames back to the VLM. We use a VLM-as-judge that uses CoT reasoning to determine which of these images is the most accurate FINAL FRAME. The final RMSE for each stimuli within each domain is calculated using this final frame. Averaging across all stimuli for each domain, this forms a new baseline. Here is the comparison between the experimental data shown in Fig.~\ref{fig:mse} and the new baseline.

\begin{table}[htbp]
\centering
\label{tab:domain_results}
\resizebox{\textwidth}{!}{%
\begin{tabular}{lcccccccc}
\toprule
\textbf{Domains} &
\textbf{Rolling on Track} &
\textbf{Bouncing} &
\textbf{Rolling on Slope} &
\textbf{Pulley System} &
\textbf{2D Gravity} &
\textbf{Pendulum} &
\textbf{Fluid} &
\textbf{Collision} \\
\midrule

Nano-Banana Pro (Original Baseline)
& $667 \pm 53$
& $93 \pm 5.7$
& $98 \pm 6.2$
& $462 \pm 10.4$
& $226 \pm 12.8$
& $288 \pm 47.2$
& $34 \pm 2.3$
& $142 \pm 12.4$ \\

Nano-Banana Pro (New Baseline)
& $703$
& $108$
& $94$
& $502$
& $197$
& $253$
& $23$
& $152$ \\

Nano-Banana Pro (Chain-of-Time Half)
& $582 \pm 51$
& $72 \pm 5.2$
& $69 \pm 6.2$
& $411 \pm 15.1$
& $74 \pm 12.5$
& $187 \pm 16.9$
& $43 \pm 1.9$
& $104 \pm 10.6$ \\

Nano-Banana Pro (Chain-of-Time Quarter)
& $428 \pm 27$
& $65 \pm 4.3$
& $46 \pm 5.1$
& $339 \pm 11.9$
& $65 \pm 8.4$
& $169 \pm 14.6$
& $72 \pm 1.8$
& $58 \pm 9.9$ \\

\midrule

GPT-Image-1.5 (Original Baseline)
& $693 \pm 97$
& $74 \pm 2.8$
& $112 \pm 9.7$
& $378 \pm 8.1$
& $246 \pm 11.7$
& $306 \pm 57.1$
& $37 \pm 1.8$
& $152 \pm 10.9$ \\

GPT-Image-1.5 (New Baseline)
& $682$
& $57$
& $103$
& $355$
& $214$
& $317$
& $34$
& $131$ \\

GPT-Image-1.5 (Chain-of-Time Half)
& $562 \pm 68$
& $48 \pm 3.7$
& $88 \pm 7.2$
& $320 \pm 18.5$
& $93 \pm 11.2$
& $260 \pm 24.5$
& $47 \pm 2.5$
& $90 \pm 12.1$ \\

GPT-Image-1.5 (Chain-of-Time Quarter)
& $446 \pm 31$
& $29 \pm 4.2$
& $52 \pm 6.6$
& $297 \pm 13.2$
& $80 \pm 9.4$
& $243 \pm 20.2$
& $90 \pm 3.5$
& $68 \pm 9.9$ \\

\bottomrule
\end{tabular}%
}
\caption{Prediction errors for the eight domains included in this analysis, averaged across all data for each domain, compared to the new baseline. Prediction error is measured by taking the average RMSE between the ground-truth positions (location of focal object, or water level) and the final positions predicted by the IGM. Error bars are 95\% Bootstrapped CI. We generally find a monotonic relationship between \textit{Chain-of-Time} precision and performance when compared against the new baseline as well, consistent with the results in Fig.~\ref{fig:mse}.}
\label{new_baseline}
\end{table}
\label{app:new_baseline}
\section{Statements}  ~\label{app:statements}

\subsection{Language Model Use}

LLMs were used in this work for literature review, for coding assistance with constructing the computer vision algorithm described in App.~\ref{app:fluid-cv-algo}, and for proofreading the manuscript.

\subsection{Reproducibility}

Code and data will be shared with the camera-ready version of this work. While the reproducibility of our results on closed source IGMs is limited by whether these models are still publicly available in the future, in App.~\ref{app:open-source}, we show results with an open-source IGM. The closed-source IGMs were tested using their respective official APIs, and the LTX-Video was run on a single A100 for 3 hours.

\subsection{Impact and Risks}

We provide methods for improving physical simulation in IGMs, and for systematically evaluating accuracy in simulated frames.
While our results are unlikely to have immediate societal impacts, in the future, this may shape how we evaluate physical reasoning and related abilities in IGMs. Further, this work paves the way towards more powerful IGMs that may optimize directly for Chain-of-Time simulation ability, similar to how reasoning LLMs optimize directly for Chain-of-Thought Reasoning ability~\citep{jaech2024openai}.

\end{document}